\documentclass{article}

\PassOptionsToPackage{authoryear,round}{natbib}
\usepackage[preprint]{icml2026}

\usepackage{cuted}
\usepackage{capt-of}
\usepackage{subcaption} 
\usepackage{standalone}
\usepackage{multirow} 
\usepackage[utf8]{inputenc} 
\usepackage[T1]{fontenc}    
\usepackage{hyperref}       
\usepackage{xurl}
\usepackage{booktabs}       
\usepackage{amsfonts}       
\usepackage{nicefrac}       
\usepackage{microtype}      
\usepackage{xcolor}         
\usepackage{tikz}
\usepackage{amsmath} 
\usepackage{float}
\usepackage{ragged2e}
\usepackage{booktabs}
\usepackage{caption}
\usepackage{siunitx} 
\usepackage[normalem]{ulem}
\usepackage{soul}
\def\losslandscape(#1){0.2*sin(deg(5*#1+2))-0.6*exp(-8*(#1+1.0)^2)+0.2
                      -0.5*exp(-8*(#1-0.6)^2)+0.1*sin(deg(10*#1+1))}

\usepackage{afterpage}
\usepackage[nodisplayskipstretch]{setspace}
\usepackage{enumitem}

\icmltitlerunning{Telescope: Improving Zero-Shot Detection of LLM Generated Content By Measuring Token Repetition Probability}

\begin{document}

\twocolumn[
  \icmltitle{Telescope: Improving Zero-Shot Detection of LLM Generated Content By Measuring Token Repetition Probability}



  \icmlsetsymbol{equal}{*}

  \begin{icmlauthorlist}
    \icmlauthor{Christopher Nassif}{equal,ECE}
    \icmlauthor{Josh F. Cooper}{equal,ISE}
  \end{icmlauthorlist}

  \icmlaffiliation{ISE}{Virginia Tech Department of Industrial and Systems Engineering}
  
  \icmlaffiliation{ECE}{Virginia Tech Department of Electrical and Computer Engineering}

  \icmlcorrespondingauthor{Josh F. Cooper}{joshfcooper@vt.edu}

  \icmlkeywords{LLM Text Detection, Interpretability}

  \vskip 0.3in
]

\printAffiliationsAndNotice{\icmlEqualContribution}
















\begin{abstract}
Distinguishing Large Language Model (LLM) generated text from human writing is a critical and difficult challenge. While LLMs are trained to write like humans, we hypothesize that this training leaves an indelible mark. LLMs develop a particularly strong aversion to token repetition very early in training. This bias persists as a ``Vestigial Heuristic'' (a developmental artifact) that is activated in LLM generated text, separating LLM from human writing. To probe this phenomenon, we introduce Telescope Perplexity, a metric that evaluates the token repetition of the model, $P(s_i | s_{1:i})$.  Our empirical investigation reveals that the Telescope Perplexity signature emerges early in pre-training, and Telescope Perplexity effectively enables highly effective zero-shot LLM detection. We show state-of-the-art or competitive performance across diverse datasets (including modern evaluation sets we introduce), reference models, and perturbation schemes with greater efficiency than other methods. 
\end{abstract}

\section{Introduction}

\subsection{Background and Motivations}

Large Language Models (LLMs) captured the attention of the general public in 2022 when OpenAI released ChatGPT. This step forward showed the world how well deep learning models could learn to understand and respond to human text \cite{ouyang2022training}. In the time since ChatGPT was released, the industry around Artificial Intelligence has been thrown into the cultural zeitgeist, while the capabilities of large language models have improved significantly. Language models have continued to become smarter \cite{hoffmann2022trainingcomputeoptimallargelanguage}, faster \cite{dao2022flashattentionfastmemoryefficientexact}, and cheaper \cite{cai2024surveymixtureexperts}. They are now being used constantly for tasks from real time translation to helping explain difficult topics and problems for students. However, LLMs have also been used for deploying spam bots \cite{liyanage-etal-2024-detecting}, disseminating fake news online \cite{sallami2024deceptiondetectiondualroles}, and writing essays for students \cite{jelson2025empiricalstudyunderstandstudents}. Humans simply cannot distinguish between LLM and human written text \cite{ippolito2020automaticdetectiongeneratedtext}, so algorithmic methods are needed to help control misuses of LLMs, and ideally these methods will continue to be effective and keep pace with new model releases. For this reason, we focus on \textbf{Zero-Shot Detection}. Zero-shot methods perform without the need to continuously update their training data, making them ideal for handling the constant churn of model releases.


\subsection{Core Contributions}
In this work, we propose and investigate the ``Vestigial Heuristics'' hypothesis, which presents a novel way to view how early training dynamics can affect the final model's behaviors. We introduce Telescope Perplexity (Eq.~\ref{eq:telescope_ppl}) as a simple, conceptually grounded probe specifically targeting one of these hypothesized \textbf{Vestigial Heuristics} that lead to a particular and extreme aversion to repetition. Additionally, we provide extensive testing to verify our hypothesis of the ``Vestigial Heuristics'' and the signature's early emergence/ locality, which allows us to better understand how Telescope Perplexity works, where it can fail, and how it can be improved in the future. Finally, we provide extensive empirical validation demonstrating Telescope Perplexity's state-of-the-art or competitive zero-shot detection performance and robustness across diverse scenarios, including on novel evaluation sets using contemporary LLMs (GPT4o Mini, Deepseek-V3). The code required for reproducing all of the experiments of this paper can be found in the following GitHub repository: \url{https://github.com/ChrisNassif/Telescope}.

\section{Related Work}

Techniques for detecting whether or not a piece of text is generated by some unknown \hyperref[Vocabulary]{``target model''} (also known as a black-box setting) can largely be grouped into 2 main categories: zero-shot detectors or supervised classifiers.

Zero-shot detectors run inference on the text with a ``reference model'' and then analyze the output tokens of that reference model using some statistical technique. The goal of the statistical analysis is to try to measure some fundamental and interpretable pattern behind LLM generated text that exists for a wide variety of target language models. Additional clarification on the relationship between reference models and target models can be found in the Appendix Section \ref{Vocabulary}. 

Supervised detectors utilize some trained machine learning model to identify whether or not the text originates from an LLM. These supervised detectors can either use zero-shot detectors as a heuristic input to the model to learn patterns that signal whether a piece of text is LLM generated \cite{verma2024ghostbusterdetectingtextghostwritten} or they can be directly finetuned from another language model to directly operate on the text without using any zero-shot heuristics \cite{su2024hc3plussemanticinvarianthuman}. The major limitation that these supervised detectors struggle with is that as new target models are released, they may quickly become obsolete and struggle to detect the outputs from new target LLMs. For this reason, we choose to focus this work on just zero-shot methods, and only benchmark against zero-shot methods.

Early attempts at creating a zero-shot detector for LLM generated text consisted of using the log Perplexity or loss of a \hyperref[Vocabulary]{reference language model} to detect text generated by a \hyperref[Vocabulary]{target language model}. As previous work by \citet{xu2024investigatingefficacyperplexitydetecting} has shown, a higher Perplexity score from a reference model means that the piece of text is less likely to be output by the reference model. Higher Perplexity can also help indicate that it has never seen any piece of text in its training similar to the piece of text in inference time, since large language models tend to write in distributions similar to the distributions found during training. Additionally, since large language models are generally trained on very similar, wide ranging datasets, if a reference LLM is perplexed by a piece of text and has never seen anything like it before, that means that other language models likely haven't seen anything like it in their training data. If other language models are less likely to have seen similar text, they would also be less likely to write the text \cite{Grammar-Learning}, and this creates a separation between human written and LLM generated text.  

Alongside Perplexity based techniques, there also exist rank based techniques, such as DetectLLM LRR \cite{su2023detectllmleveraginglogrank}, which utilize the log rank of a logits distribution. The rank of a token in a distribution is the index of a token in a logits distribution sorted by their probabilities. The token with the highest probability in a distribution has rank 1, the token with the second highest probability in a distribution has rank 2, etc. The log rank is simply the natural logarithm of the computed rank. Using token rank, \citet{su2023detectllmleveraginglogrank} devised a novel detector called LRR (Log Rank Ratio), which is highly accurate at distinguishing LLM generated and human written text. 

\citet{su2023detectllmleveraginglogrank} introduced a second technique named DetectLLM-NPR combining rank information and text perturbation. Perturbation based techniques perturb the text in some way through word swaps or minor rewrites, and then analyze how zero-shot statistics change as you perturb the text. DetectLLM-NPR and similar techniques like DetectGPT \cite{mitchell2023detectgptzeroshotmachinegeneratedtext} are able to achieve high performance detecting LLM generated text, but at a high computational cost. DetectGPT and DetectLLM-NPR each need to inference the reference model many times to get good performance, which balloons the inference costs tremendously. We choose not to benchmark against perturbation based techniques, since the tremendous computational cost combined with our testing across a wide variety of datasets and reference models would exceed our available compute.

More recent zero-shot work avoids the perturbation cost of DetectGPT
and DetectLLM-NPR. Fast-DetectGPT \cite{bao2024fastdetectgpt}
replaces perturbation with a single sampling step, scoring text via
conditional probability curvature, while DNA-GPT
\cite{yang2024dnagpt} regenerates a truncated continuation and
measures its $n$-gram or probability divergence from the original.
Both extract a strong signal from few forward passes but do not
target repetition aversion specifically. Beyond statistical scoring,
\citet{ackerman2025inspection} show an instruct-tuned Llama3-8B
can recognize its own writing via an identifiable residual-stream
direction, consistent with our claim that LLM generated text leaves
a learned, locally detectable trace. We refer the
reader to \citet{wu-etal-2025-survey} for a broader survey.
The state of the art in zero-shot black box detectors was achieved using a key refinement to Perplexity. This technique is called the Binoculars Score \cite{hans2024spottingllmsbinocularszeroshot}, and it attempts to normalize the Perplexity score with a Cross Perplexity score from two models, so that it works better across domains. Computing the Binoculars Score requires two language models instead of one, which doubles the compute requirements relative to simply using Perplexity; however, \citet{hans2024spottingllmsbinocularszeroshot} show that it is worth the extra cost since the performance gains are very significant. To the best of our knowledge, the Binoculars Score with a Falcon 7B reference model currently stands as the state-of-the-art for zero-shot detection of LLM generated text.

\newpage

\section{Understanding Vestigial Heuristics}
\label{sec:telescope_probe} 

It is well understood that neural networks often learn simpler, local patterns before mastering long-range dependencies \cite{Grammar-Learning, belrose2024neuralnetworkslearnstatistics}. This suggests a compelling question: 

\begin{quote}
\centering
    \textit{Do optimization pressures during early LLM pre-training instill fundamental statistical biases that are never unlearned later in training?}
\end{quote}

We know from previous work by \citet{belrose2024neuralnetworkslearnstatistics} that most language models start out in early training as bigram models, directly predicting the next token from the one or maybe two tokens that precede it. We hypothesize that early in training, while they are developing these bigram models, LLMs learn an important heuristic that aids them in modeling text as a bigram model: tokens very rarely repeat themselves. Therefore, these early stage bigram models will consistently have a uniquely strong aversion to token repetition as they learn that tokens very rarely repeat themselves in natural language.

A couple key results from previous work show that what these bigram models learn persists throughout later stages of training, and that most LLMs will develop the same bigram model as one another even though they are trained through different datasets and architectures \cite{Grammar-Learning}, \cite{belrose2024neuralnetworkslearnstatistics}. This heuristic persists over time even though it is often not entirely necessary to vehemently reject token repetitions at early stages of training. We call this a \textbf{Vestigial Heuristic} because it was developed early in training to aid the initial bigram model but is not strictly needed by the model in later stages of training.
Both the target and reference model will have a ``Vestigial Heuristic'' that prevents token repetition, so the text produced by the target model will activate the neurons responsible for the ``Vestigial Heuristic'' more strongly than in human text, which does not have the same ``Vestigial Heuristic''.

To investigate this hypothesis and its power in separating human written and LLM generated text, we propose Telescope Perplexity. While the formulation is strikingly simple, we argue its power lies in probing the specific hypothesized aversion to repetition that forms very early in training. 

\newpage

\begin{figure}
    \centering
    \includegraphics[width=1\linewidth]{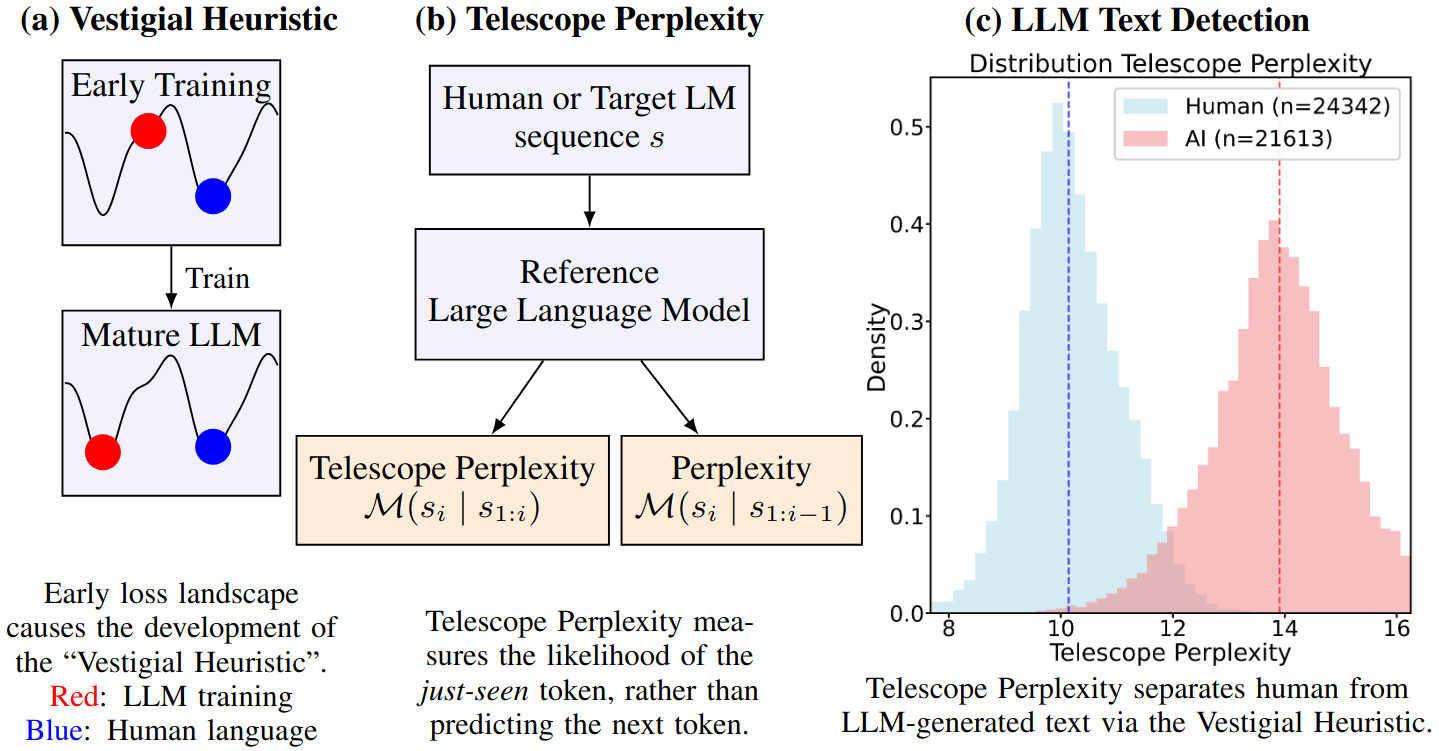}
    \caption{An overview of the ``Vestigial Heuristic'' hypothesis and measuring the average token repetition probability for the Telescope Perplexity.}
    \label{fig:overview}
\end{figure}

Telescope Perplexity measures the \hyperref[Vocabulary]{reference language model}'s learned likelihood of outputting the last token it processed given its context up to that point. For a language model $\mathcal{M}$ and a token sequence $\vec{s} = (s_1, ..., s_L)$ of length $L$, the Telescope Perplexity is defined as:
\begin{equation}
    \label{eq:telescope_ppl}
    \text{Telescope PPL}_\mathcal{M}(\vec{s}) = -\frac{1}{L} \sum_{i=1}^{L} \log \mathcal{M}(s_{i} \mid s_{1:i})
\end{equation}
$\mathcal{M}(s_{i} \mid s_{1:i})$ is the probability assigned by the model $\mathcal{M}$ to token $s_i$ given the sequence $s_1...s_i$. This provides a targeted look into the model's token repetition likelihood, whereas standard Perplexity works by predicting the \textit{next} token ($\mathcal{M}(s_i | s_{1:i-1})$). Figure \ref{fig:overview} provides a conceptual overview of the Telescope Perplexity. Please note that the Telescope Perplexity deviates from standard Perplexity due to the fact that the negative log likelihood is not exponentiated; however, this is more efficient to compute and does not affect any detection results with a fixed threshold since exponentiation is a monotonic function.

\subsection{Training Dynamics}
\label{subsec:temporal_origins}

\begin{figure}[h]
    \centering
    \begin{subfigure}{0.4\textwidth}
        \centering
        \includegraphics[width=\textwidth]{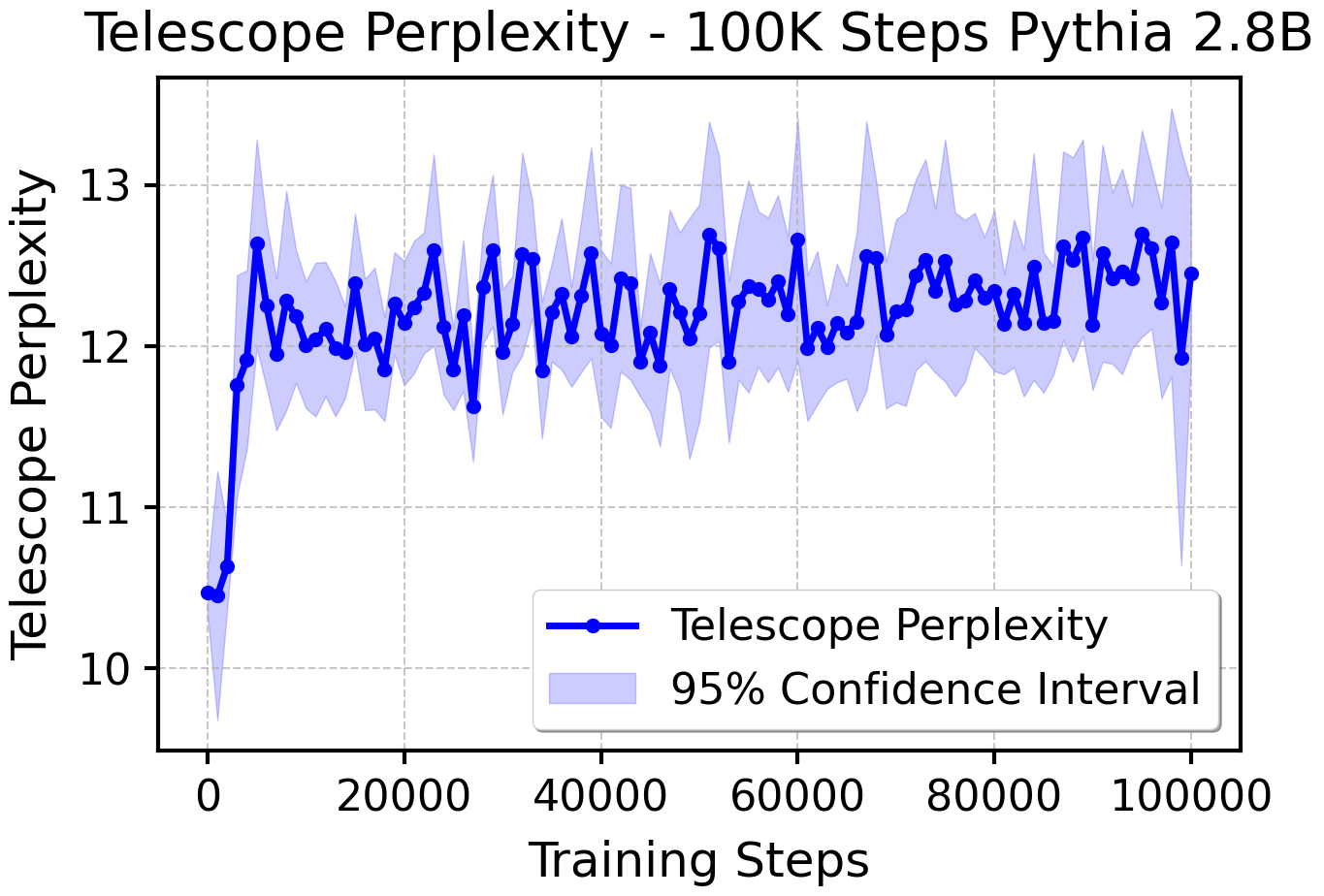}
        \caption{Final Pythia checkpoint as reference model}
        \label{fig:pythia-wrt-pythia}
    \end{subfigure}
    \hfill
    \begin{subfigure}{0.4\textwidth}
        \vspace{0.4cm}
        \centering
        \includegraphics[width=\textwidth]{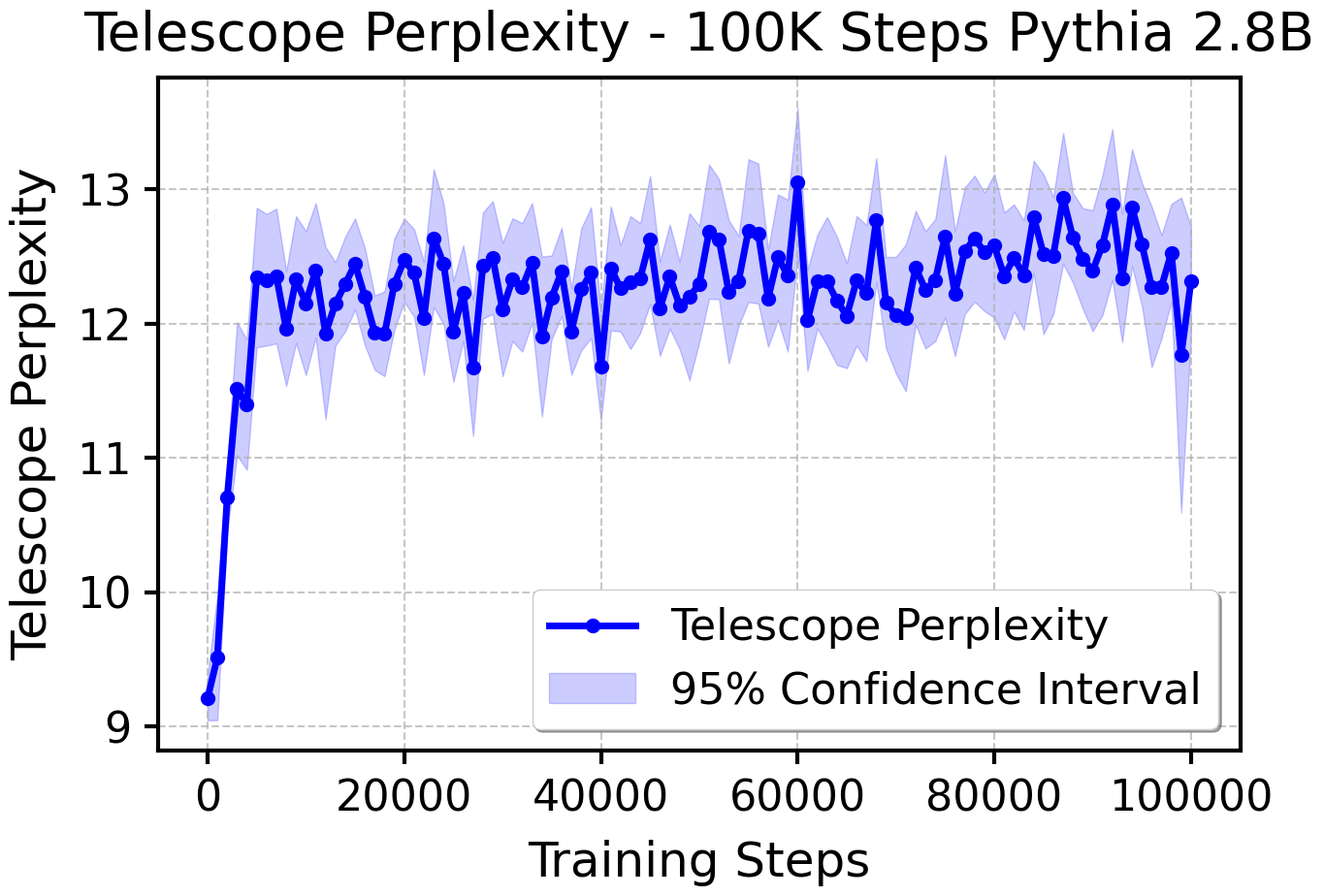}
        \caption{SmolLM 360M as reference model}
        \label{fig:pythia-wrt-smollm}
    \end{subfigure}
    \caption{Telescope Perplexity evaluated on text generated by Pythia-2.8B model checkpoints throughout training. Note the early stabilization of the Telescope Perplexity.}
    \label{fig:Training_Pythia}
\end{figure}

If the statistical biases Telescope Perplexity detects are indeed ``Vestigial Heuristics'' established early in training, we would expect the signature to emerge relatively quickly and remain stable throughout the later stages of pre-training. To investigate this, we analyzed the Telescope Perplexity of generated text using checkpoints from different stages of LLM pre-training for Amber-7B \cite{liu2023llm360} and Pythia \cite{biderman2023pythia}. Figure~\ref{fig:Training_Pythia} shows the evolution of Telescope Perplexity over text generated using the training checkpoints of Pythia-2.8B.

The results in Figure~\ref{fig:Training_Pythia} reveal a key characteristic: the Telescope Perplexity rises sharply early in training and then largely plateaus, indicating stability through later training stages. Similar stabilization was observed across various model sizes and architectures in Appendix~\ref{app:additional_training_dynamics}. This early emergence and subsequent persistence strongly support the hypothesis that Telescope Perplexity captures a ``Vestigial'' characteristic established during the foundational learning phase, rather than a property that evolves continuously with model capability.

\subsection{Signature Locality}
\label{subsec:structural_basis}

The ``Vestigial Heuristics'' hypothesis posits that the aversion to repetition that we are probing relates to local, statistical patterns learned early in training. If Telescope Perplexity primarily captures behaviors from an early stage bigram model, its effectiveness should largely persist even when the reference model is provided with very limited context. We tested this by computing Telescope Perplexity and standard Perplexity using only the preceding one (bigram context) or two (trigram context) tokens, compared to using the full preceding context available. Table~\ref{tab:limited_context} presents the detection performance (AUC) on the HC3 dataset under these conditions.

Table~\ref{tab:limited_context} reveals that Telescope Perplexity maintains a remarkably high AUC (0.897) even with purely bigram context, significantly outperforming standard Perplexity under the same constraint (0.761 AUC). This strong performance with extremely limited context confirms that the signature Telescope Perplexity measures is indeed fundamentally rooted in \textit{local} token relationships, aligning with the hypothesis that it reflects simple pattern-formation biases learned early in training.

\begin{table}[htbp]
\small
\centering
\caption{Detection performance (AUC) on the HC3 dataset using Telescope Perplexity and standard Perplexity with full context versus limited (Bigram, Trigram) context provided to the reference model (SmolLM 360M). Confidence Intervals (95\%) are shown in parentheses. Results demonstrate Telescope Perplexity's strong reliance on local information.}
\label{tab:limited_context} 
\begin{tabular}{@{}lcc@{}}
\toprule
\textbf{Method} & \textbf{Context} & \textbf{AUC (95\% Confidence Interval)} \\
\midrule
Telescope & Full & 0.995 (0.993-0.996) \\
Telescope & Bigram & 0.897 (0.893-0.905) \\
Telescope & Trigram & 0.921 (0.915-0.926) \\
\midrule
Perplexity & Full & 0.991 (0.990-0.993) \\
Perplexity & Bigram & 0.761 (0.753-0.770) \\
Perplexity & Trigram & 0.925 (0.915-0.926) \\
\bottomrule
\end{tabular}
\normalsize
\end{table}

\subsection{Generality of the Phenomenon}
\label{subsec:generality}

Is the ``Vestigial Heuristic'' detected by Telescope Perplexity an idiosyncrasy of specific models or a more general artifact? Our experiments provide two lines of evidence for generality. First, as detailed in Section~\ref{sec:empirical_validation} (Table~\ref{tab:main_auroc_results}), Telescope Perplexity demonstrates strong detection performance when using a wide variety of reference models, spanning different architectures (Gemma, Llama, Falcon, SmolLM, etc) and parameter counts (135M to 9B). This suggests the underlying signature is not confined to one model family.

Second, using the SmolLM 360M model and its distinct training corpus components FineWeb (human written data) and Cosmopedia V2 (synthetic LLM data) \cite{benallal2024smollmcorpus}, we found that Telescope Perplexity could distinguish text originating from these different internal training sources with very high accuracy (AUC 0.996, F1 0.987). See Appendix \ref{appendix:Training_Data_Separability} for additional details. This confirms that models \textbf{do} internalize fine-grained statistical properties related to their training experience, lending credence to the idea that persistent, detectable artifacts like ``Vestigial Heuristics'' can indeed be generally learned. Importantly, this demonstrates that we are not measuring the text's similarity to the training data, but a property learned in training! 

Taken together, these results suggest that the persistent local biases probed by Telescope Perplexity are likely a relatively general characteristic associated with current deep learning approaches to language modeling.


\section{Experimental Setup}
\label{sec:experimental_setup}

We formulate LLM generated text detection as a \textbf{binary
classification problem}: given a passage $\vec{s}$, decide whether it
was produced by a large language model or written by a human. Each
detector we study outputs a single real-valued
score $f(\vec{s})$ that is intended to be larger (or smaller, depending
on sign convention) on LLM generated text than on human text. A
classifier is obtained by thresholding: predict ``LLM generated'' iff
$f(\vec{s}) > \tau$ for some threshold $\tau$, and ``human written''
otherwise.

This framing motivates the two complementary metrics we report in
Section~\ref{sec:empirical_validation}. \textbf{AUROC} measures the
detector's ability to rank LLM generated text above human written text
\emph{without committing to any particular $\tau$}; it is the most
informative single summary of a score's discriminative power and lets us
compare detectors without considering how the
threshold is set. \textbf{F1-Score}, by contrast, requires a concrete
$\tau$ and so reflects classification performance at an operating
point. We report (i) the maximum F1 achievable on each test set,
which characterises the best-case behaviour of the score, and (ii) a
\emph{Transferability F1} that fixes $\tau$ from a logistic regression
fit on all \emph{other} datasets, so as to characterise how well a
threshold chosen without access to the target distribution generalises.
Together these answer two practically relevant questions: ``how
separable are the two classes under this score?'' (AUROC) and ``how
well does a threshold calibrated elsewhere transfer to a new domain?''
(Transferability F1).

To empirically validate the effectiveness of Telescope Perplexity in
detecting the hypothesized ``Vestigial Heuristics'' and compare it
against existing methods, we designed a comprehensive experimental
setup covering diverse datasets, contemporary language models, and
evaluation procedures.

\subsection{Datasets}
\label{subsec:datasets}
We gathered an array of reputable datasets that distinguish between AI-generated text and human-generated text. These datasets include HC3 \cite{guo-etal-2023-hc3}, HC3 Plus \cite{su2024hc3plussemanticinvarianthuman}, a popular Kaggle competition dataset named LLM - Detect AI Generated Text \cite{llm-detect-ai-generated-text}, another, more difficult, Kaggle dataset named AI Vs Human Text \cite{ai-vs-human}, and the set of Ghostbusters paper datasets \cite{verma2024ghostbusterdetectingtextghostwritten}, whose names will be hereafter preceded by ``GB'' to identify them easily. 

\subsubsection{Novel Datasets with Stronger Target Models}
One of the key limitations of the datasets in the literature is that they are generated using rather old language models, which are not commonly in use today; some have even been deprecated. This makes many of the publicly available datasets somewhat unrepresentative for modern LLM detection applications. Therefore, in addition to our existing suite of datasets, we also generated several novel datasets using the GPT4o Mini \cite{openai2024gpt4ocard} and Deepseek V3 models \cite{deepseekai2025deepseekv3technicalreport}. These models are regarded as highly capable by the research community, placing near or at the top in a variety of benchmarks, and importantly, the free tiers for both GPT4o Mini and Deepseek are quite generous for users and allow anyone to access them easily. Many of the datasets that we used did not provide the prompts given to the language models to generate the data; however, the essays and creative writing portions of the Ghostbusters dataset did contain the prompts used in their creation. Using these prompts, we generated the GB Essays GPT4o Mini, GB Creative Writing Deepseek, GB Creative Writing GPT4o Mini, and GB Essays Deepseek datasets. 


\subsubsection{English as a Second Language Text}
To evaluate our model's robustness to ESL (English as a Second Language) text, which is a known failure mode for AI-detection schemes \cite{liang2023gptdetectorsbiasednonnative}, we introduce the ESL GPT4o Mini dataset. The human responses for this dataset are adapted from an existing corpus of ESL student essays \cite{feedback-prize-english-language-learning}. We then tasked GPT4o Mini to rewrite the essays to improve clarity and structure. Rewriting is a difficult environment for AI detection models since entire sentences and ideas are often re-used from the human written text, possibly obfuscating the statistical fingerprints. 

\subsubsection{Adversarial Perturbations}
A student asking an LLM to write their essay may submit a rephrased version of an LLM's text. We wish to evaluate each detector's robustness to such adversarial attacks, using  the ``perturbations'' datasets by \citet{verma2024ghostbusterdetectingtextghostwritten}, which contains attacks that can range from rephrasing a word or sentence to changing the ordering of paragraphs. This dataset also contains varying levels of each perturbation, which allows us to test how each detector performs with varying levels of perturbations. We additionally test each method with an adversarial ``humanizer'' system described in Section \ref{appendix:Adversarial AI Humanizers}.

\subsection{Detection Methods and Reference Models}
\label{subsec:detectors}
We evaluate our proposed Telescope Perplexity method (Eq.~\ref{eq:telescope_ppl}) against several established zero-shot baselines representing diverse approaches. These include standard Perplexity based on next-token prediction ($P(s_i|s_{1:i-1})$), the rank-based DetectLLM LRR, which utilizes log rank \cite{su2023detectllmleveraginglogrank}, Binoculars, which utilizes Cross-Perplexity between two models \cite{hans2024spottingllmsbinocularszeroshot}, and Fast-DetectGPT, which uses conditional probability curvature \cite{bao2024fastdetectgpt}. To assess the generality and robustness of detection methods with respect to the underlying inference engine, we employed a diverse set of 12 reference models varying in size, architecture, and origin:
Gemma 2 2B, Gemma 2 9B, Llama 3.1 8B, Falcon 7B, SmolLM 135M, SmolLM 360M, SmolLM 1.7B, SmolLM2 135M, SmolLM2 360M, SmolLM2 1.7B, GPT-J 6B, GPT-Neo 2.7B. When a single reference model is needed, we use the instruct variant, and in Binoculars, which requires two reference models, we use the instruct and pretrained variant. 

\subsection{Evaluation Metrics and Procedure}
\label{subsec:metrics}

We primarily evaluate detection performance using the Area Under the Receiver Operating Curve (AUROC) score. We also report the maximum F1-Score achievable by finding the optimal threshold on the test set itself. To assess practical threshold robustness across domains, we employ a Transferability Test: for each target dataset, an optimal threshold is determined via logistic regression trained on scores from all \textit{other} datasets, and the resulting F1-Score on the held out target dataset is reported. Unless otherwise stated, reported AUROC and Transferability F1-Scores are averaged across all 12 reference models. We do not include true positive rate at very low false positive rates due to the fact that the confidence intervals are often inconclusive from these metrics when using a small false positive rate value like 0.05\%. See Appendix Section \ref{sec:unused_performance_metrics} for more information. Larger fixed false positive rate values such as 5\% offer more substantial confidence intervals, but are limited in how useful they actually are, since LLM content detectors that flag content as LLM generated when they are not 5\% of the time are generally not very useful. For this reason we include all of the experiment results with true positive rate at 5\% false positive rate in Appendix Section \ref{sec:full_results}.

\section{Experimental Results}
\label{sec:empirical_validation}

We now present the empirical results evaluating Telescope Perplexity's effectiveness as a zero-shot detector, its robustness under various conditions, and its practical limitations.

\subsection{Detection Performance}
\label{subsec:core_efficacy}
First, we assess the overall detection performance of Telescope Perplexity compared to baselines across our suite of evaluation sets. Table~\ref{tab:main_auroc_results} summarizes the average AUROC scores across all reference models for each dataset. Full results per reference model are available in Appendix Section~\ref{sec:full_results}.

\begin{table*}[t]
\centering
\caption{Detection performance (Average AUROC across 12 reference models) of Telescope Perplexity, Binoculars, Perplexity, DetectLLM LRR, and Fast-DetectGPT across diverse datasets. Bold indicates the best performance per dataset. Since we test on so many different variations of SmolLM, our averaging will be inherently biased and overvalue performance on the SmolLM architecture. The headline numbers in this table also hide variation that matters operationally. On Detect LLM Text, Telescope Perplexity attains $0.99219$ AUROC against $0.89307$ for standard Perplexity, which corresponds to error rates of $6.8\times 10^{-3}$ and $1.0\times 10^{-1}$ respectively, a roughly $15\times$ reduction. On ESL GPT4o Mini, the gap from $0.9999$ to $0.82523$ corresponds to an error-rate reduction of more than three orders of magnitude. Averaging across reference models also obscures the largest per-model gains. Readers are strongly encouraged to view our full results with the performance of each reference model on each dataset with each detection technique in Appendix Section \ref{sec:full_results}.}
\label{tab:main_auroc_results}
\resizebox{0.8\linewidth}{!}{
\begin{tabular}{l|ccccc}

\toprule
    \multirow{2}{*}{\textbf{Dataset}} & \multicolumn{5}{c}{AUROC} \\ 
    & Telescope (ours) & Binoculars & Perplexity & DetectLLM & Fast-DetectGPT \\
    \midrule
    

Detect LLM Text& \textbf{0.99219}& 0.76588& 0.89307& 0.92981& 0.70085 \\
AI vs Human& \textbf{0.95143}& 0.86297& 0.90743& 0.90316& 0.75608 \\
HC3& 0.99155& 0.99441& \textbf{0.99471}& 0.98436& 0.95584 \\
HC3 Plus& \textbf{0.98451}& 0.88510& 0.90999& 0.87758& 0.83575 \\
ESL GPT4o Mini& \textbf{0.99983}& 0.79637& 0.82523& 0.69051& 0.60603 \\
GB Essay ChatGPT& 0.98628& 0.88434& \textbf{0.99810}& 0.99730& 0.55624 \\
GB News ChatGPT& 0.90480& 0.98773& 0.98817& \textbf{0.99050}& 0.91940 \\
GB Creative ChatGPT& \textbf{0.99397}& 0.91846& 0.94990& 0.91852& 0.52336 \\
GB Essay GPT4o& 0.98136& 0.85505& \textbf{0.99365}& 0.99163& 0.51477 \\
GB Creative GPT4o& \textbf{0.99271}& 0.91276& 0.92303& 0.87374& 0.63813 \\
GB News Claude& 0.88038& \textbf{0.89263}& 0.87211& 0.86317& 0.77787 \\
GB Creative Claude& \textbf{0.96604}& 0.82929& 0.89304& 0.87449& 0.60276 \\
GB Essay Claude& 0.94223& 0.77288& 0.94310& \textbf{0.95988}& 0.61633 \\
GB Essay Deepseek V3& 0.98484& 0.99225& \textbf{0.99881}& 0.99680& 0.82763 \\
GB Creative Deepseek V3& 0.98199& \textbf{0.99569}& 0.98852& 0.96391& 0.90439 \\
    \bottomrule
\end{tabular}
}
\normalsize
\end{table*}

The results demonstrate that Telescope Perplexity consistently achieves high AUROC scores, often outperforming the current state-of-the-art method, Binoculars, as well as other strong baselines like Perplexity and DetectLLM LRR, when averaged across reference models and on each reference model. Notably, Telescope Perplexity shows exceptional performance on our novel evaluation sets featuring contemporary models like GPT4o Mini and Deepseek-V3 (e.g., GB Essay GPT4o Mini, ESL GPT4o Mini). This validates that probing the hypothesized ``Vestigial Heuristic'' via Telescope Perplexity provides a powerful and broadly effective signal for zero-shot LLM detection, even against modern target models.

\begin{figure}[h]
   \begin{minipage}{0.52\textwidth}
     \centering
     \includegraphics[width=.90\linewidth]{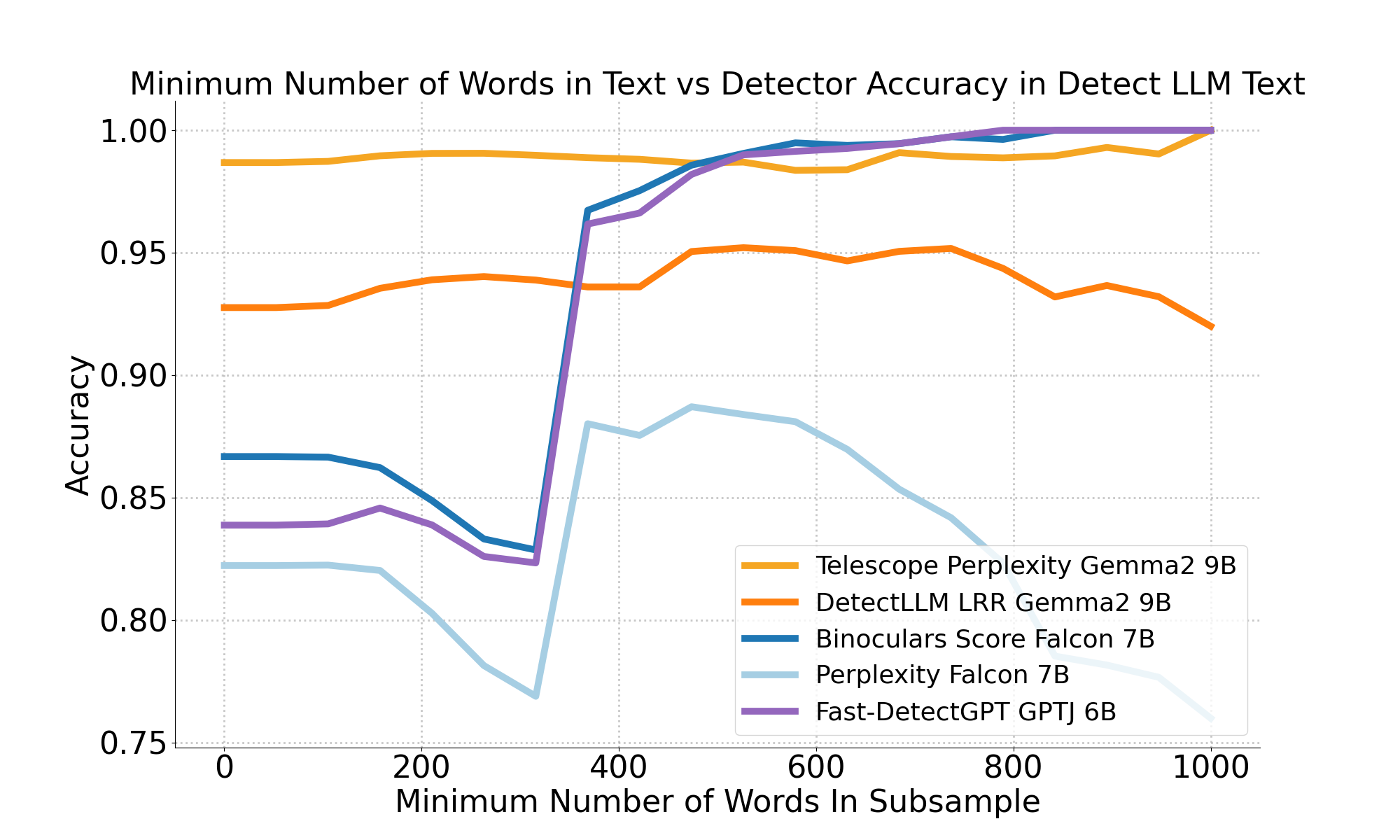}
   \end{minipage}\hfill
   \begin{minipage}{0.52\textwidth}
     \centering
     \vspace{0.3cm}
     \includegraphics[width=.90\linewidth]{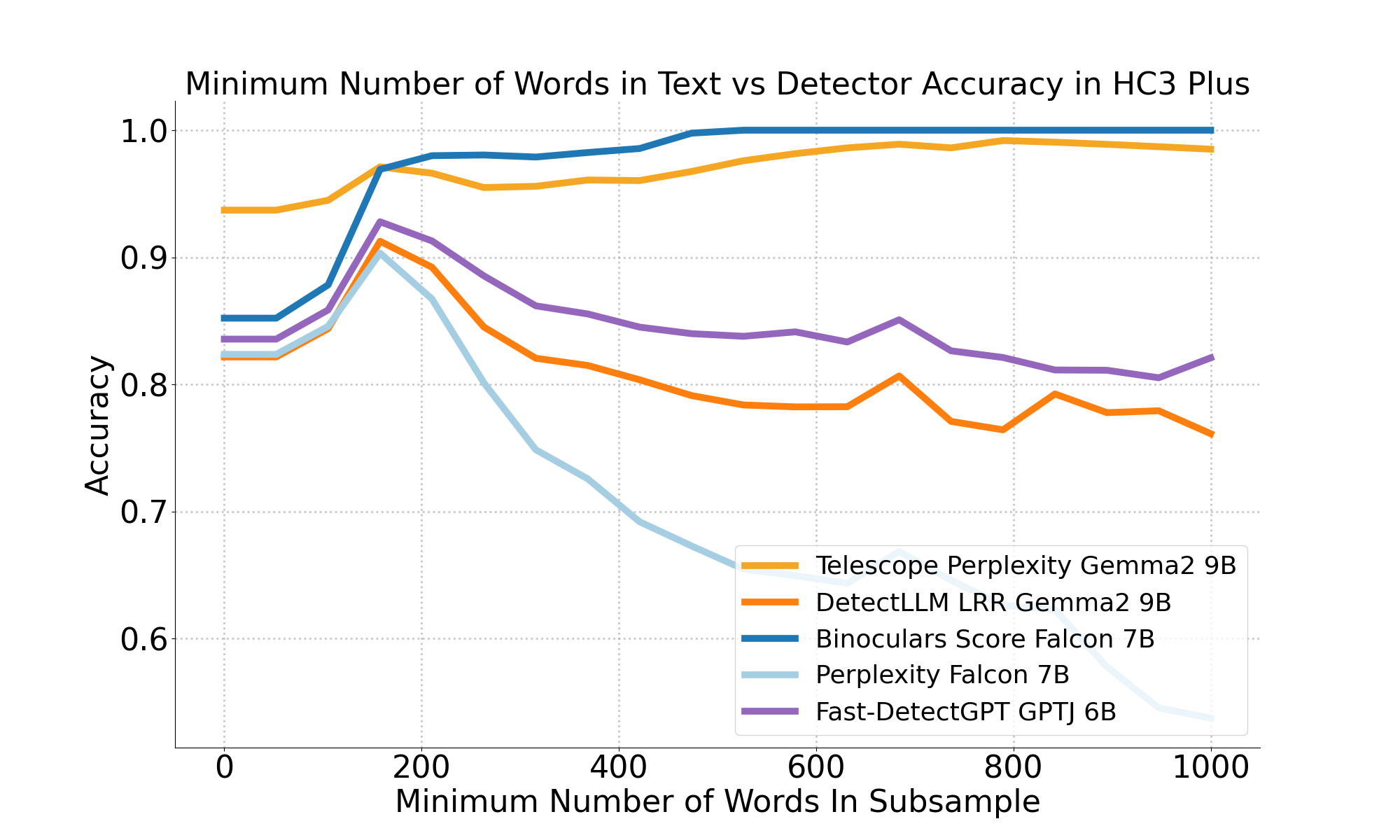}
   \end{minipage}
   \caption{Impact of minimum text length on the AUROC performance of several detectors on the Detect LLM Text dataset (top) and the HC3 Plus dataset (bottom).}
   \label{fig:text_length_impact}
\end{figure}

\begin{figure}[h]
   \begin{minipage}{0.52\textwidth}
     \centering
     \includegraphics[width=.975\linewidth]{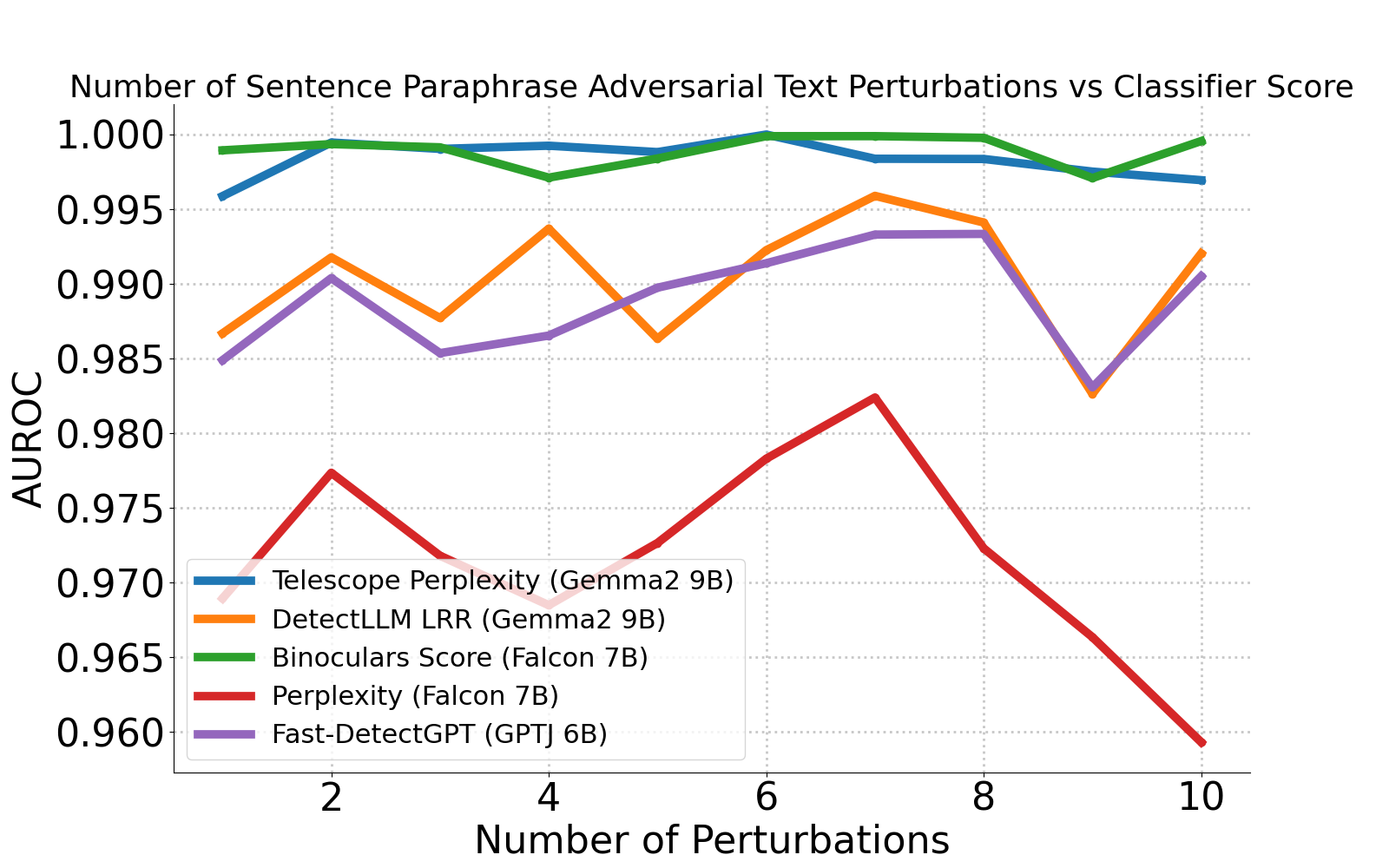}
   \end{minipage}\hfill
   \begin{minipage}{0.52\textwidth}
     \centering
     \vspace{0.3745cm}
     \includegraphics[width=.975\linewidth]{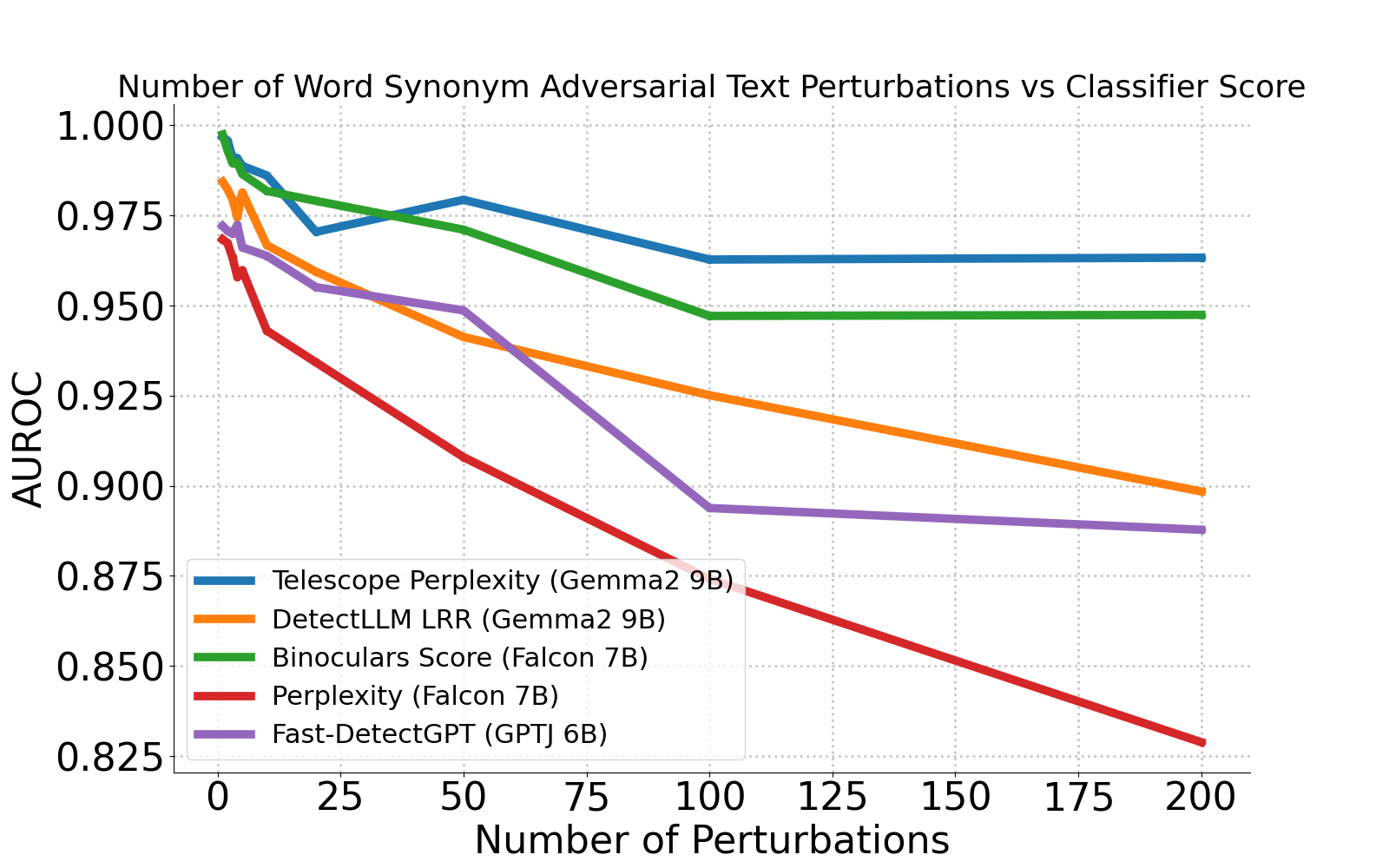}
   \end{minipage}
   \caption{Impact of paraphrasing sentences (top) and swapping words with synonyms (bottom) on detector AUROC performance on the Ghostbusters dataset.}
   \label{fig:perturbation_analysis}
\end{figure}

\subsection{Robustness}
\label{subsec:robustness}

Figure~\ref{fig:text_length_impact} demonstrates the impact of minimum text length on detector performance. While longer texts generally provide more evidence, Telescope Perplexity maintains strong performance even on relatively shorter texts (e.g. $<$ 100 or $<$ 200 words), suggesting the underlying local bias aggregates reliably. Its performance degrades less sharply with decreasing length compared to some baselines in certain datasets. For more examples, see Appendix Section~\ref{sec:additional_ablations}.

To assess resilience to simple obfuscation, we evaluated performance on texts where words were increasingly replaced by synonyms and where sentences were increasingly paraphrased. Figure~\ref{fig:perturbation_analysis} illustrates that while performance degrades for all detectors as perturbations increase, Telescope Perplexity maintains relatively high AUROC compared to baselines, indicating the local bias it measures is not solely dependent on exact lexical choice. Further evaluation on dedicated perturbation datasets by \citet{verma2024ghostbusterdetectingtextghostwritten} confirms Telescope Perplexity's robustness against various character, word, and paragraph-level modifications, which can be found in Appendix Section ~\ref{sec:additional_ablations}.
Our tests of sophisticated ``AI Humanizer'' attacks show only small degradations in performance (see Appendix Section \ref{appendix:Adversarial AI Humanizers}).

These results suggest the ``Vestigial Heuristic'' signature captured by Telescope Perplexity is reasonably robust to variations in text length and common perturbation strategies.

\begin{table*}[t]
\small
\caption{Transferability of Telescope Perplexity, Binoculars, Perplexity, DetectLLM LRR, and Fast-DetectGPT when tuned on every other dataset and tested on a specific dataset. We report the F1-Score of an algorithm's performance on the test dataset averaged across all of the reference models tested. Similarly to Table \ref{tab:main_auroc_results}, we also strongly recommend that readers view our full results in Appendix Section \ref{sec:full_results}.}
\centering
\label{tab:transferability_f1}
\resizebox{0.8\linewidth}{!}{%
\begin{tabular}{l|cccccc}
\toprule
    \multirow{2}{*}{\textbf{Test Dataset}} & \multicolumn{6}{c}{F1 Score} \\ 
    &Telescope (ours) & Binoculars & Perplexity & DetectLLM & Fast-DetectGPT\\
    \midrule


        GB Essay ChatGPT& \textbf{0.94317}& 0.84594& 0.93620& 0.94249& 0.76713& \\
        GB News ChatGPT& 0.86115& \textbf{0.93763}& 0.81355& 0.87625& 0.63212& \\
        GB Creative ChatGPT& \textbf{0.96255}& 0.85858& 0.89799& 0.85935& 0.72314& \\
        GB Essay GPT4o& \textbf{0.93922}& 0.82922& 0.93493& 0.93341& 0.73008& \\
        GB Creative GPT4o& \textbf{0.96269}& 0.85953& 0.73004& 0.74011& 0.72738& \\
        GB News Claude& 0.69429& \textbf{0.79833}& 0.76009& 0.74903& 0.60027& \\
        GB Creative Claude& \textbf{0.89028}& 0.76001& 0.69536& 0.75085& 0.62287& \\
        GB Essay Claude& 0.87984& 0.74559& 0.86685& \textbf{0.88829}& 0.69177& \\
        GB Essay Deepseek V3& 0.94006& \textbf{0.94689}& 0.93718& 0.94351& 0.71785& \\
        GB Creative Deepseek V3& 0.93721& 0.90241& \textbf{0.95466}& 0.90851& 0.69879& \\
        
        \bottomrule
\end{tabular}
}
\normalsize
\end{table*}

\subsection{Domain Sensitivity}
\label{subsec:boundaries}

While generally effective, Telescope Perplexity's performance varies across domains. As seen in Table~\ref{tab:main_auroc_results}, Telescope Perplexity achieves near-perfect scores when detecting ESL rewriting with an average 0.99983 AUROC across every reference model tested. We hypothesize this is a result of the ESL writing triggering the ``Vestigial Heuristic'' even more weakly than standard human text. \linebreak \linebreak

Conversely, Telescope Perplexity's performance, while still often strong, is comparatively lower on some news writing datasets (e.g., Ghostbusters News Claude with an average AUROC of 0.88038 across every reference model tested). Additionally, Telescope Perplexity consistently misclassified AI-generated poetry as human written, as seen in Appendix Section~\ref{sec:misclassification}. This suggests that genres governed by rigid lexical constraints or highly stylized registers (such as poetry or specific journalistic styles) might obscure the typical ``Vestigial Heuristic'' signature, which defines limitations for the current probe. Even though Telescope Perplexity seems to struggle with news writing, it also seems to perform better on essay writing than creative writing. Since essay writing is generally more formulaic than creative writing, this suggests that there isn't a clear direct correlation between detection performance and formulaicity.

\subsection{Practical Considerations and Recommendations for Deployment}
\label{subsec:deployment}
\label{subsec:practicalities}

Finally, we examine practical aspects relevant to deployment. Table~\ref{tab:transferability_f1} presents the results of our Transferability Test, evaluating how well a threshold tuned on N-1 datasets generalizes to the held-out Nth dataset.

While Telescope Perplexity often performs well even in this challenging scenario, its F1 score sometimes drops compared to its potential maximum F1 (achieved when tuning on the test set itself). This indicates that while the ``Vestigial Heuristic'' signature is generally present, its baseline level (and thus the optimal threshold) can vary slightly depending on the domain, style, or target model.

Like most other zero-shot detectors we tested, Telescope Perplexity's raw output scores are poorly calibrated and should not be interpreted as true probabilities. They serve as effective discriminative scores for classification but exhibit overconfidence as shown in Appendix Section~\ref{sec:calibration}.

Our experiments suggest several concrete recommendations for anyone
deploying Telescope Perplexity in practice.

\paragraph{Reference-model selection.}
Per-reference-model results in Appendix~\ref{sec:full_results} show
that no single reference model is uniformly best, and the spread
across reference models on a fixed dataset can exceed the spread
between detectors on the same row. We therefore recommend evaluating
candidate reference models on a small held-out portion of the target
domain before committing to one. When inference budget allows, taking
the mean or median Telescope Perplexity over a small ensemble of
reference models from different families (e.g.\ one Gemma, one Llama,
one SmolLM) trades a constant-factor compute cost for substantially
reduced variance across domains.

\paragraph{Threshold calibration.}
The Transferability results in Table~\ref{tab:transferability_f1}
indicate that thresholds tuned on unrelated datasets degrade,
sometimes substantially, when transferred to a new domain. We
therefore recommend the following calibration procedure for any new
deployment:
\begin{enumerate}[leftmargin=*,topsep=2pt,itemsep=2pt]
    \item Collect a small labeled validation set ($\sim 200$--$500$
    examples, balanced between LLM generated and human written) drawn
    from the same distribution the detector will see in deployment.
    \item Score every example with the chosen reference model and
    detector.
    \item Choose $\tau$ to maximize F1 in this validation set, or, if
    a fixed false positive budget is required (e.g., academic-misconduct
    settings where false positives carry severe cost), choose $\tau$ as
    the smallest threshold whose validation set false positive rate is
    below the operational budget.
    \item Re-calibrate periodically as the target-model and writing
    distributions drift.
\end{enumerate}
For settings where labeled validation data is genuinely unavailable,
the transferability scores in Table~\ref{tab:transferability_f1}
give a reasonable lower bound on what to expect from an
externally-tuned threshold.

\subsection{Failure Modes and Misclassification Analysis}
\label{subsec:failure_modes}

To complement the aggregate metrics above, we examine where Telescope
Perplexity errs and how confidently. Across all reference models the
expected distance from the decision threshold for a misclassified
example is approximately one standard deviation of the score
distribution, with human written texts misclassified as LLM generated
typically sitting \emph{closer} to the threshold (i.e.\ low-confidence
errors) and LLM generated texts misclassified as human written
typically sitting \emph{further} from the threshold (high-confidence
errors). Full per-model error-distance plots are deferred to
Appendix Section \ref{sec:misclassification}; here we highlight the two
qualitative patterns that matter most for deployment.

\paragraph{Stylised and formulaic text.}
The most confident false negatives we observed were in stylized or
formulaic genres. Figure~\ref{fig:poetry_misclass} (deferred to
Appendix Section \ref{sec:misclassification} for space) shows an LLM generated pirate-themed poem in the
GB Creative ChatGPT split that Telescope Perplexity scored at $12.03$
— well below the dataset decision boundary of $13.36$. Stylized
constraints (rhyme, meter, register) appear to suppress exactly the
local repetition aversion that our score probes, in part by forcing
end-line repetition (``Ahoy, me hearties \ldots'' echoed across
stanzas) that human written verse would also exhibit. This is
consistent with the weaker performance on highly formulaic news
writing (e.g.\ GB News Claude, average AUROC $0.880$) observed in
Section~\ref{subsec:boundaries}.

\paragraph{Direction of error and operational implications.}
The asymmetry described above has a concrete deployment implication:
the most damaging error mode for an academic-integrity use case is the
high-confidence false negative (LLM generated text passed off as
human). Practitioners who care about this asymmetry should set
thresholds against a target false-negative rate on stylized genres
specifically, rather than against the data set-averaged decision
boundary used for our reported maximum F1.

A more detailed dataset-by-dataset error decomposition, including
representative misclassified samples and per-model error-distance
distributions, is provided in
Appendix Section \ref{sec:misclassification_analysis} and the surrounding subsections.

\section{Limitations}
While we have attempted to rigorously test Telescope Perplexity against other LLM text detectors, there are still a couple of limitations to our experimentation methods due to compute constraints and practical limitations.


 \textbf{1)} \textbf{Perturbation-Based Detectors:} We do not benchmark on DetectLLM NPR, DetectGPT or other perturbation based detection algorithms because of their massive computational cost. Preliminary testing showed worse performance on NPR compared to LRR, and there have been studies that show that DetectGPT heavily underperforms compared to other modern techniques \cite{li2024magemachinegeneratedtextdetection}.

\textbf{2)} \textbf{Datasets Representative of Deployment:} We attempt to address data limitations in AI detection by introducing a suite of new datasets. Although these datasets use modern LLMs as target models, they do not accurately reflect the data distribution in deployment environments,  where students may employ multiple adversarial methods to avoid detection alongside newer, more powerful language models.

\section{Conclusion}

In this work, we present the Telescope Perplexity metric, which we hypothesize probes an underlying behavior that forms early in training which discourages token repetition with a ``Vestigial Heuristic''. We demonstrate the local nature of this ``Vestigial Heuristic'' and track its development in LLM training.
Through rigorous experimentation, we empirically compare our method with baselines and show that Telescope Perplexity is effective at discriminating between LLM generated and human written text and is particularly effective when tuned to a particular domain, while being robust to the choice of reference model.

\section*{Impact Statement}
Advances in the detection of synthetic text, if adopted in high-stakes
settings (e.g., academic integrity), carry significant risks alongside their
intended benefits. The primary benefit of improved detection capabilities is
defensive, weakening the value proposition of misuses of language models such
as automated disinformation and academic dishonesty. More effective detection
is also expected to have stronger negative impacts on the authors of
misclassified texts. As more reliable techniques, like Telescope Perplexity,
become available in non-lab settings, we caution against the exclusive use of
automated single-point-of-failure systems in favor of using such systems as a
first line of defense. In addition, we note that on our ESL evaluation set Telescope Perplexity
is less susceptible to the non-native-speaker bias documented in prior work
\cite{liang2023gptdetectorsbiasednonnative}; (Section~\ref{subsec:boundaries}).
\section*{Acknowledgements}

We thank Dr. Ming Jin for his guidance throughout this work. We also gratefully acknowledge the Advanced Research Computing (ARC) at Virginia Tech and Mobius Logic for providing the computational resources that made this research possible.




\nocite{*}
\bibliography{references}
\bibliographystyle{icml2026}

\clearpage
\section{Appendices}

\subsection{Clarifications on Vocabulary} \label{Vocabulary}

Neural networks can become arbitrarily good at a task with enough data \cite{Cybenko1989-fv}, and we can apply them to classify whether or not a piece of text has been generated by LLMs; however they do have some very major drawbacks. Simply using a neural network to classify text generated by one language model may not perform well on language models it hasn't been trained on. In a world where new language models are released constantly, it is uniquely valuable to create an LLM generated text detector that will work on any language model trained in the future. In addition, data for these tasks on a wide variety of current language models in a wide variety of tasks is often difficult to find which can limit these detectors' reliability. This is why in the literature, there is a distinction between zero-shot methods and non-zero-shot methods; zero-shot methods promise to correctly classify LLM generated text without training on the specific language model that may have generated the text, while non-zero-shot methods must be trained on specific language models and may not be able to generalize to newly released language models. Since consumers will switch to whichever language model performs the best at a given point in time, zero-shot methods are generally thought to be more robust but are more difficult to design and make accurate.

\begin{figure}[!h]
    \centering

    \includegraphics[width=1\linewidth]{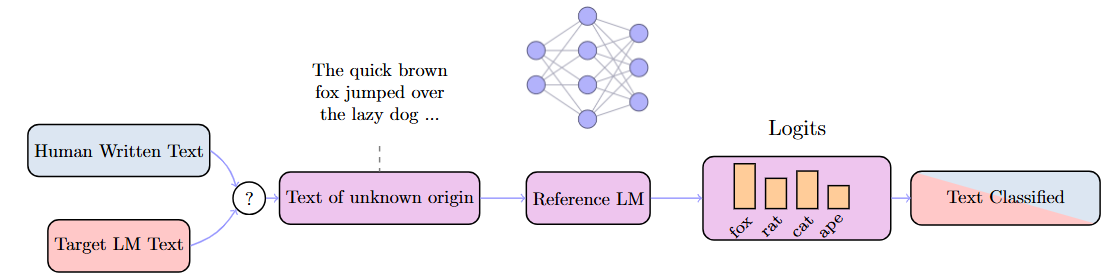}
    \caption{A graphical illustration of the relationship between the reference language model and the target language model.}
    \label{fig:Telescope_Relationship_Between_Reference_Model_And_Target_Model}
\end{figure}

Because of their desirable properties, many researchers have focused on developing and improving the performance of these zero-shot detectors. Zero-shot detection is a difficult problem since it requires directly analyzing the high dimensional space of text, so in an attempt to compress this space many of the zero-shot detectors in the literature opt to use a reference model to help detect the text generated by a given target model. A target model is simply the model that may have generated the text under question, while the reference model is a model that a specific AI generated text detector may use to produce the probabilities for the next token for every word in the sentence. For example, in the sentence ``The quick brown fox jumped over the lazy dog'', the reference language model will produce the probability for any word to come after ``The'', ``The quick'', ``The quick brown'', etc. These probabilities can then be analyzed using metrics such as the average Perplexity of each word that the model predicted. It turns out that depending on the reference model used, the logits distributions can often convey very interesting information for whether the text under question was generated by some target language model. The target model is generally unknown, since we generally don't have a good idea which language models may have been used to generate text out in the wild; however, the reference model is chosen by the algorithm designer, and different reference models may have different properties that make them better or worse at detecting AI generated content from a variety of target models. The relationship between reference models and target models is illustrated in Figure \ref{fig:Telescope_Relationship_Between_Reference_Model_And_Target_Model}. In this paper, we mainly focus on zero-shot methods and attempt to push the boundaries of the accuracy that can be achieved with them while still being able to easily generalize to any language model.

\vspace{5mm}

\subsection{Analysis of Frequency Components of the Telescope Score} 
We now consider the single token Telescope Perplexity:

\begin{equation}
    \text{Single Token Log Telescope PPL}_\mathcal{M}(s_i)  = -\log\mathcal{M}(s_i \mid s_{1:i})
\end{equation}

\begin{figure}[!h]
    \centering
    \includegraphics[width=1.0\linewidth]{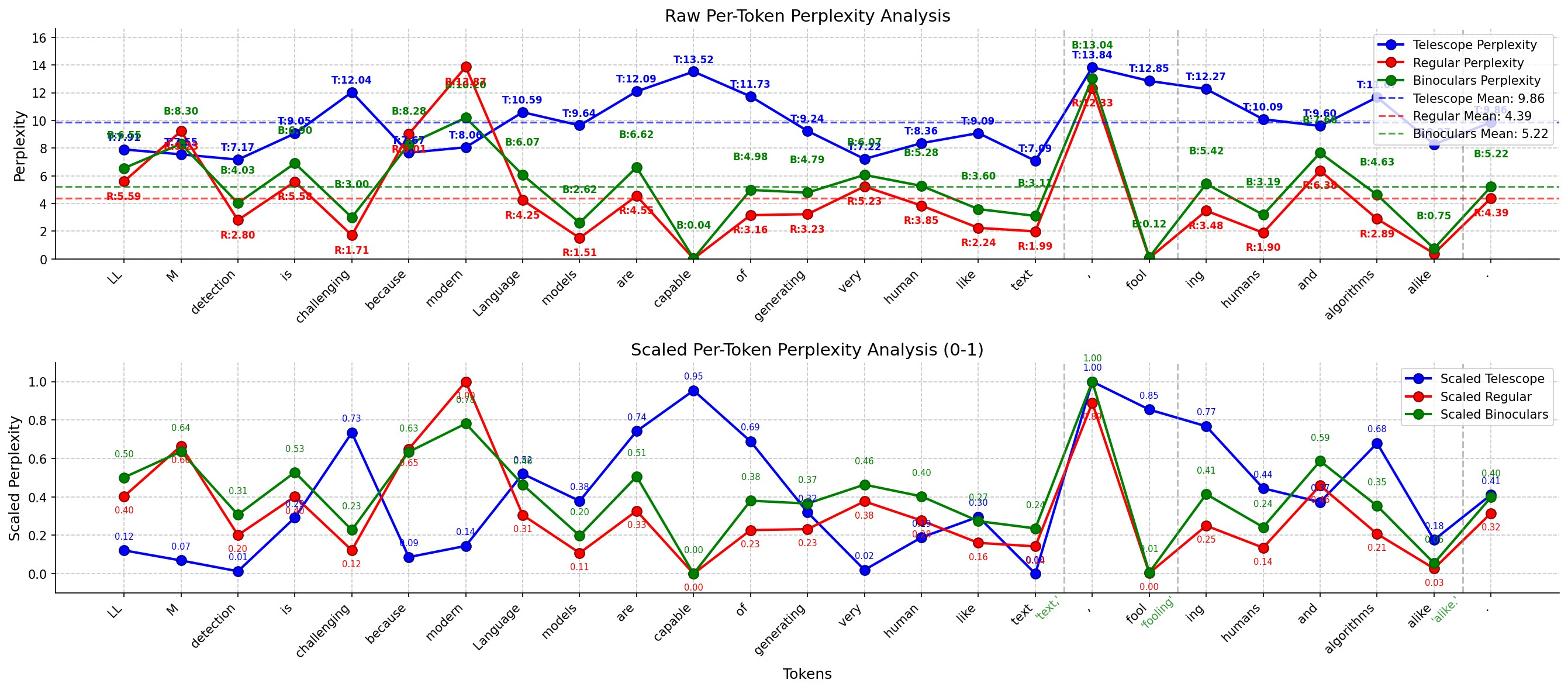}
    \caption{Per-token Telescope Perplexity, Perplexity, and Binoculars Score over the sequence}
    \label{fig:Telescope_perplexity_signal_AI}
\end{figure}

We can use this to plot our per-token Telescope Perplexity as in Figure \ref{fig:Telescope_perplexity_signal_AI}. Readers familiar with signal processing may notice that the Telescope Perplexity over the token position resembles a signal. Interrogating this similarity, we found that the only truly relevant component of this signal is its mean. Future work applying sequence models in this domain is needed to fully evaluate if only the mean of this signal is truly important; our analysis applying signal processing yielded no results.

\subsection{Training Data Separability} 
\label{appendix:Training_Data_Separability}
Here we present additional results on using SmolLM 360M to perform detection on a 10k subsample of its synthetic and human written training data.
We find that, interestingly, standard Perplexity is able to somewhat effectively separate its own training data into synthetic and human data, implying Perplexity's effectiveness in detecting LLM generated text needs further study. The Binoculars Score does not effectively delineate between AI and human training data.

\begin{table}[h]
\centering
\begin{tabular}{l | c c}
\hline
\textbf{Method} & \textbf{AUC} & \textbf{F1 Score} \\
\hline
Binoculars & 0.7692 & 0.7327 \\
\hline
Perplexity & 0.9523 & 0.9056 \\
\hline
Telescope Perplexity & 0.9956 & 0.9874 \\
\hline
\end{tabular}
\caption{F1 and AUC of SmolLM 360M on its own training data.}
\label{tab:performance_on_own_training_data}
\end{table}

\subsection{Error Independence Analysis}
\label{appendix:error_independence}

We analyzed error independence across all detection methods to identify potential ensemble opportunities. For each classifier pair, we computed Cohen's Kappa, Yule's Q-statistic, and Normalized Mutual Information (NMI) to measure how independently they make errors. The Q-statistic, calculated as $Q = \frac{n_{11} \cdot n_{00} - n_{10} \cdot n_{01}}{n_{11} \cdot n_{00} + n_{10} \cdot n_{01}}$, ranges from -1 to 1, with lower values indicating greater error independence between classifiers.

Table~\ref{tab:error_independence} shows aggregated statistics across all datasets, sorted by Q-statistic. DetectLLM using Falcon 7B exhibited the highest independence (Q = 0.017), making errors on almost completely different examples than other methods. Telescope Perplexity and standard Perplexity from the same model showed moderate dependence (Q = 0.7-0.8), while Binoculars showed lower dependence due to its Cross-Perplexity normalization.
\begin{table}[!h]
\small
\resizebox{.95\linewidth}{!}{
\begin{tabular}{l|l|ccc}
\toprule
\textbf{Detector} & \textbf{Reference Model} & \textbf{Kappa} & \textbf{Q-stat} & \textbf{MI} \\
\midrule
LRR & Falcon 7B & 0.059 & 0.081 & 0.023 \\
Fast-DetectGPT & SmolLM 1.7B & --- & 0.504 & 0.058 \\
Fast-DetectGPT & Llama3 8B & 0.339 & 0.518 & 0.057 \\
Fast-DetectGPT & SmolLM 360M & --- & 0.583 & 0.059 \\
Fast-DetectGPT & SmolLM 135M & 0.309 & 0.590 & 0.062 \\
Binoculars & SmolLM 1.7B & 0.420 & 0.605 & 0.079 \\
Fast-DetectGPT & SmolLM2 135M & 0.410 & 0.637 & 0.086 \\
Fast-DetectGPT & SmolLM2 1.7B & 0.405 & 0.639 & 0.070 \\
Fast-DetectGPT & SmolLM2 360M & 0.440 & 0.678 & 0.082 \\
Binoculars & SmolLM 360M & 0.480 & 0.683 & 0.085 \\
Binoculars & SmolLM 135M & 0.539 & 0.748 & 0.086 \\
Binoculars & SmolLM2 1.7B & 0.604 & 0.818 & 0.083 \\
Telescope & Falcon 7B & 0.640 & 0.830 & 0.054 \\
Fast-DetectGPT & Gemma2 2B & 0.620 & 0.832 & 0.105 \\
Binoculars & Llama3 8B & 0.626 & 0.835 & 0.082 \\
LRR & SmolLM2 135M & 0.630 & 0.836 & 0.129 \\
Telescope & SmolLM2 135M & 0.677 & 0.850 & 0.075 \\
LRR & SmolLM2 360M & 0.671 & 0.853 & 0.149 \\
Fast-DetectGPT & Gemma2 9B & 0.647 & 0.855 & 0.105 \\
LRR & Llama3 8B & 0.684 & 0.859 & 0.138 \\
LRR & SmolLM 135M & 0.671 & 0.860 & 0.126 \\
Telescope & SmolLM 135M & 0.686 & 0.860 & 0.078 \\
Telescope & SmolLM2 360M & 0.693 & 0.861 & 0.090 \\
Telescope & Gemma2 2B & 0.681 & 0.861 & 0.055 \\
LRR & SmolLM2 1.7B & 0.684 & 0.861 & 0.154 \\
Fast-DetectGPT & Falcon 7B & 0.667 & 0.862 & 0.086 \\
LRR & Gemma2 2B & 0.686 & 0.863 & 0.130 \\
Binoculars & SmolLM2 360M & 0.669 & 0.864 & 0.097 \\
Telescope & SmolLM2 1.7B & 0.704 & 0.868 & 0.082 \\
Telescope & SmolLM 360M & 0.704 & 0.870 & 0.086 \\
Binoculars & SmolLM2 135M & 0.683 & 0.871 & 0.106 \\
Telescope & SmolLM 1.7B & 0.705 & 0.871 & 0.082 \\
Perplexity & SmolLM2 135M & 0.695 & 0.872 & 0.183 \\
Perplexity & Gemma2 9B & 0.705 & 0.872 & 0.160 \\
Perplexity & SmolLM 135M & 0.702 & 0.873 & 0.181 \\
LRR & SmolLM 360M & 0.697 & 0.874 & 0.138 \\
Telescope & Llama3 8B & 0.707 & 0.874 & 0.080 \\
Telescope & Gemma2 9B & 0.711 & 0.876 & 0.057 \\
LRR & SmolLM 1.7B & 0.705 & 0.878 & 0.130 \\
LRR & Gemma2 9B & 0.705 & 0.879 & 0.111 \\
Perplexity & Falcon 7B & 0.709 & 0.880 & 0.181 \\
Perplexity & SmolLM 360M & 0.715 & 0.880 & 0.192 \\
Perplexity & SmolLM 1.7B & 0.718 & 0.881 & 0.196 \\
Binoculars & Gemma2 2B & 0.705 & 0.881 & 0.107 \\
Perplexity & SmolLM2 360M & 0.718 & 0.883 & 0.197 \\
Binoculars & Falcon 7B & 0.715 & 0.884 & 0.090 \\
Binoculars & Gemma2 9B & 0.707 & 0.885 & 0.097 \\
Perplexity & SmolLM2 1.7B & 0.726 & 0.887 & 0.192 \\
\end{tabular}}
\caption{Error independence statistics (average Kappa, average Q-statistic, and average Mutual Information (MI)) aggregated across all datasets. Lower values for the Kappa, Q-statistic, and Mutual Information (MI) indicate greater error independence.}
\label{tab:error_independence}
\end{table}

These results suggest promising ensemble combinations: DetectLLM (especially with Falcon 7B) paired with any Perplexity-based method, different model sizes within the same architecture, and Telescope Perplexity combined with Binoculars. The low mutual information values (mostly $<$ 0.2) indicate substantial opportunities for ensemble improvements through complementary error patterns.
\subsection{The Calibration of Zero-Shot Detection Methods} \label{sec:calibration}
\begin{figure}[!htbp]
    \centering
    \begin{minipage}{0.48\textwidth}
        \centering
        \includegraphics[width=\linewidth]{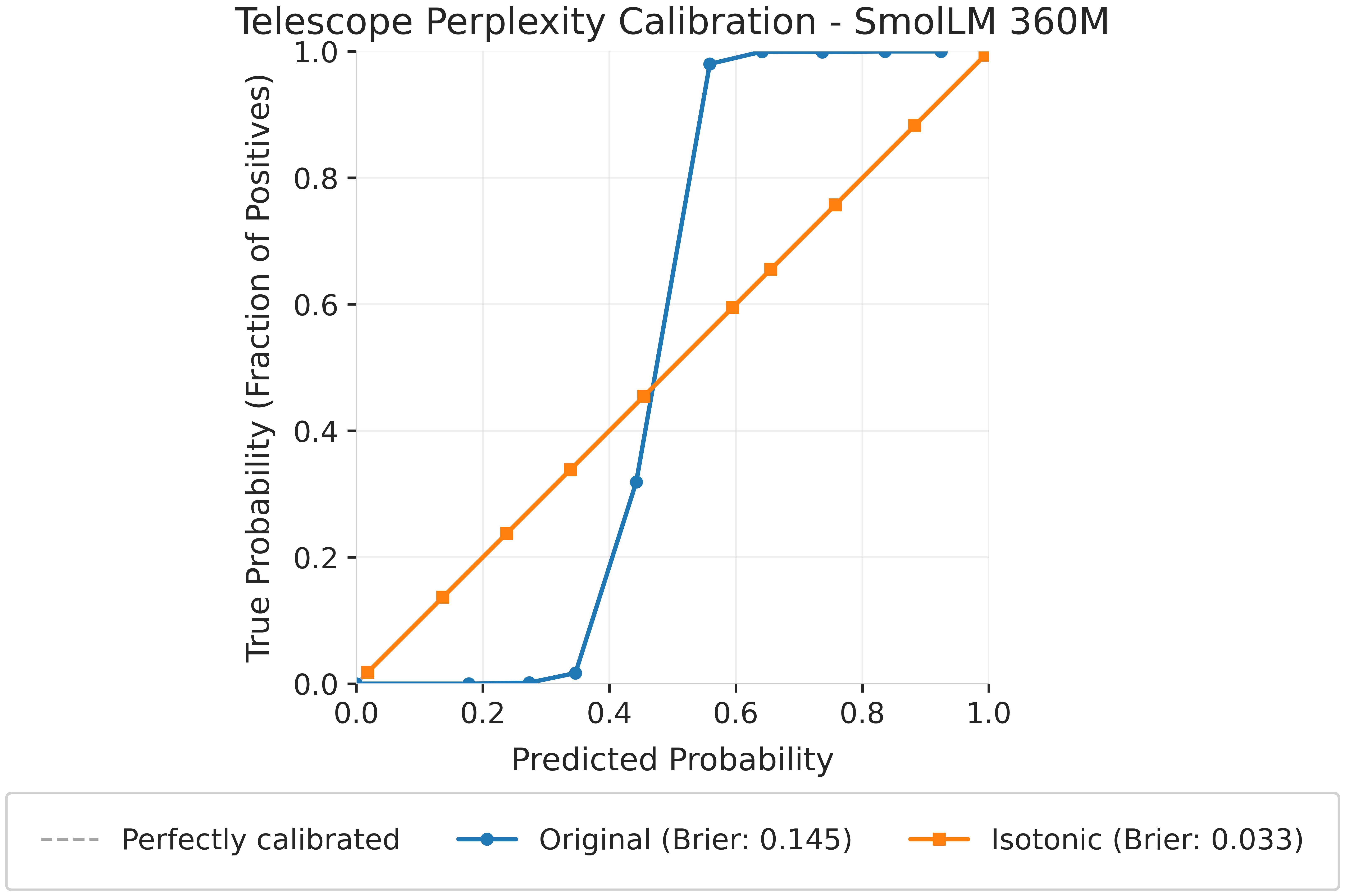}
    \end{minipage}%
    \hfill
    \begin{minipage}{0.48\textwidth}
        \centering
        \includegraphics[width=\linewidth]{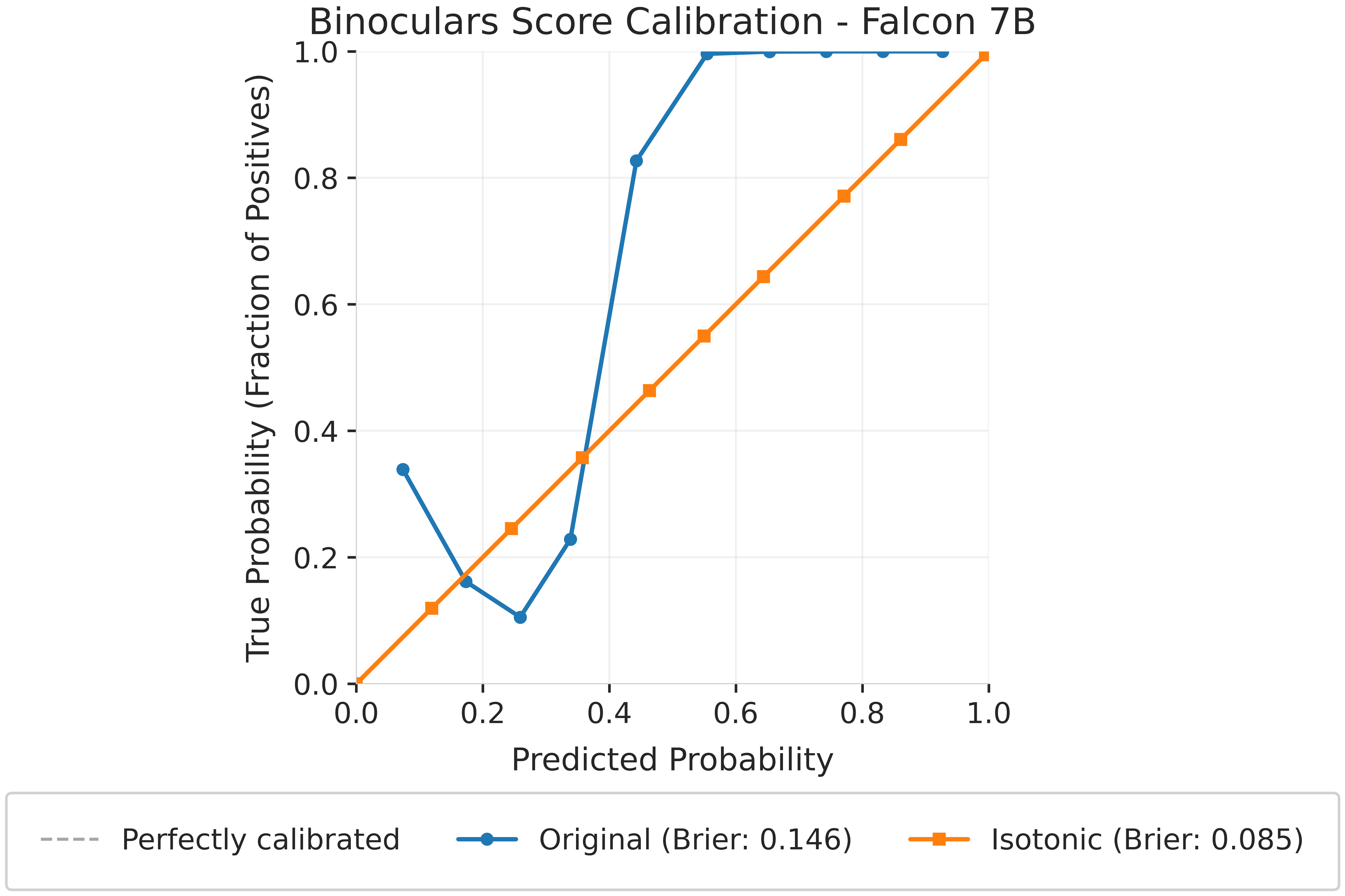}
    \end{minipage}
    \caption{Calibration comparison between SmolLM 360M Telescope Perplexity (top) and Falcon Binoculars method (bottom)}
    \label{fig:telescope_calibration}
\end{figure}

\begin{figure}[!htbp]
    \centering
    \includegraphics[width=1\linewidth]{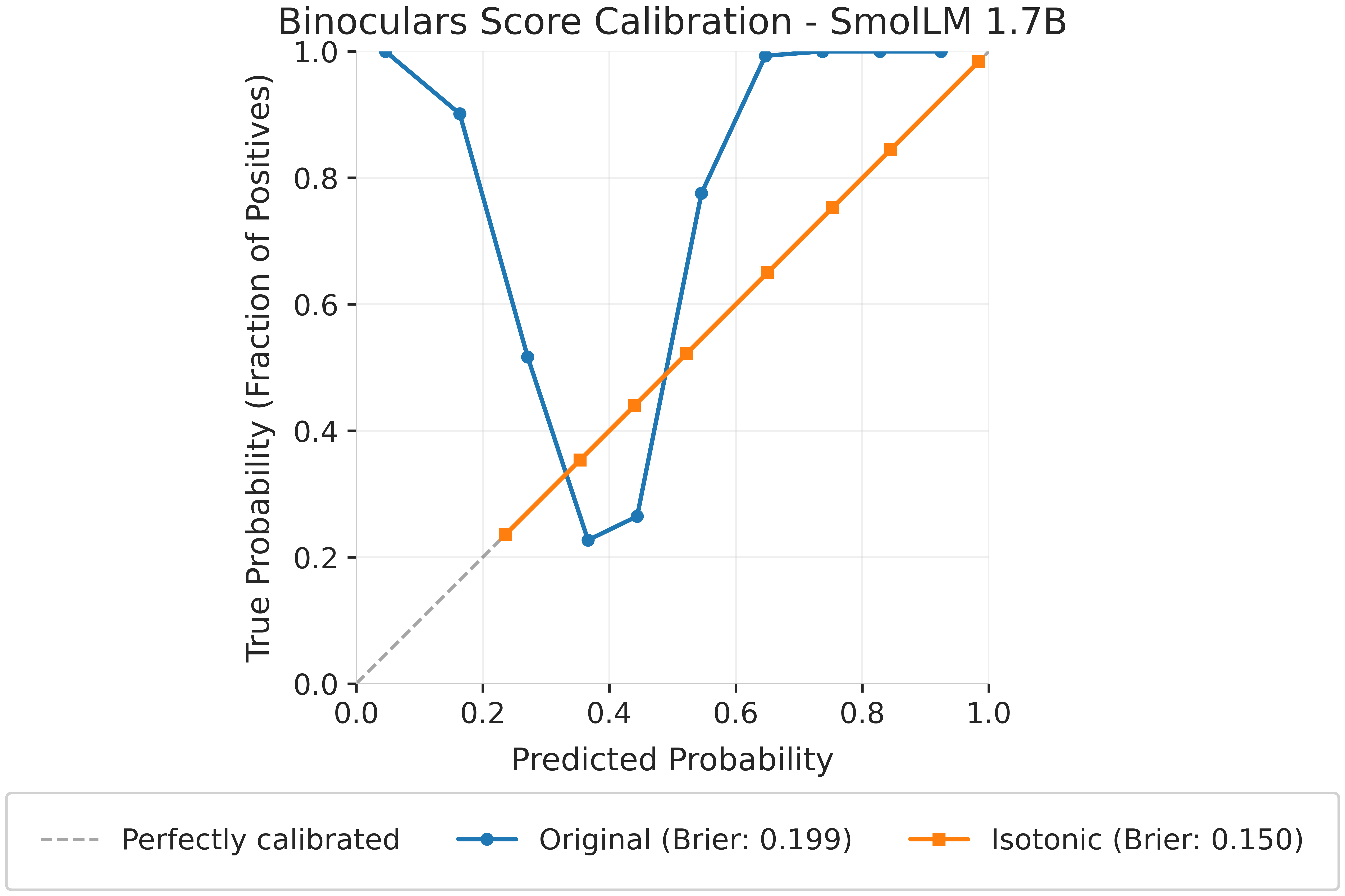}
    \caption{An extremely poorly calibrated Binoculars example using SmolLM 1.7B as a reference model.}
    \label{fig:binoculars_poorly_calibrated}
\end{figure}

Both Binoculars and Telescope Perplexity are natively very poorly calibrated, meaning the scores produced by these models do not accurately reflect the real probabilities of a given text being LLM generated. The calibration behavior of both detectors is shown in Figure \ref{fig:telescope_calibration} with one particularly poorly calibrated example shown in Figure \ref{fig:binoculars_poorly_calibrated}.

\subsection{Datasets} \label{sec:datasets}

For each dataset, we filter out samples that are less than 100 words long since when text is shorter, it may be statistically impossible to determine whether the text is written by a large language model. Shorter pieces of text have fewer different ways to write them and because of this, the way that a human is likely to write the text may be very similar to the way that an LLM would have written the text. Intuitively, this is similar to trying to make a decision based off of too little evidence. Including these short samples introduces unwanted variance to our measurements since the best the models can do some of the time is simply guess. To the best of our knowledge, there isn't any research specifically targeting how long the input to these detectors should be to minimize the data-points that are statistically impossible or difficult to classify for a theoretically perfect classifier, so we choose to filter out all text samples with fewer than 100 words from our datasets. A popular commercial detector called GPTZero also recommends that text be longer than 100 words for accurate classification. Other than following the lead of GPTZero, this decision is fairly arbitrary, and we decided on it at the beginning of testing and have not changed it since. In addition, to save on computation cost, we also filter out any samples that are larger than 5000 words, since some of the LLM generated samples were incredibly long and would often induce out of memory errors. To reduce out of memory errors on our limited compute resources, we decided to simply take out the samples which were long enough to cause errors. For extremely long datasets with hundreds of thousands of samples, we only use the first 10,000 samples to save on compute.

\begin{figure}[h]
\centering
\includegraphics[width=0.475\textwidth]{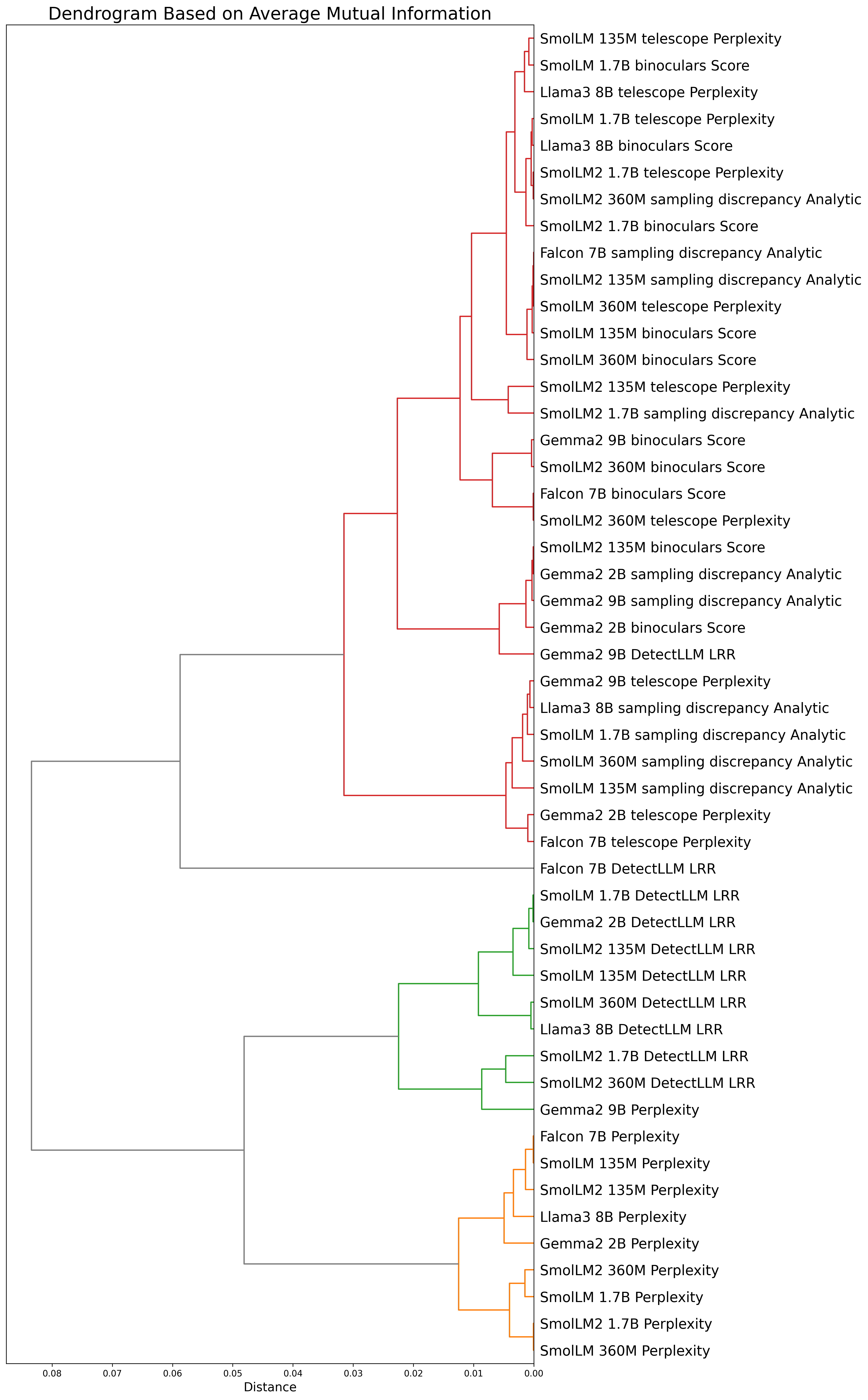}
\caption{MI for aggregated datasets with different colors showing model groups that are error independent with other groups.}
\label{fig:error_independence_heatmaps}
\end{figure}

Figure~\ref{fig:error_independence_heatmaps} visualizes pairwise error independence, using a dendrogram with less error independent models being closer in distance.

\subsection{Example of Misclassification on Poetry/ Stylized Text} \label{sec:misclassification}

In this section, we show an example of an extremely misclassified AI generated text sample in the GB Creative Writing ChatGPT dataset that Telescope Perplexity very confidently classified as human written. This text sample is stylized to be pirate-like, which may indicate that this is a way to avoid detection (although this may be a rather impractical way to avoid detection in many circumstances).

\begin{figure}[!h]
\fbox{
\parbox{8cm}{
\footnotesize
\textbf{Text:} \\
Ahoy, me hearties, gather round, 
I'll spin for ye a tale profound, 
Of an island in the distant deep, 
Where dragons fly and secrets keep. 
'Twas a land surrounded by azure seas, 
Where the dragon isles danced with the breeze, 
Guarded fiercely by three women bold, 
Their stories, legends yet untold. 
The first was Rosie, with flaming hair, 
A tempest rage, none could compare, 
With fiery spirit, she stood her ground, 
To protect the island, her dragons, renowned. 
Then there was Lily, with eyes so keen, 
A huntress born, her aim, unseen, 
With bow in hand and arrow true, 
No enemy could escape her view. 
Lastly was Bella, the wise and serene, 
Her words like gold, her heart a queen, 
She bore the knowledge of ancient roam, 
And in her presence, wisdom will bloom. 
Together they sailed amidst the tide, 
With dragon wings, they took their stride, 
The island shrouded in mystery rare, 
Its secrets guarded with utmost care. 
Dragons gleamed in the sun's warm glow, 
With scales that shimmered, a mighty show, 
In harmony, they danced and soared, 
Their fiery breath, their mighty hoard. 
But one day came a pirate crew, 
With greed and darkness to pursue, 
To snatch the dragons, to rule with might, 
And claim the island, lost from sight. 
But Rosie, Lily, and Bella stood, 
United as one, they knew they should, 
With swords unsheathed and hearts aflame, 
They fought for the dragons they held no blame. 
As the battle raged with thunderous sound, 
The pirates fled, their treasure unfound, 
The island stood proud, the dragons remained, 
A testament of the women unchained. 
So hear me shanty, a tale of courage bold, 
Of an island guarded against pirates untold, 
Through legends sung, let their bravery live, 
Three fierce women, their souls will thrive. 
Ahoy, me hearties, let our voices ring, 
Of the island where dragons took wing, 
And when the sea sings this tale divine, 
May their bravery echo throughout all time.
\noindent\rule{8cm}{0.4pt}
\textbf{Telescope Perplexity:} 12.03 \\
\noindent\rule{8cm}{0.4pt}
\textbf{Label:} LLM Generated \\
\noindent\rule{8cm}{0.4pt}
\textbf{Decision Boundary for This Dataset:} 13.36
\normalsize
}}
\caption{Example of Misclassification}
\label{fig:poetry_misclass}
\end{figure}

\subsection{Misclassifications Analysis}
\label{sec:misclassification_analysis}
\begin{figure*}[t]
    \centering
    \begin{subfigure}[b]{0.48\textwidth}
        \centering
        \includegraphics[width=\textwidth]{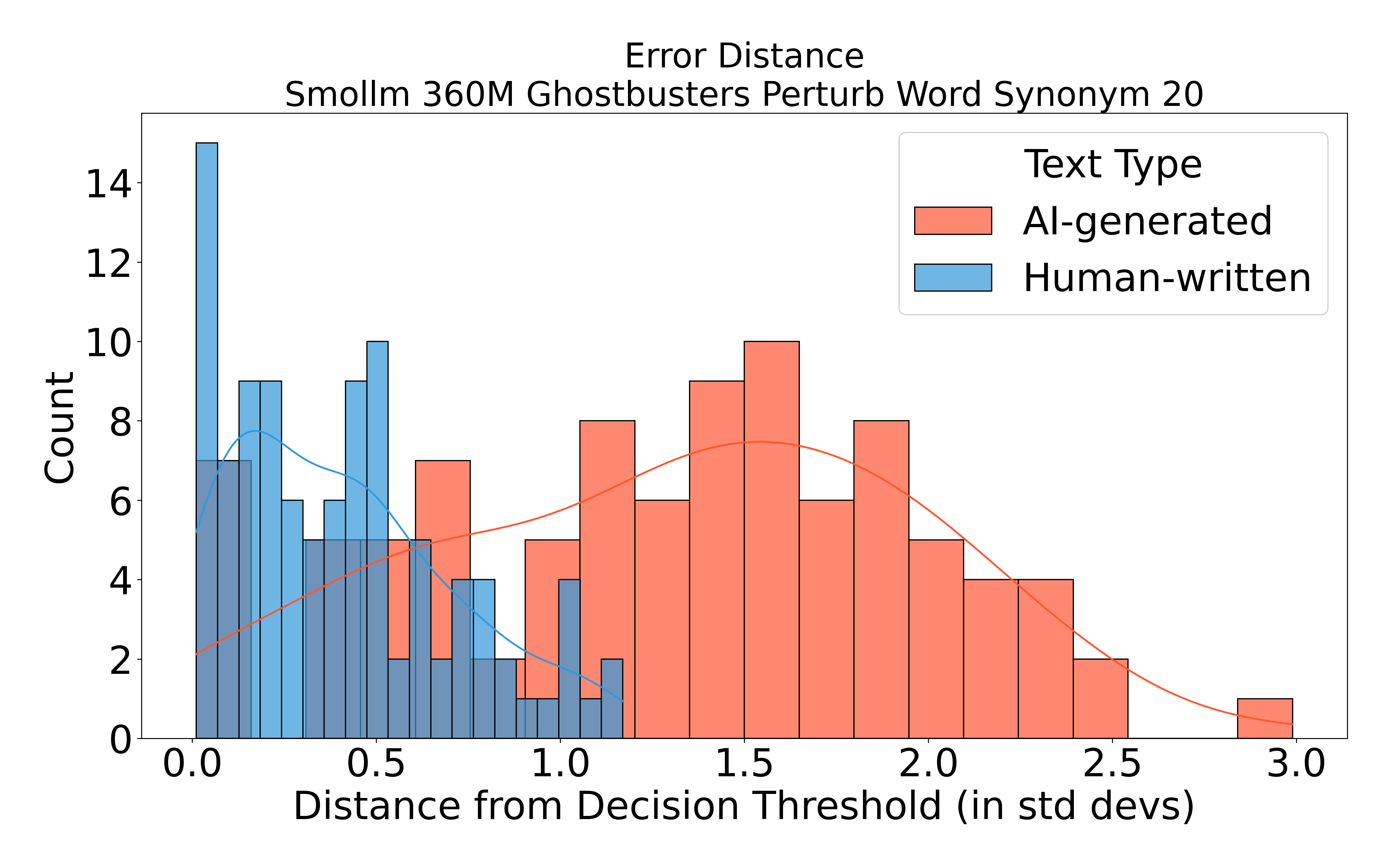}
        \caption{Word Synonym Perturbation}
        \label{fig:smollm_synonym}
    \end{subfigure}
    \hfill
    \begin{subfigure}[b]{0.48\textwidth}
        \centering
        \includegraphics[width=\textwidth]{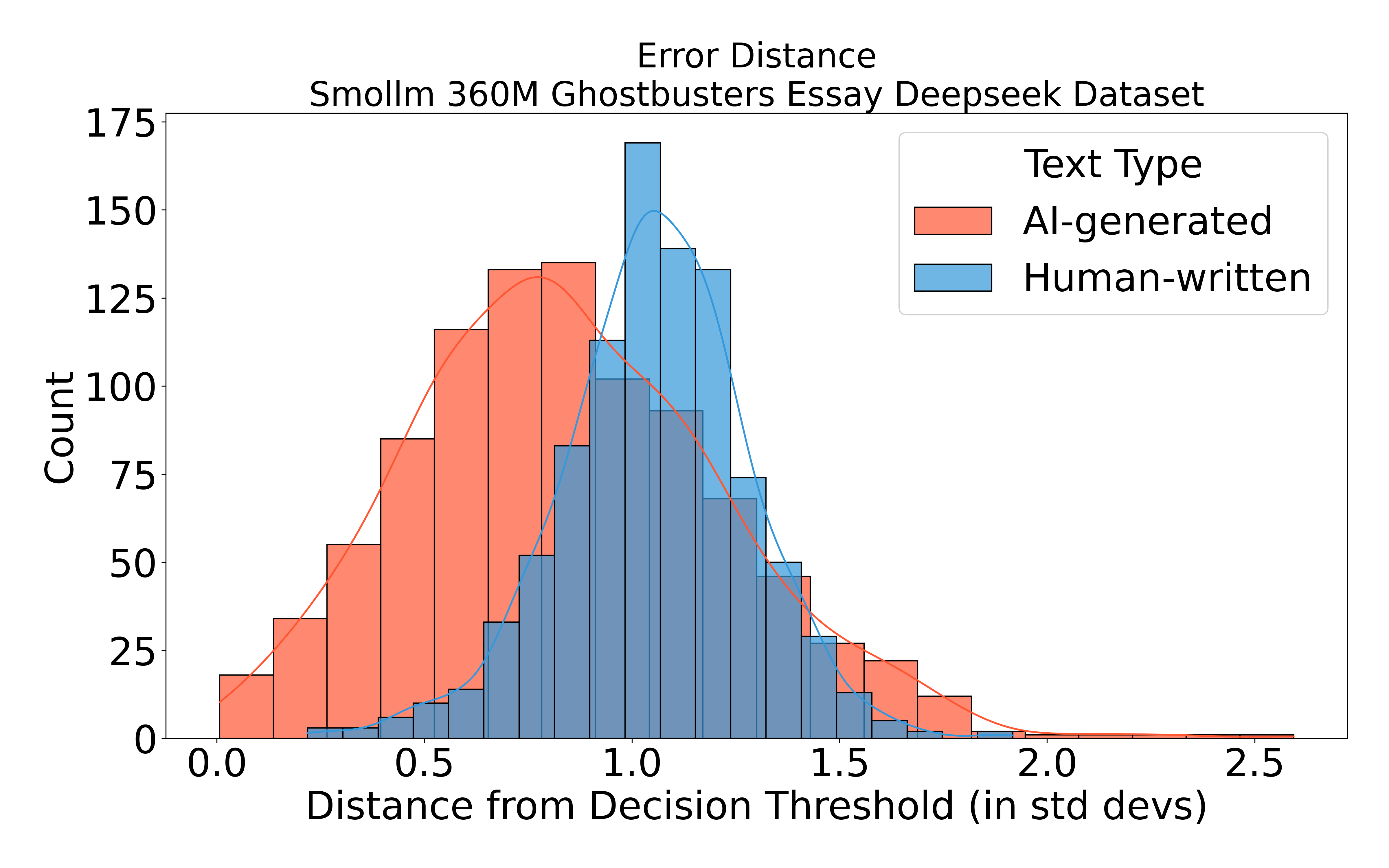}
        \caption{Deepseek Essay Dataset}
        \label{fig:smollm_deepseek}
    \end{subfigure}
    
    \vspace{1.5em}
    
    \begin{subfigure}[b]{0.48\textwidth}
        \centering
        \includegraphics[width=\textwidth]{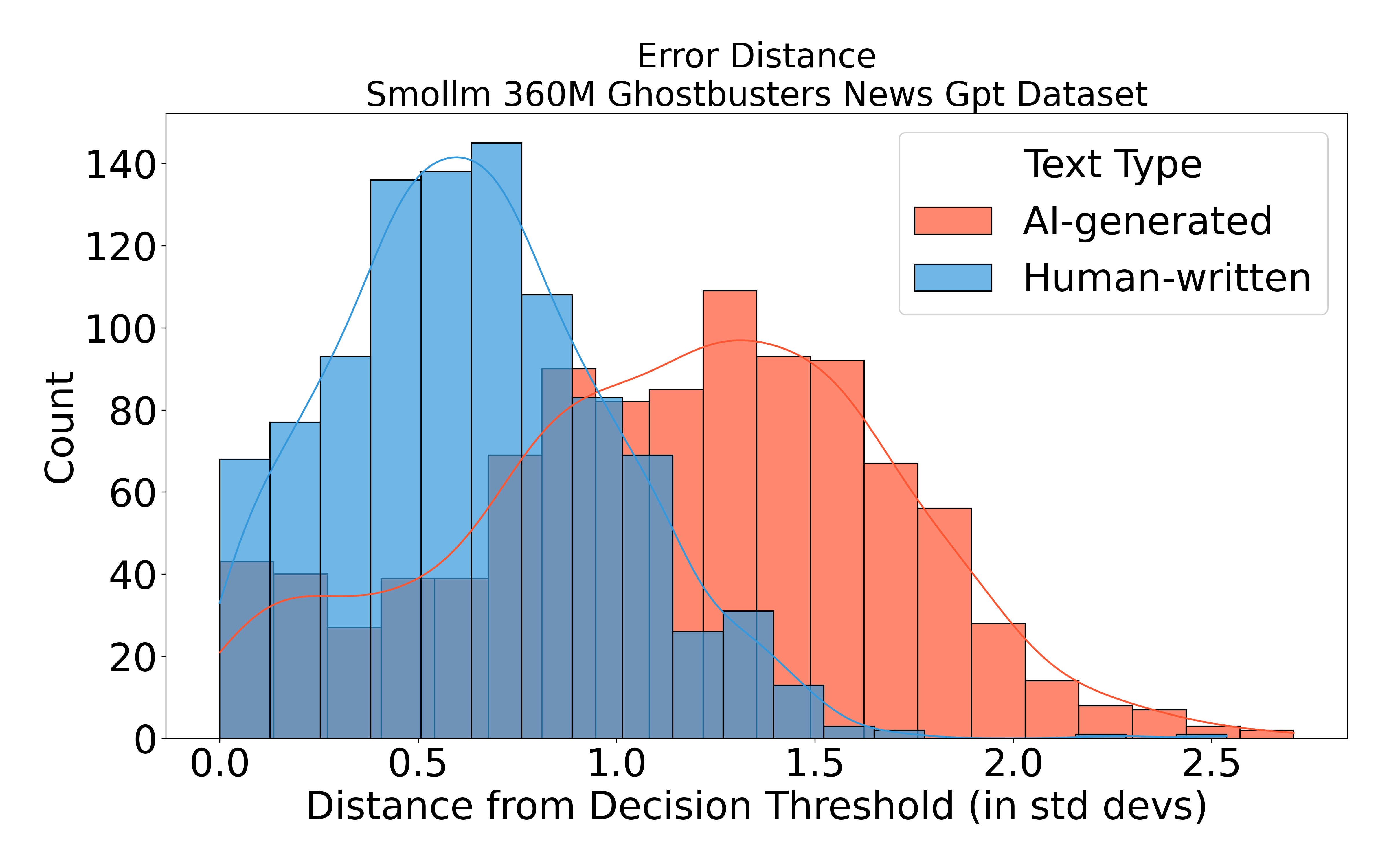}
        \caption{ChatGPT News dataset}
        \label{fig:smollm_news}
    \end{subfigure}
    
    \caption{Error distance metrics for SmolLM 360M across different datasets and perturbation types.}
    \label{fig:combined-error-distances}
\end{figure*}

Understanding misclassifications is important in any high impact application. We find that for Telescope Perplexity the expected distance from the  threshold for a classification made in error is \begin{math}\approx 1\end{math} standard deviation, with human written texts misclassified as AI being less than that on average and AI text misclassified as human typically being more than a single standard deviation from the threshold. Figure \ref{fig:combined-error-distances} shows a couple of examples of the distribution of misclassifications with Deepseek-v3 being the only model in the entire suite of reference models to have a lower 
 \begin{math}\sigma\end{math}  for the AI generated misclassifications than human ones.
 

\subsection{Statistical Significance and Confidence Intervals}

To compute our confidence intervals, we use a python package by \citet{jacobgildenblatconfidenceinterval}, which contains functions to compute the 2 standard deviation confidence intervals of an AUROC score using the Delong method \cite{65f9f828-9f33-36dc-9429-5d215792ea89}, \cite{6851192} and the 2 standard deviation confidence intervals of the F1-Score using the Takahashi method \cite{10.1007/s10489-021-02635-5}. The 2 standard deviation confidence intervals for the true positive rate at 5\% false positive rate score was found simply by bootstrapping.

\subsection{Poor Transfer Examples}

\begin{figure*}[!htbp]
\centering
\begin{minipage}{0.48\textwidth}
    \centering
    \includegraphics[width=\linewidth]{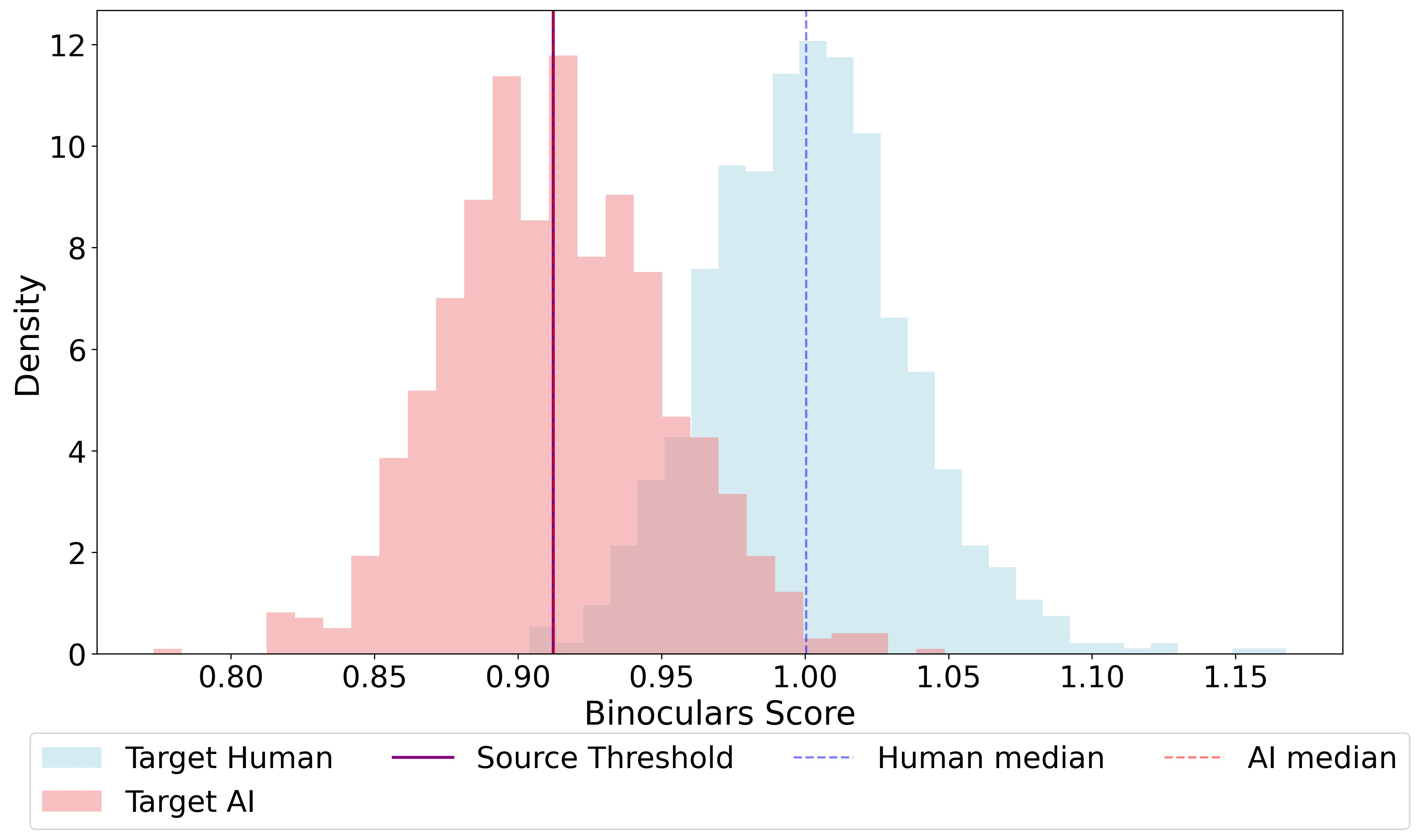}
\end{minipage}\hfill
\begin{minipage}{0.48\textwidth}
    \centering
    \includegraphics[width=\linewidth]{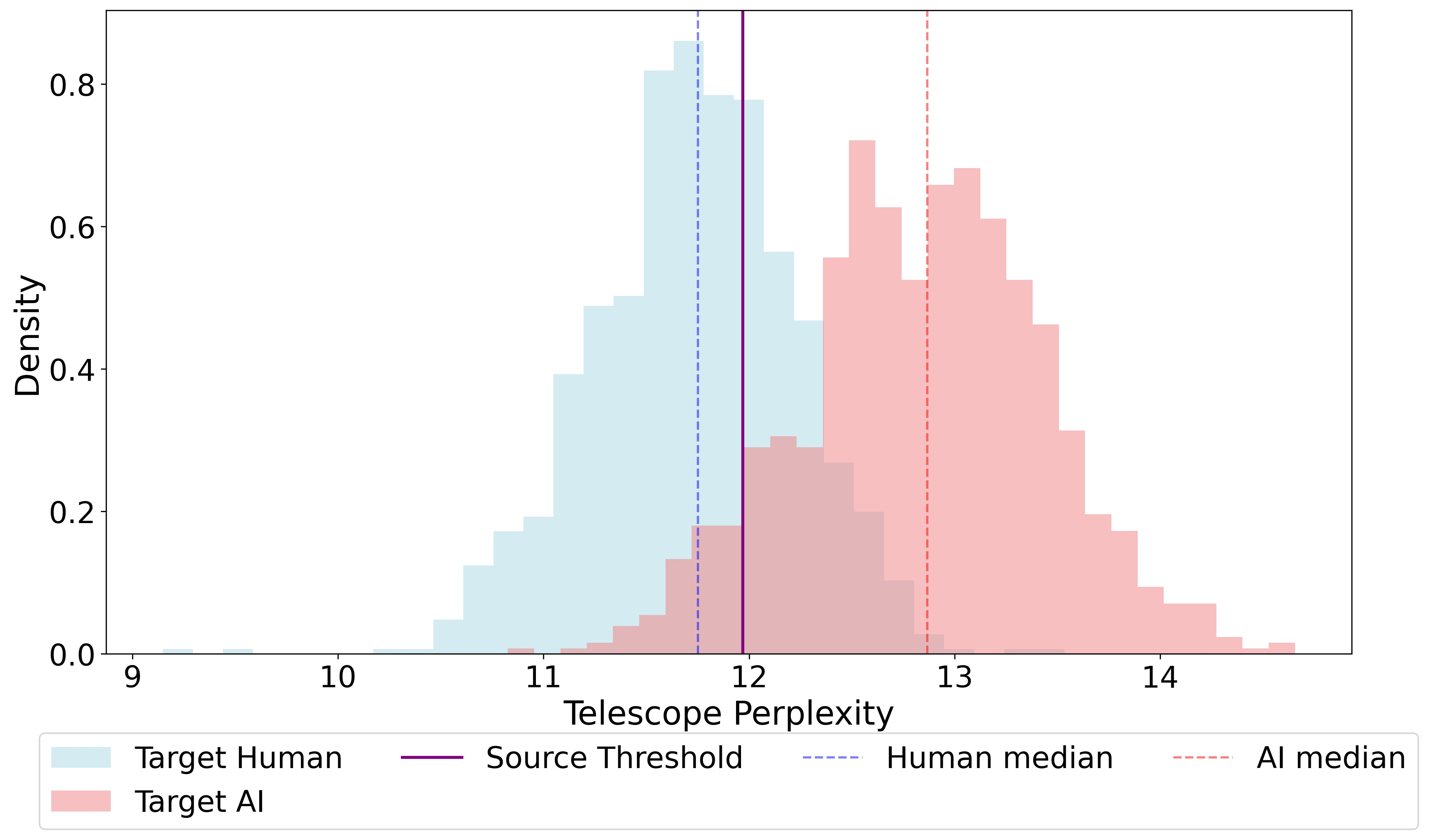}
\end{minipage}
\caption{Example of poor threshold transfer. Comparing transfer from AI-Human to GB Creative Claude}
\label{fig:transfer_plots}
\end{figure*}

If only shown a specific target model on a specific type of text, Telescope Perplexity can have poor threshold transfer capabilities. An illustrative example of poor threshold transfer for Telescope Perplexity is shown in  Figure \ref{fig:transfer_plots}. This highlights the importance of using a diverse set of data to tune Telescope Perplexity's threshold.

Even with this single example of poor transfer, it is important to note that Telescope Perplexity's transfer performance is still generally better than other approaches as seen in Appendix Section \ref{sec:full_results}. 

\subsection{Computational Resources And Execution Time}

The majority of these experiments were performed on Nvidia L40S GPUs with 48 GB of VRAM. Every model was run using 8 bit quantization using the BitsAndBytes\footnote{https://huggingface.co/docs/bitsandbytes/en/index} library to ensure that there was minimal loss of precision when benchmarking each detector, while also allowing us to run two models at once on single GPU. A subset of the Ghostbusters datasets with large models such as Gemma2 9B were prone to out of memory exceptions, so for this reason, we utilized A100 GPUs with 80 GB of VRAM instead when necessary. Compute resource constraints did not allow us to test on reference models larger than 9 billion parameters. We have shown that in this range there is no direct correlation between model size and detector accuracy, but this relationship may not hold at larger model sizes.

The execution time of the experiments were highly dependent on the reference model used, the number of reference models a detection algorithm used, and the dataset. Different datasets contain different numbers of samples while different reference models may have more or less active parameters. In order to save time and effort while running these experiments, many of these experiments were run while batching multiple reference models and detection techniques in a single run, which further complicates matters, and makes it difficult to ascertain the run-time of a single experiment. Therefore in Table \ref{tab:exec_time} we report the total runtime of experiments on an L40S GPU on the Detect LLM Text dataset (10,000 samples). Knowing these execution times should provide a rough sense of how long all of the other experiments in this paper would take to run individually.

\begin{table}[H]
    \caption{Execution times of each technique used in this study on the Detect LLM Text dataset (10,000 samples) with L40S GPUs.
    } \label{tab:exec_time}
    \centering
    \resizebox{8.5cm}{!}{
    \begin{tabular}{l|ccccc}
        \toprule
        \multirow{2}{*}{Reference Model} & \multicolumn{5}{c}{Execution Time (minutes)} \\
        & Telescope & Binoculars & Perplexity & DetectLLM LRR & Fast-DetectGPT\\

        \midrule    


        Gemma2 2B & 63.43 & 72.18 & 63.54 & 35.67 & 33.05 \\
        Gemma2 9B & 73.63 & 88.69 & 73.91 & 44.60 & 42.84 \\
        Llama3 8B & 53.24 & 64.04 & 52.80 & 42.12 & 37.16 \\
        Falcon 7B & 44.95 & 65.62 & 45.86 & 45.43 & 37.02 \\
        GPT-Neo 2.7B & 45.51 & 65.42 & 45.52 & 47.70 & 40.93 \\
        GPT-J 6B & 64.98 & 105.00 & 65.21 & 69.37 & 61.10 \\
        SmolLM 135M & 31.98 & 35.39 & 32.34 & 35.76 & 24.81 \\
        SmolLM 360M & 31.86 & 36.08 & 31.15 & 34.66 & 28.74 \\
        SmolLM 1.7B & 30.03 & 33.39 & 30.25 & 32.24 & 24.80 \\
        SmolLM2 135M & 28.58 & 31.65 & 28.58 & 31.18 & 24.92 \\
        SmolLM2 360M & 28.29 & 31.70 & 28.24 & 30.65 & 24.77 \\
        SmolLM2 1.7B & 30.66 & 34.33 & 30.62 & 31.12 & 25.06 \\

    \midrule    
    \end{tabular}
    }
\end{table}

\subsection{Unused Performance Metrics}
\label{sec:unused_performance_metrics}

Unlike Binoculars, we decide to completely disregard using the TPR at ultra low FPR score, such as 0.05\% FPR since we simply do not have enough data in our datasets to make accurate and reliable measurements of behavior that only happens once in every 10,000 samples. We initially attempted to use these metrics, but during testing, we noticed substantial jumps of around 40\% near the end of testing a dataset, which helps corroborate that using these TPR at ultra low FPR scores as performance metrics to rate classifiers is unreliable and high variance for the number of samples that we have from each dataset. Other works by \citet{tufts2025practicalexaminationaigeneratedtext} show that the confidence intervals for this metric are massive and inconclusive.



\subsection{Additional Training Dynamics} 
\label{app:additional_training_dynamics}
\begin{figure}
    \centering
    \begin{subfigure}{0.48\textwidth}
        \centering
        \includegraphics[width=\linewidth]{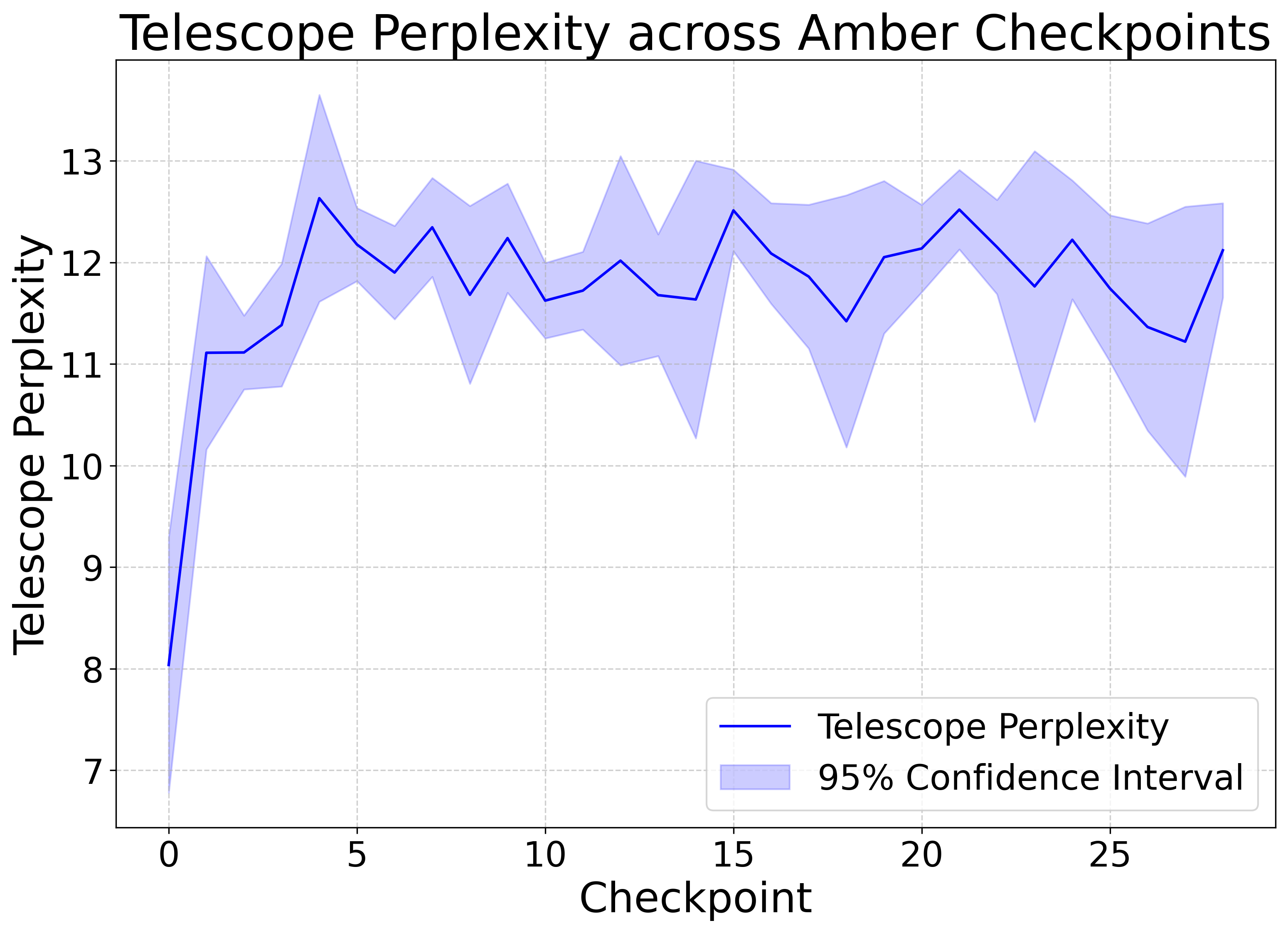}
        \caption{Early training of \citet{liu2023llm360}'s Amber-7B}
        \label{fig:Amber_training}
    \end{subfigure}
    \hfill
    \begin{subfigure}{0.48\textwidth}
        \centering
        \includegraphics[width=\linewidth]{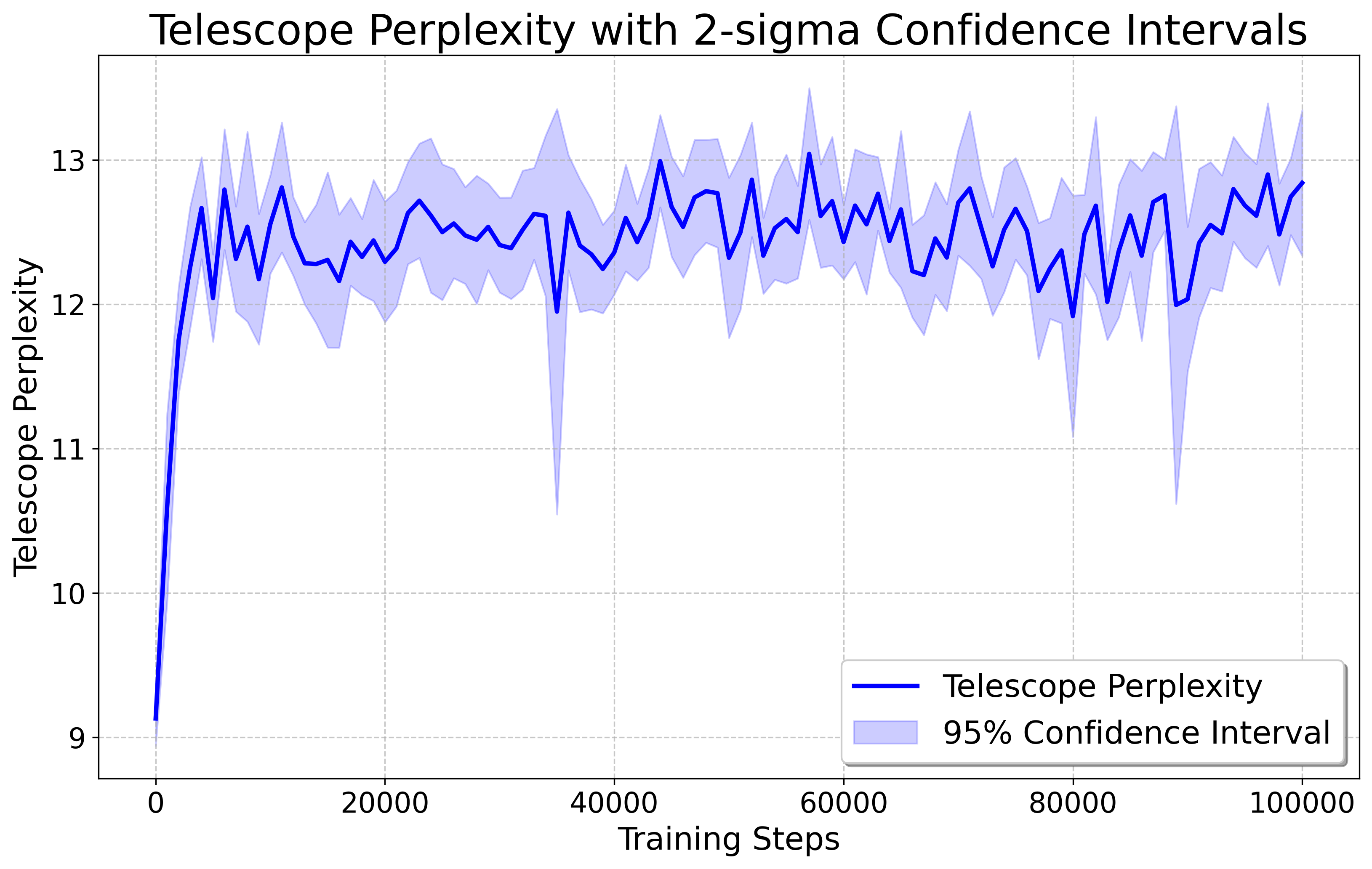}
        \caption{First 100k training steps of \citet{biderman2023pythia}'s Pythia-160M model}
        \label{fig:Pythia_training}
    \end{subfigure}
    \caption{Training curves for language models}
    \label{fig:training_comparison}
\end{figure}
\citet{liu2023llm360} provide checkpoints throughout the training of their 7 billion parameter model Amber, these are taken regularly but the number of steps per checkpoint is not given. Compute constraints make evaluating all the Amber-7B model checkpoints infeasible so we instead focus on the early training steps to validate our hypothesis. We also present a chart showing the emergence of the ``Vestigial Heuristic'' in a much smaller model (Pythia-160M). As shown in Figure \ref{fig:training_comparison} we observe Telescope Perplexity rising early in training before stabilizing across both model architectures and across a few model sizes.

\subsection{Multilingual Ablation}  
The Telescope Perplexity probe is designed to apply to English-language models. For completeness, we include limited multilingual testing on \citet{wang2024m4multigeneratormultidomainmultilingual}'s M4 dataset using \citet{gemmateam2024gemma2improvingopen}'s Gemma2-2b. Variance in the original data sources across languages is large enough to make comparison between languages difficult. Additionally, \citet{gemmateam2024gemma2improvingopen}'s models are ``not trained specifically for state-of-the-art multilingual capabilities,'' making the performance of each technique quite remarkable.

\begin{table}[htbp]
    \centering
    \caption{Performance Metrics by Language and Method with Gemma2-2B as reference model}
    \label{tab:performance_metrics}

    \resizebox{8.5cm}{!}{
    \begin{tabular}{llcc}
    \toprule
    \textbf{Language} & \textbf{Metric} & \textbf{AUC} & \textbf{Confidence Interval} \\
    \midrule

    \multirow{5}{*}{Chinese} & Telescope Perplexity & 0.94944 & (0.94354, 0.95533) \\
                             & Binoculars Score     & 0.98895 & (0.98677, 0.99114) \\
                             & Perplexity           & 0.96257 & (0.95722, 0.96792) \\
                             & LRR                  & 0.95358 & (0.94848, 0.95868) \\
                             & Fast-DetectGPT       & 0.94745 & (0.94208, 0.95282) \\
    \midrule
    \multirow{5}{*}{Russian} & Telescope Perplexity & 0.66543 & (0.65152, 0.67935) \\
                             & Binoculars Score     & 0.89772 & (0.88992, 0.90551) \\
                             & Perplexity           & 0.70169 & (0.68836, 0.71502) \\
                             & LRR                  & 0.63027 & (0.61578, 0.64476) \\
                             & Fast-DetectGPT       & 0.78909 & (0.77760, 0.80057) \\
    \midrule
    \multirow{5}{*}{Urdu}    & Telescope Perplexity & 0.93901  & (0.93226, 0.94576) \\
                             & Binoculars Score     & 0.99996  & (0.99991, 1.00000) \\
                             & Perplexity           & 0.97867  & (0.97517, 0.98217) \\
                             & LRR                  & 0.98020  & (0.97694, 0.98346) \\
                             & Fast-DetectGPT       & 0.99975 & (0.99962, 0.99988) \\
    \midrule
    \multirow{5}{*}{Arabic}  & Telescope Perplexity & 0.94478 & (0.93926, 0.95030) \\
                             & Binoculars Score     & 0.97650 & (0.97324, 0.97977) \\
                             & Perplexity           & 0.92182 & (0.91478, 0.92887) \\
                             & LRR                  & 0.90266 & (0.89520, 0.91012) \\
                             & Fast-DetectGPT       & 0.93667 & (0.93080, 0.94254) \\

    \bottomrule
    \end{tabular}
    }
\end{table}

\subsection{Adversarial Temperature and Prompt Attacks}

In order to gauge how the temperature of the target model affects Telescope Perplexity's performance on detecting text from that target model, we have decided to provide a small ablation study where we generate new datasets using different target model sampling temperatures. To conduct this ablation, we regenerate the Ghostbusters Essay dataset with the GPT4o target model with sampling temperatures of 0, 1 (the default for the OpenAI API), and 1.2 (the highest temperature that was also stable). Using a reference model of SmolLM 360M, we find that with a temperature of 0 corresponds to an AUROC of 0.999992, a temperature of 1 corresponds to an AUROC of 0.99993, and a temperature of 1.2 corresponds to an AUROC of 0.9988. Overall, we see a trend that a lower sampling temperature leads to better detection performance, which makes sense, since as the temperature of the target language model decreases, the chance for the language model to accidentally mimic human text via random sampling also decreases.

In addition, we also wish to study how prompt attacks affect the performance of the Telescope Perplexity. One simple prompting attack that people may employ to dodge detection is asking the target model to attempt to repeat words, which may interfere with the token repetition signal. To accomplish this, we regenerated the Ghostbusters Essay dataset with the GPT4o target model, but instead of using the system prompt ``You are a helpful assistant.'', we use the system prompt ``You are a helpful assistant. Try to repeat key words one after another while still following the prompt''. In doing so, we found that the AUROC on the dataset with the reference model of SmolLM 360M degraded from 0.99993 to 0.996. This suggests a slight but statistically significant degradation; however, this is not as bad as one may expect from these types of adversarial prompting attacks.
\subsection{LLM-as-classifier}\label{appendix:LLM-as-classifier}
We performed a simple statistical test using Gemini 3.5-Flash as a classifier. This regime fails to beat chance (p = 0.184 > 0.05) in our simple 50 sample test drawn from \citet{verma2024ghostbusterdetectingtextghostwritten}'s GPT essay dataset.
\subsection{Performance Drift}
\begin{figure}
    \centering
    \includegraphics[width=1\linewidth]{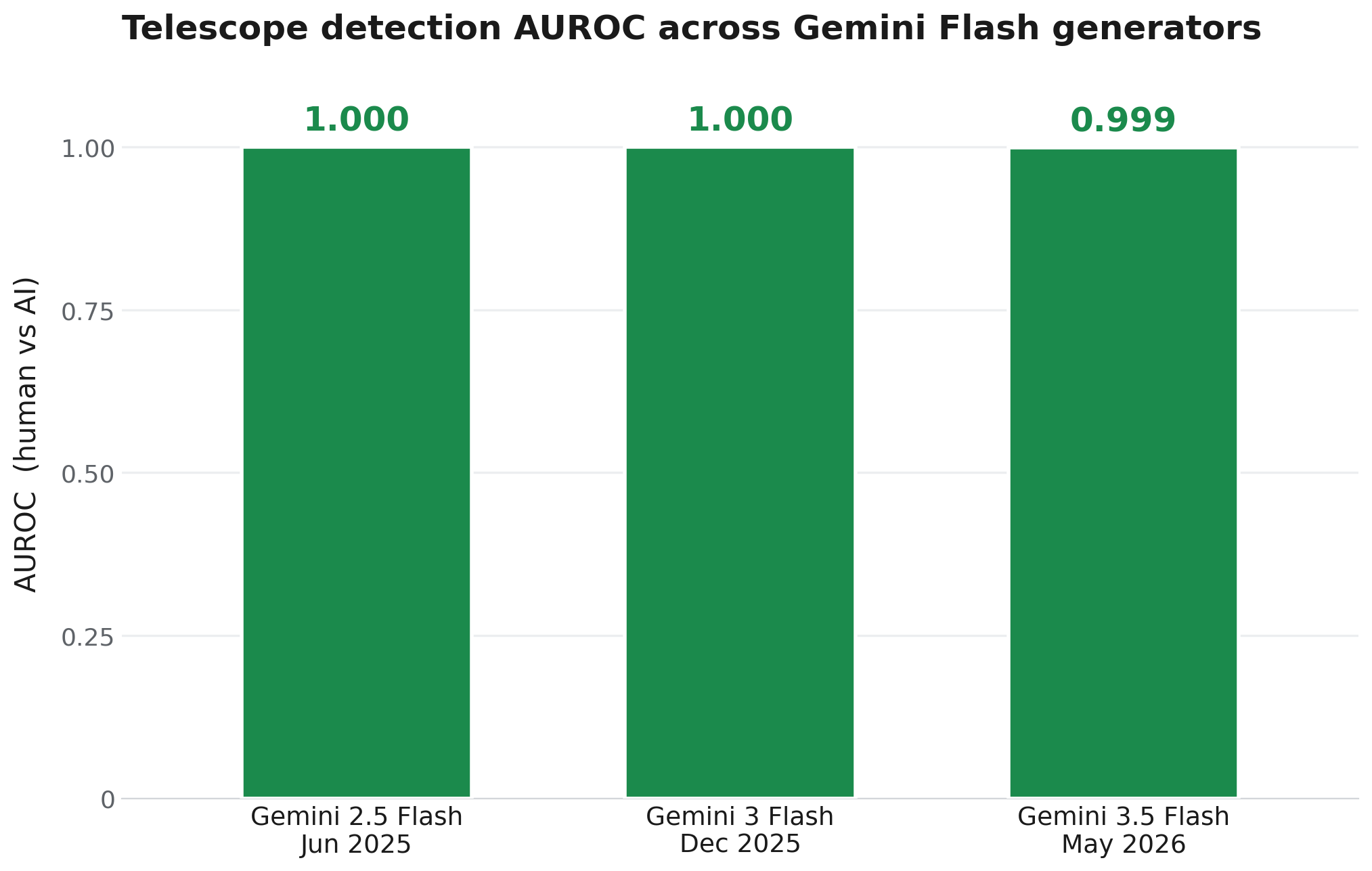}
    \caption{Telescope Perplexity maintains its high AUC across many model releases.}
    \label{fig:auc_drift}
\end{figure}

We measure Telescope Perplexity's performance on the last years slate of Gemini flash model releases (2.5, 3.0,3.5). We find no statistically significant change in model performance over these model releases as shown in Figure \ref{fig:auc_drift}.

\subsection{Adversarial AI Humanizers}
\label{appendix:Adversarial AI Humanizers}
To assess detection robustness under adversarial conditions, we employ a BERT-based humanization system that iteratively modifies text to minimize detection scores. The algorithm operates as follows:

Given the input text $T$ and the human baseline score $\mu_H$, the system randomly masks the tokens with probability $p_m = 0.08$. For each masked position, a BERT model generates $k=2$ candidate replacements from the top-$k$ most probable tokens. Candidates are filtered by semantic similarity $\text{sim}(T, T') \geq 0.98$ using sentence embeddings. Among valid candidates, the system selects the replacement that minimally decreases the detection score relative to $\mu_H$, accepting changes only if they move its score closer to the human baseline.

This process iterates for 500 steps, generating $n=4$ variants per input and selecting the optimal variant by detection score. Notably, we choose minimally destructive settings for the humanizer where semantic meaning is strongly preserved. In practice, more destructive settings can decrease detector accuracy at the cost of the text becoming incomprehensible compared to the original. Telescope maintains strong performance with degradation concentrated in datasets where its performance is typically weaker (see Tables \ref{fig:humanizer_performance} and \ref{fig:humanizer_performance2}).

\begin{table*}[t]
    \captionof{table}{Detection performance of Telescope Perplexity, Binoculars, Perplexity, DetectLLM LRR, and Fast-DetectGPT across humanizer-perturbed datasets using SmolLM 360M as reference model. We report the AUROC of each detection technique. The best performance on each dataset is bolded.} \label{fig:humanizer_performance}
    \centering
    \resizebox{1\linewidth}{!}{
    \begin{tabular}{l|lccccc}
        \toprule
         \multirow{2}{*}{Dataset} & \multirow{2}{*}{Reference Model} & \multicolumn{5}{c}{AUROC} \\
        & & Telescope & Binoculars & Perplexity & DetectLLM LRR & Fast-DetectGPT\\
        \midrule
        
\multirow{1}{*}{GB Creative Claude}
& SmolLM 360M & \textbf{0.99507} & 0.88834 & 0.95074 & 0.93432 & 0.69787  \\
\midrule
\multirow{1}{*}{GB Creative DeepSeek}
& SmolLM 360M & 0.90452 & \textbf{0.98675} & 0.97280 & 0.94070 & 0.91483  \\
\midrule
\multirow{1}{*}{GB Creative GPT-4o Adversarial Prompt}
& SmolLM 360M & \textbf{0.98217} & 0.92634 & 0.97159 & 0.96413 & 0.77258  \\
\midrule
\multirow{1}{*}{GB Creative GPT-4o}
& SmolLM 360M & \textbf{0.98831} & 0.94116 & 0.93884 & 0.90649 & 0.76582  \\
\midrule
\multirow{1}{*}{GB Creative ChatGPT}
& SmolLM 360M & \textbf{1.00000} & 0.81408 & 0.98507 & 0.98102 & 0.50416 \\
\midrule
\multirow{1}{*}{GB Essay Claude}
& SmolLM 360M & \textbf{0.97759} & 0.77987 & 0.93455 & 0.93838 & 0.63550 \\
\midrule
\multirow{1}{*}{GB Essay DeepSeek}
& SmolLM 360M & 0.92191 & 0.90351 & \textbf{0.99880} & 0.99453 & 0.72387 \\
\midrule
\multirow{1}{*}{GB Essay GPT-4o Adversarial Prompt}
& SmolLM 360M & 0.84277 & 0.72826 & 0.97662 & \textbf{0.98000} & 0.59864 \\
\midrule
\multirow{1}{*}{GB Essay GPT-4o}
& SmolLM 360M & 0.94656 & 0.76053 & 0.98266 & \textbf{0.98733} & 0.58917 \\
\midrule
\multirow{1}{*}{GB Essay ChatGPT}
& SmolLM 360M & 0.99186 & 0.70428 & 0.99373 & \textbf{0.99498} & 0.57298 \\
\midrule
\multirow{1}{*}{GB News Claude}
& SmolLM 360M & 0.89180 & 0.91674 & \textbf{0.95341} & 0.90736 & 0.87278 \\
\midrule
\multirow{1}{*}{GB News ChatGPT}
& SmolLM 360M & 0.98826 & 0.98662 & \textbf{0.99836} & 0.99618 & 0.96857 \\
        \bottomrule
    \end{tabular}
    }
\end{table*}

\begin{table*}[t]
    \captionof{table}{AUROC confidence intervals (95\%) for detection methods across humanizer-perturbed datasets using SmolLM 360M. The method with the narrowest confidence interval for the highest AUROC is bolded.}\label{fig:humanizer_performance2}
    \centering
    \resizebox{1\linewidth}{!}{
    \begin{tabular}{l|lccccc}
        \toprule
        \multirow{2}{*}{Dataset} & \multirow{2}{*}{Reference Model} & \multicolumn{5}{c}{AUROC Confidence Intervals} \\
        & & Telescope & Binoculars & Perplexity & DetectLLM LRR & Fast-DetectGPT\\
        \midrule
        
\multirow{1}{*}{GB Creative Claude}
& SmolLM 360M & (0.98407, 1.00608) & (0.77655, 1.00013) & (0.89862, 1.00286) & (0.86970, 0.99894) & (0.54258, 0.85315)  \\
\midrule
\multirow{1}{*}{GB Creative DeepSeek}
& SmolLM 360M & (0.86856, 0.94047) & (0.97454, 0.99896) & (0.95688, 0.98871) & (0.91289, 0.96851) & (0.88246, 0.94720)  \\
\midrule
\multirow{1}{*}{GB Creative GPT-4o Adversarial Prompt}
& SmolLM 360M & (0.96587, 0.99848) & (0.89467, 0.95802) & (0.95159, 0.99160) & (0.94331, 0.98494) & (0.71800, 0.82716)  \\
\midrule
\multirow{1}{*}{GB Creative GPT-4o}
& SmolLM 360M & (0.97945, 0.99717) & (0.91345, 0.96886) & (0.91276, 0.96493) & (0.87393, 0.93905) & (0.71163, 0.82001)  \\
\midrule
\multirow{1}{*}{GB Creative ChatGPT}
& SmolLM 360M & (1.00000, 1.00000) & (0.76492, 0.86325) & (0.97252, 0.99761) & (0.96723, 0.99481) & (0.43794, 0.57037)  \\
\midrule
\multirow{1}{*}{GB Essay Claude}
& SmolLM 360M & (0.96418, 0.99100) & (0.72729, 0.83245) & (0.90634, 0.96276) & (0.91165, 0.96510) & (0.57316, 0.69784)  \\
\midrule
\multirow{1}{*}{GB Essay DeepSeek}
& SmolLM 360M & (0.89131, 0.95251) & (0.86729, 0.93974) & (0.99704, 1.00056) & (0.98894, 1.00012) & (0.66676, 0.78098)  \\
\midrule
\multirow{1}{*}{GB Essay GPT-4o Adversarial Prompt}
& SmolLM 360M & (0.79225, 0.89329) & (0.66925, 0.78727) & (0.96163, 0.99160) & (0.96819, 0.99181) & (0.53448, 0.66280)  \\
\midrule
\multirow{1}{*}{GB Essay GPT-4o}
& SmolLM 360M & (0.92164, 0.97148) & (0.70457, 0.81649) & (0.96896, 0.99636) & (0.97669, 0.99797) & (0.52477, 0.65357)  \\
\midrule
\multirow{1}{*}{GB Essay ChatGPT}
& SmolLM 360M & (0.97902, 1.00470) & (0.64294, 0.76562) & (0.98717, 1.00029) & (0.98940, 1.00055) & (0.50794, 0.63803)  \\
\midrule
\multirow{1}{*}{GB News Claude}
& SmolLM 360M & (0.85601, 0.92760) & (0.88525, 0.94823) & (0.93007, 0.97676) & (0.87373, 0.94099) & (0.83372, 0.91184)  \\
\midrule
\multirow{1}{*}{GB News ChatGPT}
& SmolLM 360M & (0.98050, 0.99603) & (0.97451, 0.99873) & (0.99623, 1.00049) & (0.99279, 0.99956) & (0.94890, 0.98825)  \\
        \bottomrule
    \end{tabular}
    }
\end{table*}

\subsection{Additional Ablation Results} \label{sec:additional_ablations}

\begin{figure}[h]
    \centering
    \includegraphics[width=1\linewidth]{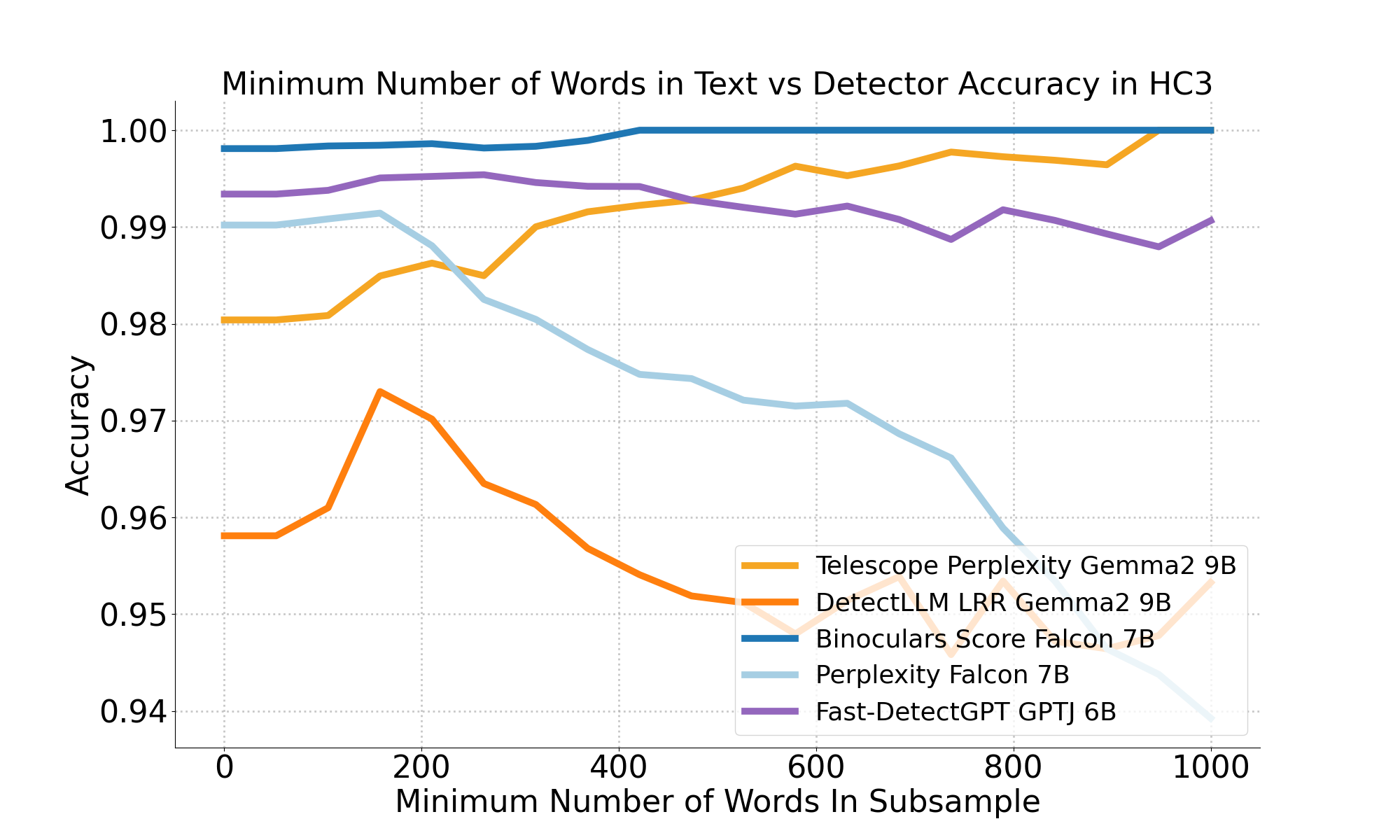}
    \caption{Impact of text length on detector performance (HC3). We filter out samples below a minimum word count and report accuracy on the remaining subset.}
    \label{fig:appendices_text_length_hc3}
\end{figure}

\begin{figure}[h]
    \centering
    \includegraphics[width=1\linewidth]{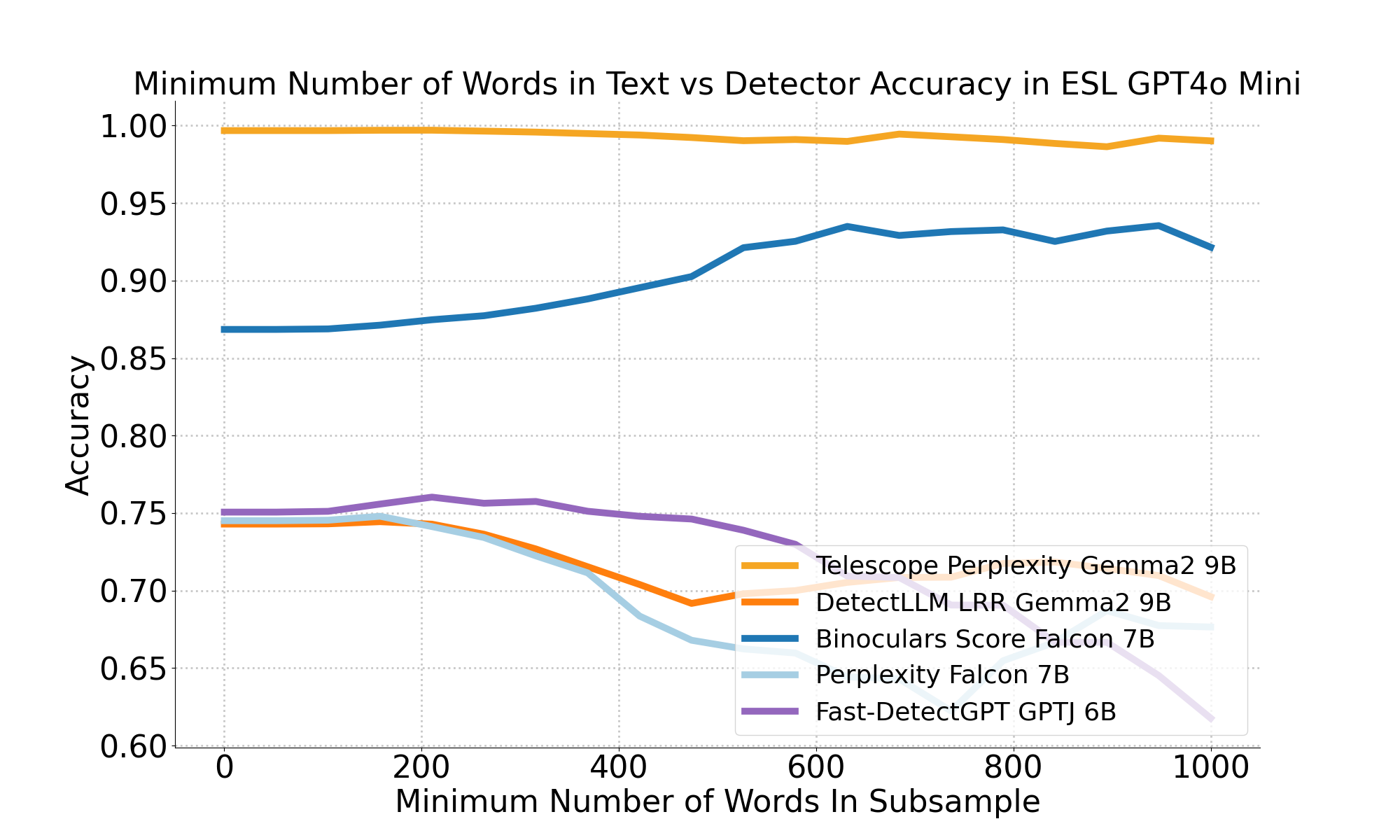}
    \caption{Impact of text length on detector performance (ESL GPT-4o Mini). We filter out samples below a minimum word count and report accuracy on the remaining subset.}
    \label{fig:appendices_text_length_esl}
\end{figure}

\begin{figure}[h]
    \centering
    \includegraphics[width=1\linewidth]{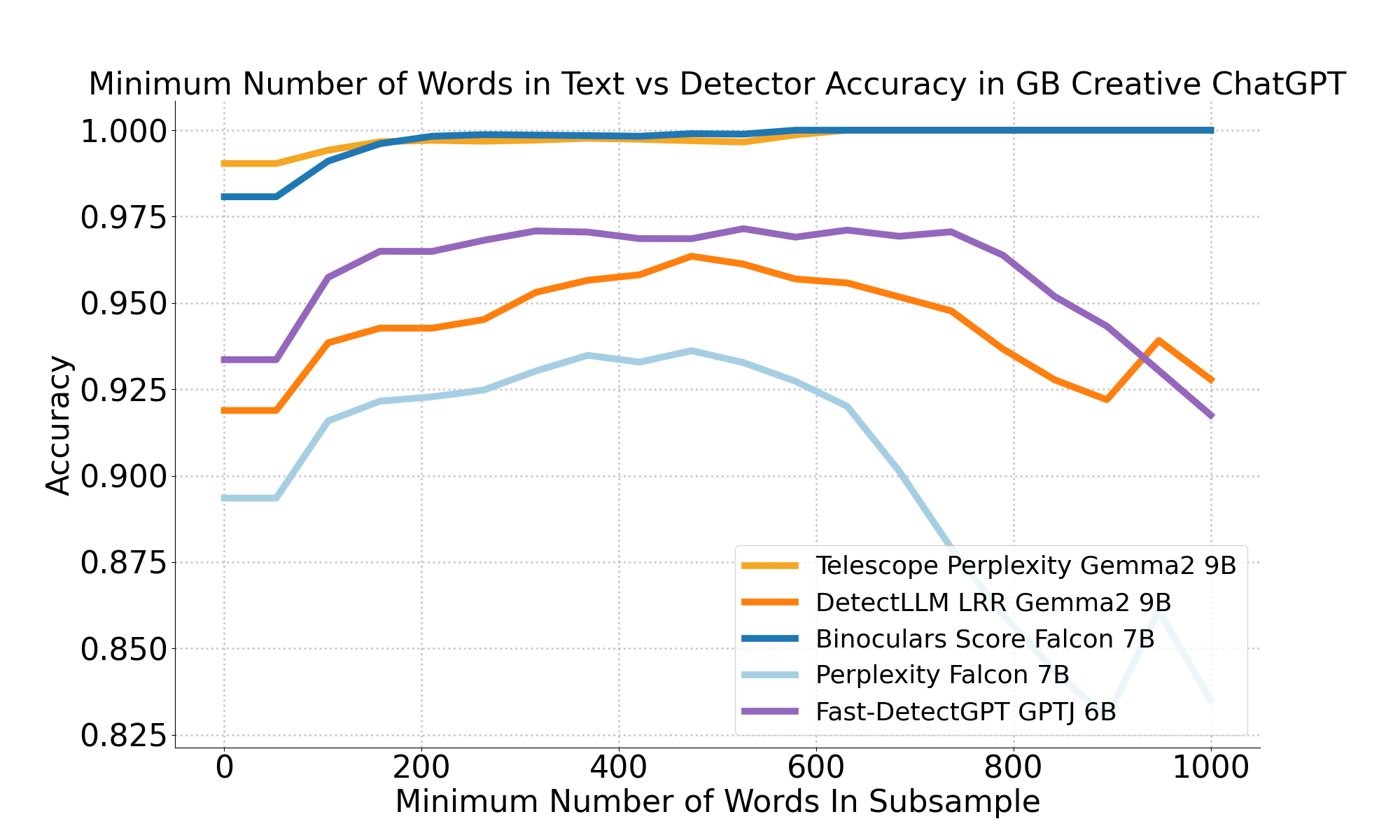}
    \caption{Impact of text length on detector performance (GB Creative GPT). We filter out samples below a minimum word count and report accuracy on the remaining subset.}
    \label{fig:appendices_text_length_gb_creative}
\end{figure}

\begin{figure}[h]
    \centering
    \includegraphics[width=\linewidth]{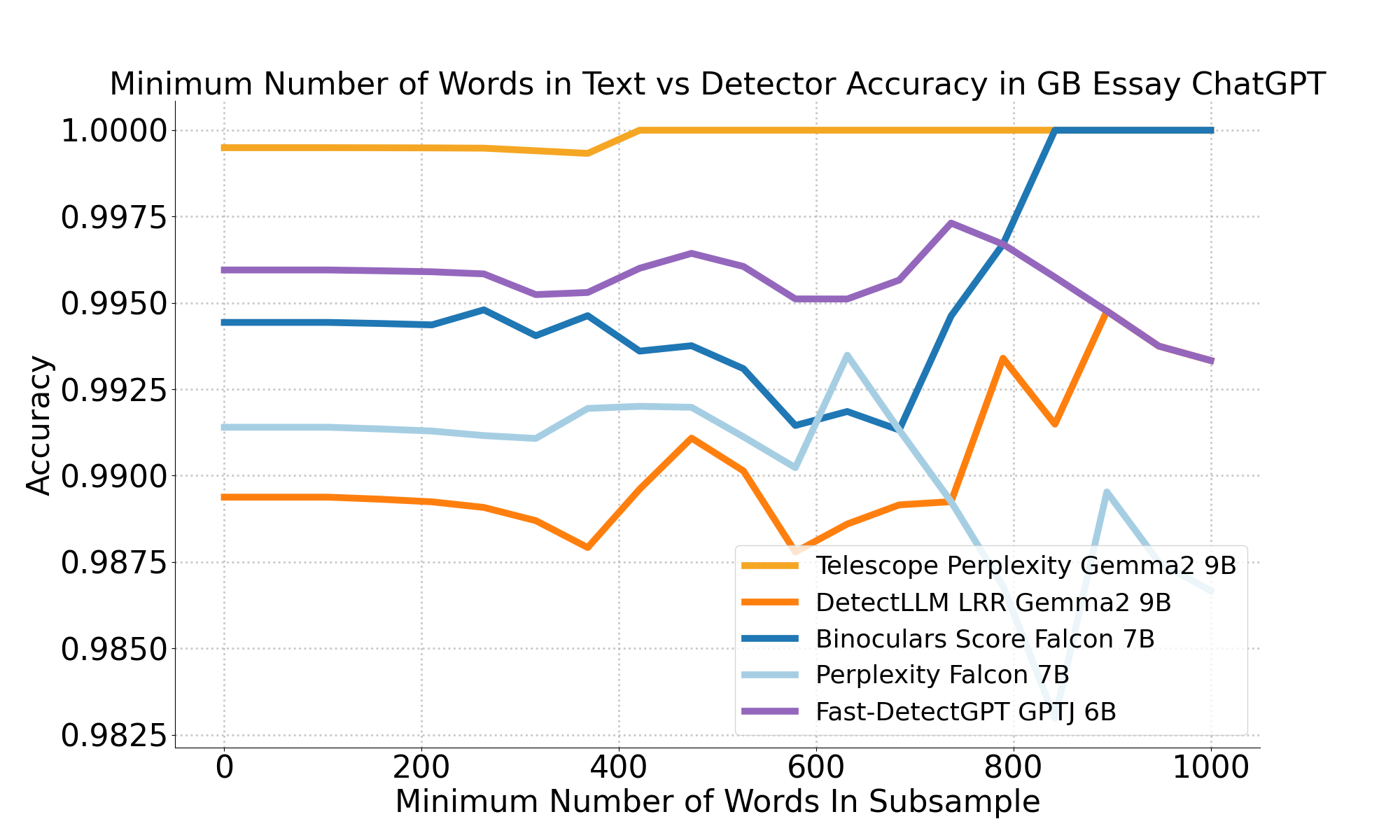}
    \caption{Impact of text length on detector performance (GB Essay GPT). We filter out samples below a minimum word count and report accuracy on the remaining subset.}
    \label{fig:appendices_text_length_gb_essay}
\end{figure}

\begin{figure*}[t]
    \centering
    \begin{subfigure}[b]{0.48\textwidth}
        \centering
        \includegraphics[width=\textwidth]{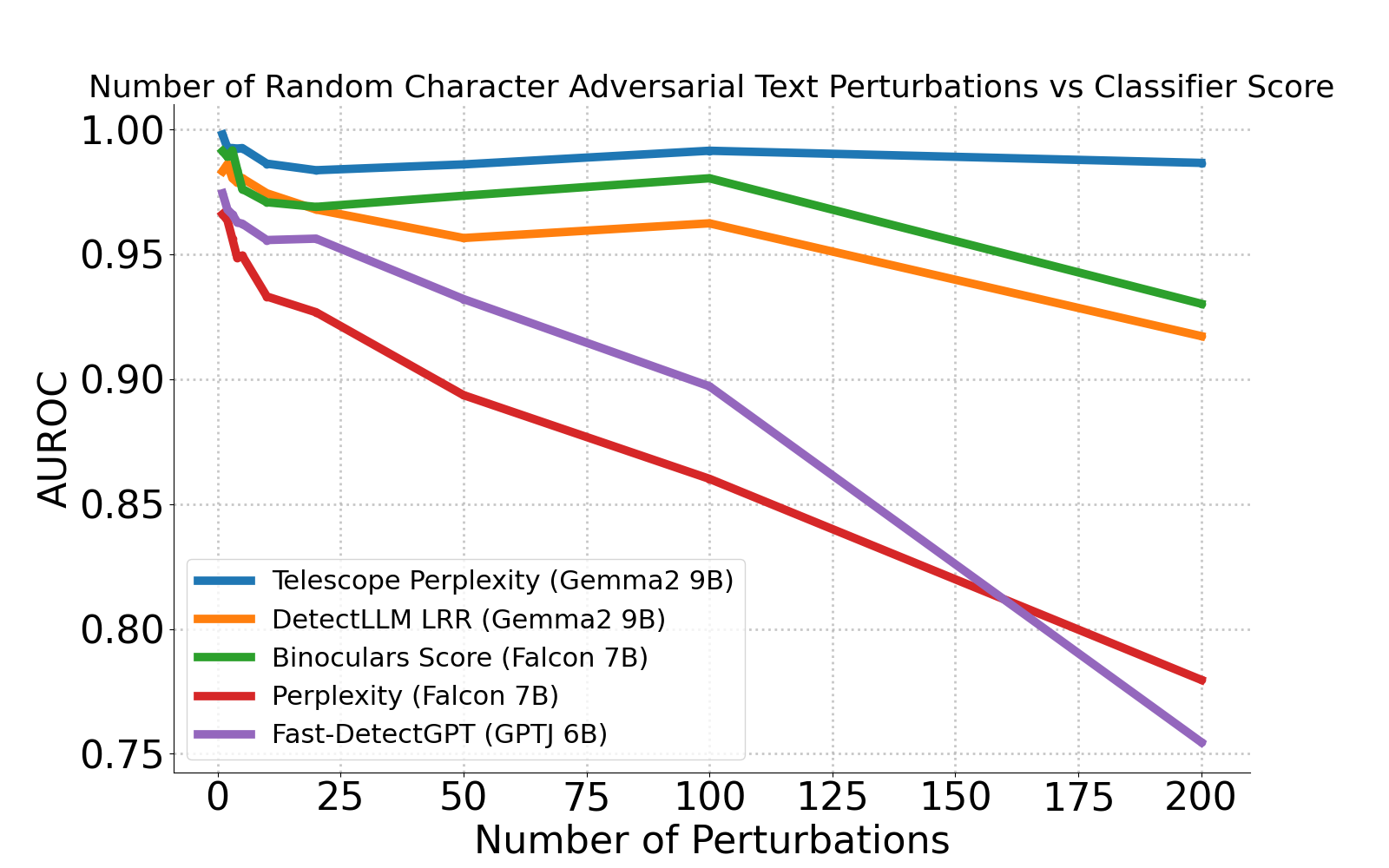}
        \caption{Random characters}
        \label{fig:appendices_perturb_char_basic}
    \end{subfigure}
    \hfill
    \begin{subfigure}[b]{0.48\textwidth}
        \centering
        \includegraphics[width=\textwidth]{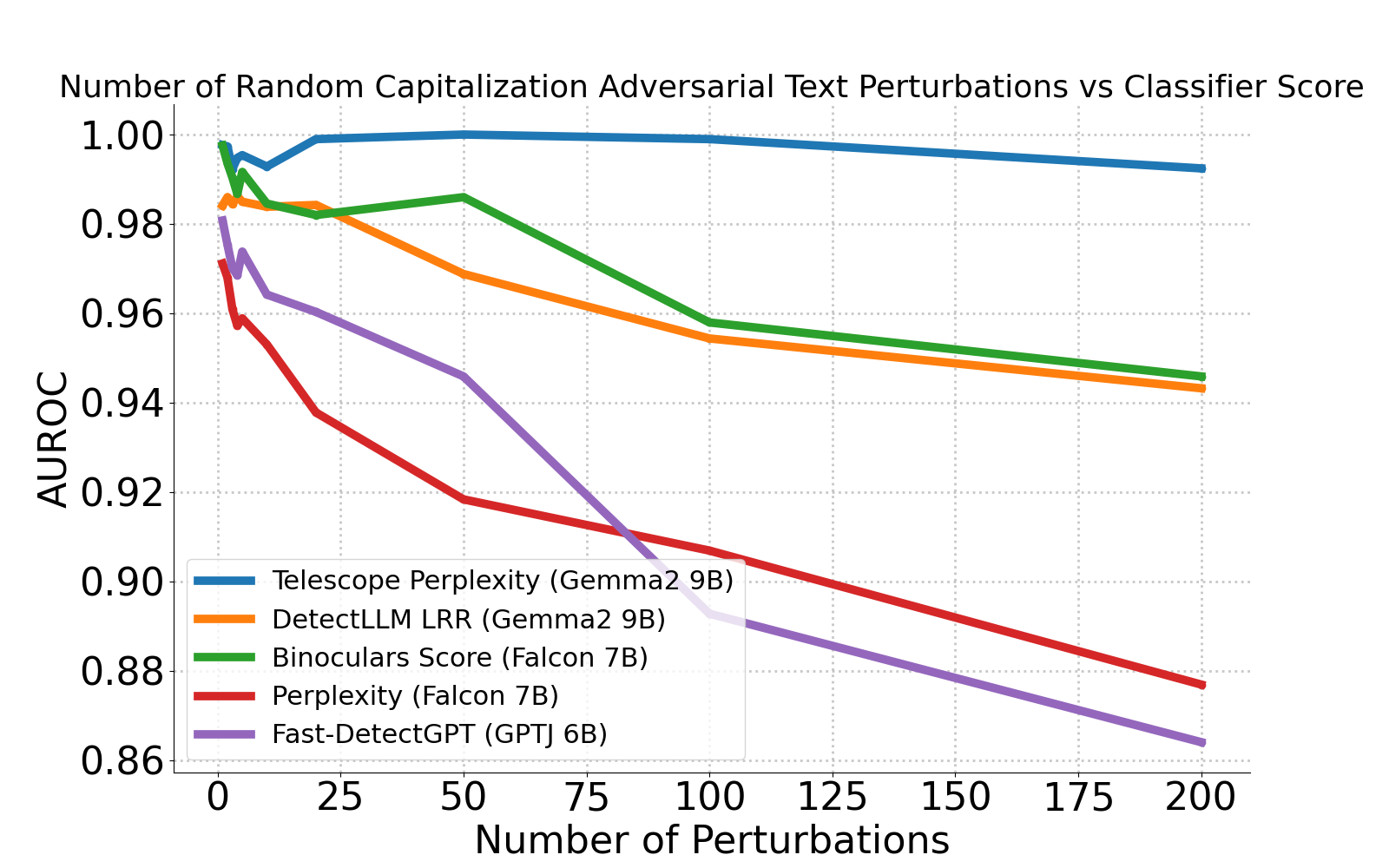}
        \caption{Random capitalization}
        \label{fig:appendices_perturb_char_caps}
    \end{subfigure}
    
    \caption{Impact of random character and random capitalization perturbations on the AUROC of each detector.}
    \label{fig:combined_character_perturbations}
\end{figure*}

\begin{figure*}[t]
    \centering
    \begin{subfigure}[b]{0.48\textwidth}
        \centering
        \includegraphics[width=\textwidth]{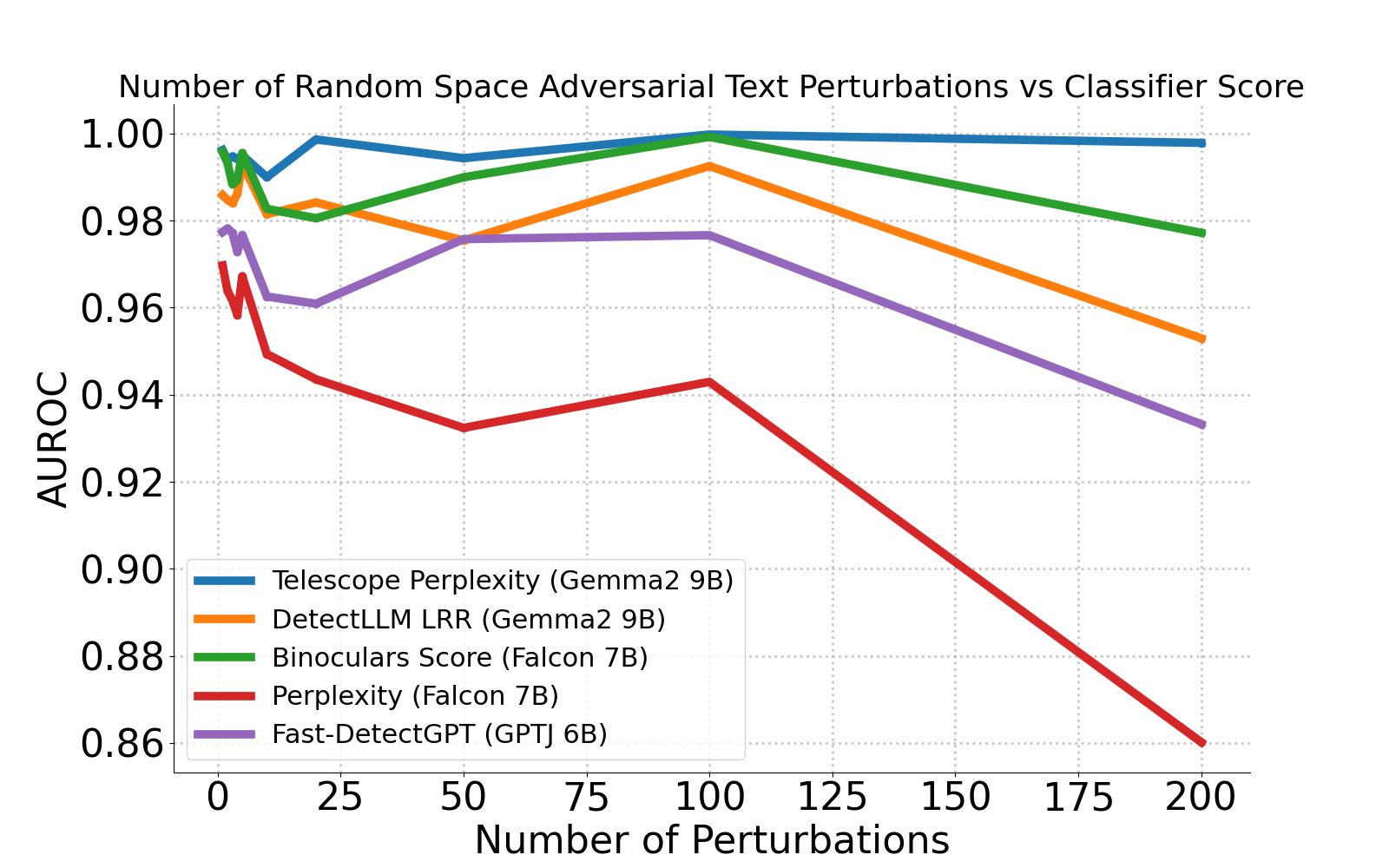}
        \caption{Random space}
        \label{fig:appendices_perturb_char_space}
    \end{subfigure}
    \hfill
    \begin{subfigure}[b]{0.48\textwidth}
        \centering
        \includegraphics[width=\textwidth]{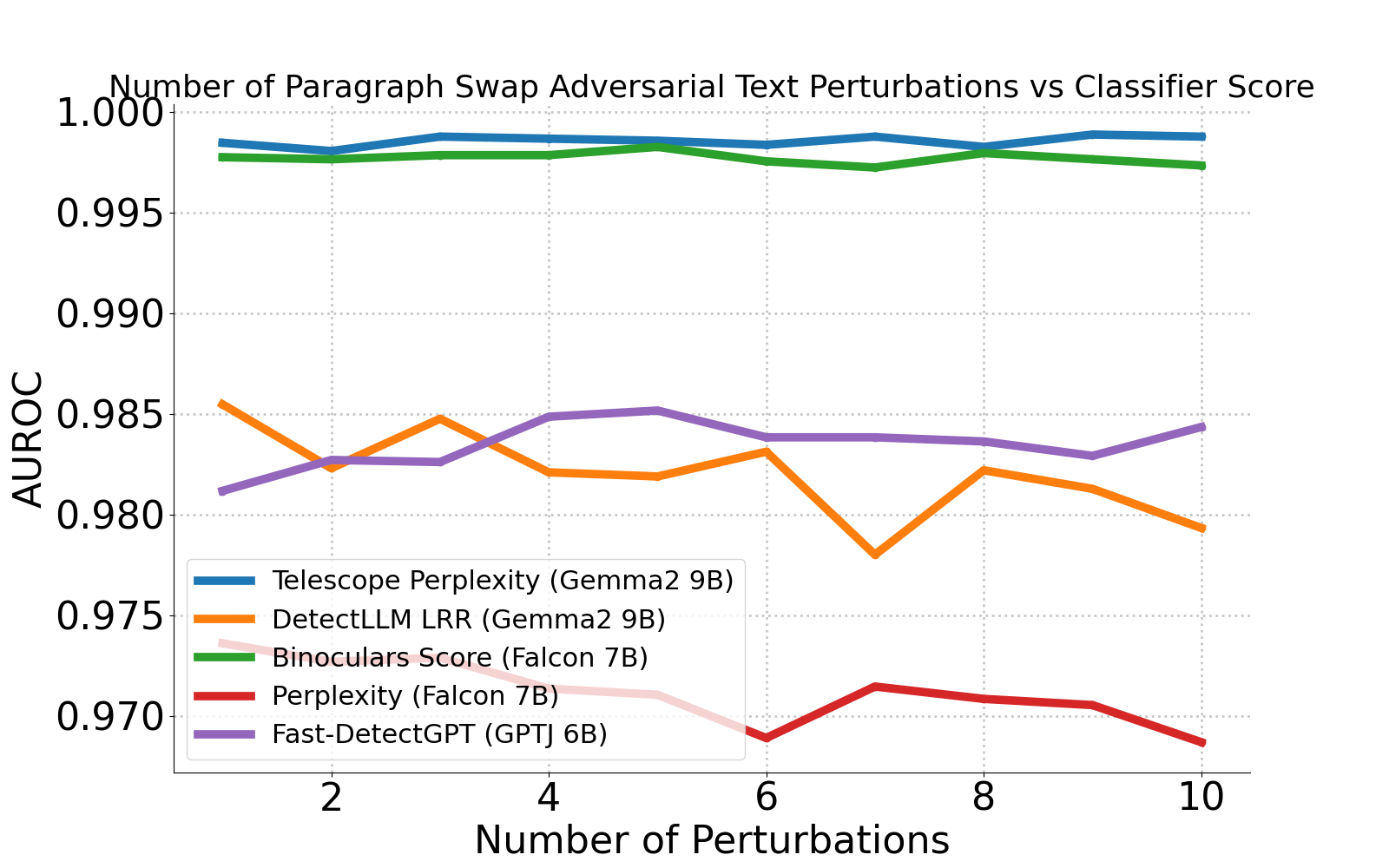}
        \caption{Paragraph reordering}
        \label{fig:appendices_perturb_paragraph}
    \end{subfigure}
    
    \vspace{1.5em} 
    
    \begin{subfigure}[b]{0.48\textwidth}
        \centering
        \includegraphics[width=\textwidth]{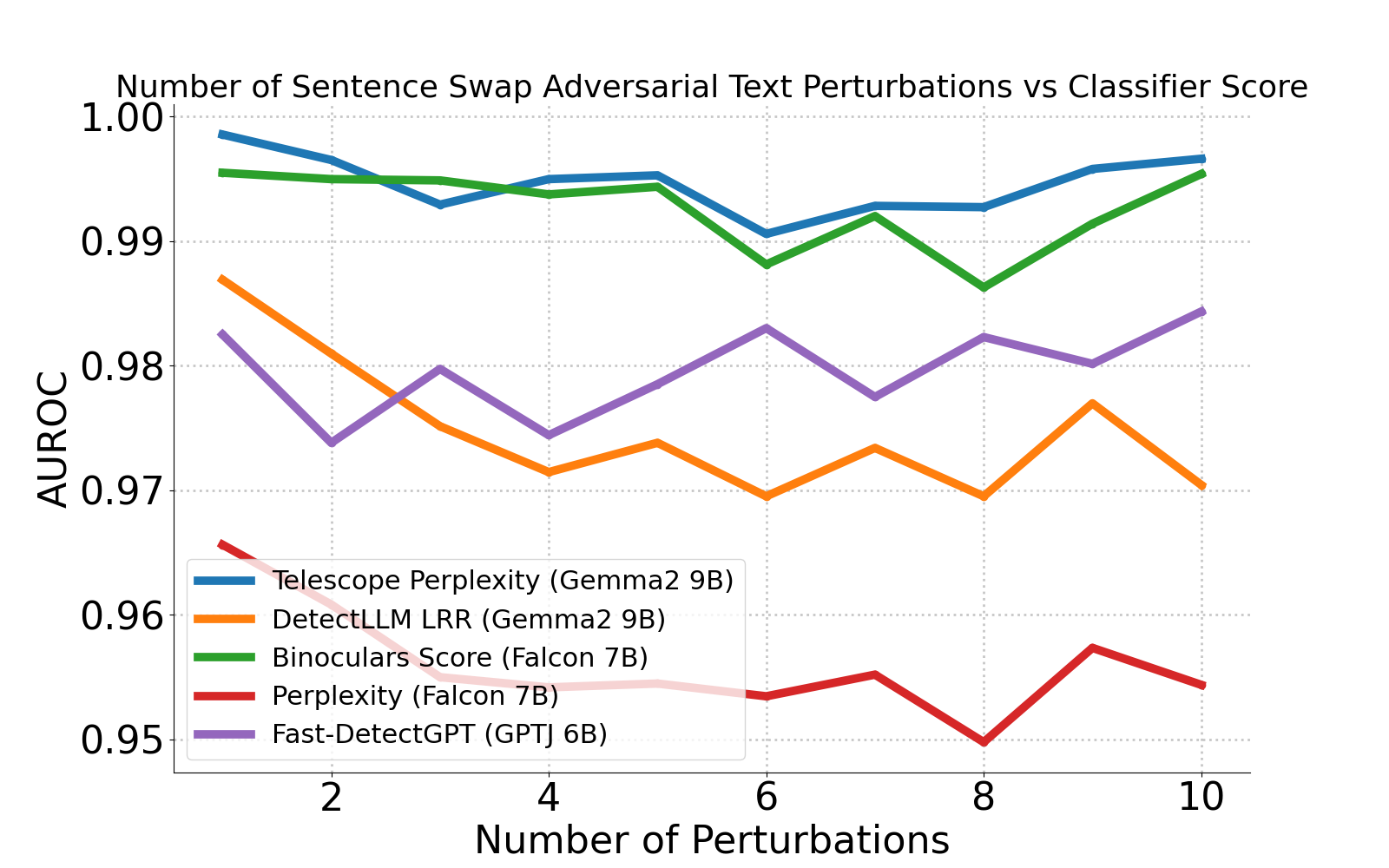}
        \caption{Sentence reordering}
        \label{fig:appendices_perturb_sentence}
    \end{subfigure}
    
    \caption{Impact of random space, paragraph reordering, and sentence reordering perturbations on the AUROC of each detector.}
    \label{fig:combined_perturbations}
\end{figure*}

Figures \ref{fig:appendices_text_length_hc3}, \ref{fig:appendices_text_length_esl}, \ref{fig:appendices_text_length_gb_creative}, and \ref{fig:appendices_text_length_gb_essay} show model performance over text length. Figures \ref{fig:appendices_perturb_char_basic}, 
\ref{fig:appendices_perturb_char_caps}, \ref{fig:appendices_perturb_char_space}, \ref{fig:appendices_perturb_paragraph}, and \ref{fig:appendices_perturb_sentence} show model performance over perturbations for a number of different perturbation regimes described in \citet{verma2024ghostbusterdetectingtextghostwritten}.

\clearpage

\subsection{Raw Experimental Results}

\label{sec:full_results}
\begin{strip}
    \captionof{table}{Detection performance of Telescope Perplexity, Binoculars, Perplexity, DetectLLM LRR, and Fast-DetectGPT across datasets and reference models. We report the AUROC of each reference model on each dataset and detection technique. The best performance on a dataset is bolded. 
    }
    \centering
    \resizebox{13cm}{!}{
\begin{tabular}{l|lcccccc}
        \toprule
        \multirow{2}{*}{Dataset} & \multirow{2}{*}{Reference Model} & \multicolumn{5}{c}{AUROC} \\
        & & Telescope & Binoculars & Perplexity & DetectLLM LRR & Fast-DetectGPT\\
        
    \midrule    
    \multirow{12}{*}{Detect LLM Text}

    & Gemma2 2B & \textbf{0.99639} & 0.93494 & 0.90786 & 0.94416 & 0.91596  \\
    & Gemma2 9B & \textbf{0.99858} & 0.95208 & 0.85133 & 0.97735 & 0.93209  \\
    & Llama3 8B & \textbf{0.99296} & 0.83669 & 0.93477 & 0.88760 & 0.76157  \\
    & Falcon 7B & \textbf{0.98158} & 0.84827 & 0.88921 & 0.91568 & 0.85181  \\
    & SmolLM 135M & \textbf{0.99547} & 0.66195 & 0.90323 & 0.96461 & 0.56232  \\
    & SmolLM 360M & \textbf{0.99618} & 0.65566 & 0.92177 & 0.96672 & 0.52354  \\
\bottomrule
    \end{tabular}
    }
\end{strip}

\begin{table*}[t]
    \caption{Detection performance of Telescope Perplexity, Binoculars, Perplexity, DetectLLM LRR, and Fast-DetectGPT across datasets and reference models. We report the AUROC of each reference model on each dataset and detection technique. The best performance on a dataset is bolded. 
    }
    \centering
    \resizebox{14cm}{!}{
    \begin{tabular}{l|lcccccc}
        \toprule
        \multirow{2}{*}{Dataset} & \multirow{2}{*}{Reference Model} & \multicolumn{5}{c}{AUROC} \\
        & & Telescope & Binoculars & Perplexity & DetectLLM LRR & Fast-DetectGPT\\
        \midrule
\multirow{12}{*}{GB Essay ChatGPT}

& Gemma2 2B & \textbf{0.99987} & 0.99845 & 0.99979 & 0.99868 & 0.98578  \\
& Gemma2 9B & \textbf{1.00000} & 0.99796 & 0.99922 & 0.99956 & 0.99347  \\
& Llama3 8B & \textbf{0.99999} & 0.90958 & 0.99953 & 0.99878 & 0.62873  \\
& Falcon 7B & 0.99846 & \textbf{0.99991} & 0.99932 & 0.99698 & 0.97445  \\
& SmolLM 135M & \textbf{0.99994} & 0.80145 & 0.99924 & 0.99726 & 0.21200  \\
& SmolLM 360M & \textbf{1.00000} & 0.62785 & 0.99938 & 0.99898 & 0.07174  \\
\bottomrule
    \end{tabular}
    }
\end{table*}

\begin{table*}[t]
    \caption{Detection performance of Telescope Perplexity, Binoculars, Perplexity, DetectLLM LRR, and Fast-DetectGPT across datasets and reference models. We report the AUROC of each reference model on each dataset and detection technique. The best performance on a dataset is bolded. 
    }
    \centering
    \resizebox{14cm}{!}{
    \begin{tabular}{l|lcccccc}
        \toprule
        \multirow{2}{*}{Dataset} & \multirow{2}{*}{Reference Model} & \multicolumn{5}{c}{AUROC} \\
        & & Telescope & Binoculars & Perplexity & DetectLLM LRR & Fast-DetectGPT\\

\midrule
\multirow{12}{*}{GB News Claude}

& Gemma2 2B & 0.92326 & \textbf{0.94816} & 0.92999 & 0.92280 & 0.80865  \\
& Gemma2 9B & \textbf{0.98566} & 0.98377 & 0.90450 & 0.97768 & 0.85818  \\
& Llama3 8B & 0.94244 & 0.89252 & \textbf{0.96908} & 0.92132 & 0.69805  \\
& Falcon 7B & 0.69352 & \textbf{0.96214} & 0.92459 & 0.89289 & 0.92435  \\
& SmolLM 135M & \textbf{0.86539} & 0.83901 & 0.79028 & 0.78966 & 0.69045  \\
& SmolLM 360M & \textbf{0.93531} & 0.81669 & 0.82855 & 0.85974 & 0.65129  \\
\bottomrule
    \end{tabular}
    }
\end{table*}

\begin{table*}[t]
    \caption{Detection performance of Telescope Perplexity, Binoculars, Perplexity, DetectLLM LRR, and Fast-DetectGPT across a variety of perturbation schemes and reference models. We report the AUROC of each reference model on each perturbation scheme and detection technique. The best performance on a dataset is bolded.  
    }
    \centering
    \resizebox{16cm}{!}{
    \begin{tabular}{l|lcccccc}
        \toprule
        \multirow{2}{*}{Dataset} & \multirow{2}{*}{Reference Model} & \multicolumn{5}{c}{AUROC} \\
        & & Telescope & Binoculars & Perplexity & DetectLLM LRR & Fast-DetectGPT\\
        \midrule
        
\multirow{12}{*}{GB Spelling Error Perturbation}

& Gemma2 2B & 0.93457 & \textbf{0.95360} & 0.91127 & 0.91503 & 0.95203  \\
& Gemma2 9B & \textbf{0.98599} & 0.96196 & 0.91074 & 0.95631 & 0.94367  \\
& Llama3 8B & \textbf{0.95360} & 0.83518 & 0.93290 & 0.95245 & 0.52696  \\
& Falcon 7B & 0.81250 & \textbf{0.97345} & 0.89360 & 0.88848 & 0.95255  \\
& SmolLM 135M & \textbf{0.94586} & 0.44095 & 0.85347 & 0.87552 & 0.22272  \\
& SmolLM 360M & \textbf{0.95307} & 0.42454 & 0.87364 & 0.90855 & 0.25867  \\
\bottomrule
    \end{tabular}
    }
\end{table*}

\begin{table*}[t]
    \caption{Detection performance of Telescope Perplexity, Binoculars, Perplexity, DetectLLM LRR, and Fast-DetectGPT across a variety of perturbation schemes and reference models. We report the AUROC of each reference model on each perturbation scheme and detection technique. The best performance on a dataset is bolded.   
    }
    \centering
    \resizebox{17cm}{!}{
    \begin{tabular}{l|lcccccc}
        \toprule
        \multirow{2}{*}{Dataset} & \multirow{2}{*}{Reference Model} & \multicolumn{5}{c}{AUROC} \\
        & & Telescope & Binoculars & Perplexity & DetectLLM LRR & Fast-DetectGPT\\
        \midrule

\multirow{12}{*}{GB Swap Adjacent Sentences Perturbation}

& Gemma2 2B & \textbf{0.98824} & 0.98476 & 0.97432 & 0.96471 & 0.93054  \\
& Gemma2 9B & \textbf{0.99529} & 0.99131 & 0.96082 & 0.97279 & 0.94006  \\
& Llama3 8B & \textbf{0.97954} & 0.90671 & 0.97054 & 0.95714 & 0.63144  \\
& Falcon 7B & 0.93351 & \textbf{0.99448} & 0.95448 & 0.93781 & 0.97647  \\
& SmolLM 135M & \textbf{0.99325} & 0.83449 & 0.96072 & 0.97719 & 0.46819  \\
& SmolLM 360M & \textbf{0.99223} & 0.80953 & 0.96195 & 0.98425 & 0.46389  \\
\bottomrule
    \end{tabular}
    }
\end{table*}

\begin{table*}[t]
    \caption{Two standard deviation confidence interval of Telescope Perplexity, Binoculars, Perplexity, DetectLLM LRR, and Fast-DetectGPT across datasets and reference models. We report the confidence interval of the AUROC of each reference model on each perturbation scheme and detection technique.
    }
    \centering
    \resizebox{16cm}{!}{
    \begin{tabular}{l|lcccccc}
        \toprule
        \multirow{2}{*}{Dataset} & \multirow{2}{*}{Reference Model} & \multicolumn{5}{c}{AUROC} \\
        & & Telescope & Binoculars & Perplexity & DetectLLM LRR & Fast-DetectGPT\\
        \midrule

\multirow{12}{*}{Detect LLM Text}

& Gemma2 2B & (0.99549, 0.99729) & (0.93036, 0.93952) & (0.90215, 0.91356) & (0.93980, 0.94852) & (0.91052, 0.92141)  \\
& Gemma2 9B & (0.99800, 0.99916) & (0.94830, 0.95585) & (0.84376, 0.85890) & (0.97478, 0.97991) & (0.92731, 0.93687)  \\
& Llama3 8B & (0.99181, 0.99412) & (0.82836, 0.84502) & (0.93030, 0.93924) & (0.88108, 0.89412) & (0.75156, 0.77157)  \\
& Falcon 7B & (0.97936, 0.98379) & (0.83910, 0.85744) & (0.88255, 0.89586) & (0.91000, 0.92137) & (0.84356, 0.86006)  \\
& SmolLM 135M & (0.99439, 0.99656) & (0.64912, 0.67478) & (0.89727, 0.90920) & (0.96130, 0.96793) & (0.54974, 0.57489)  \\
\bottomrule
    \end{tabular}
    }
\end{table*}

\begin{table*}[t]
    \caption{Two standard deviation confidence interval of Telescope Perplexity, Binoculars, Perplexity, DetectLLM LRR, and Fast-DetectGPT across datasets and reference models. We report the confidence interval of the AUROC of each reference model on each perturbation scheme and detection technique.
    }
    \centering
    \resizebox{15cm}{!}{
    \begin{tabular}{l|lcccccc}
        \toprule
        \multirow{2}{*}{Dataset} & \multirow{2}{*}{Reference Model} & \multicolumn{5}{c}{AUROC} \\
        & & Telescope & Binoculars & Perplexity & DetectLLM LRR & Fast-DetectGPT\\
        \midrule

\multirow{12}{*}{GB Essay ChatGPT}

& Gemma2 2B & (0.99974, 1.00001) & (0.99765, 0.99924) & (0.99939, 1.00019) & (0.99766, 0.99970) & (0.98195, 0.98962)  \\
& Gemma2 9B & (0.99999, 1.00000) & (0.99694, 0.99898) & (0.99839, 1.00006) & (0.99923, 0.99989) & (0.99103, 0.99591)  \\
& Llama3 8B & (0.99998, 1.00001) & (0.89672, 0.92244) & (0.99892, 1.00014) & (0.99761, 0.99995) & (0.60431, 0.65316)  \\
& Falcon 7B & (0.99746, 0.99945) & (0.99982, 1.00000) & (0.99847, 1.00018) & (0.99541, 0.99856) & (0.96788, 0.98103)  \\
& SmolLM 135M & (0.99988, 1.00000) & (0.78211, 0.82079) & (0.99865, 0.99983) & (0.99589, 0.99863) & (0.19217, 0.23184)  \\
\bottomrule
    \end{tabular}
    }
\end{table*}

\begin{table*}[t]
    \caption{Two standard deviation confidence interval of Telescope Perplexity, Binoculars, Perplexity, DetectLLM LRR, and Fast-DetectGPT across datasets and reference models. We report the confidence interval of the AUROC of each reference model on each perturbation scheme and detection technique.
    }
    \centering
    \resizebox{17cm}{!}{
    \begin{tabular}{l|lcccccc}
        \toprule
        \multirow{2}{*}{Dataset} & \multirow{2}{*}{Reference Model} & \multicolumn{5}{c}{AUROC} \\
        & & Telescope & Binoculars & Perplexity & DetectLLM LRR & Fast-DetectGPT\\
        \midrule

\multirow{12}{*}{GB Creative Claude}

& Gemma2 2B & (0.96049, 0.97455) & (0.94717, 0.96422) & (0.92490, 0.94578) & (0.90933, 0.93242) & (0.80374, 0.84019)  \\
& Gemma2 9B & (0.99398, 0.99876) & (0.97323, 0.98428) & (0.93178, 0.95168) & (0.97222, 0.98310) & (0.86452, 0.89450)  \\
& Llama3 8B & (0.98991, 0.99553) & (0.79302, 0.83197) & (0.96541, 0.97879) & (0.93697, 0.95524) & (0.39983, 0.45100)  \\
& Falcon 7B & (0.96698, 0.97903) & (0.90546, 0.93016) & (0.88176, 0.90971) & (0.84342, 0.87551) & (0.82852, 0.86276)  \\
& SmolLM 135M & (0.95176, 0.96755) & (0.77757, 0.81671) & (0.81058, 0.84661) & (0.82561, 0.85976) & (0.36485, 0.41419)  \\
\bottomrule
    \end{tabular}
    }
\end{table*}

\begin{table*}[t]
    \caption{Two standard deviation confidence interval of Telescope Perplexity, Binoculars, Perplexity, DetectLLM LRR, and Fast-DetectGPT across a variety of perturbation schemes and reference models. We report the confidence interval of the AUROC of each reference model on each perturbation scheme and detection technique.
    }
    \centering
    \resizebox{17cm}{!}{
    \begin{tabular}{l|lcccccc}
        \toprule
        \multirow{2}{*}{Dataset} & \multirow{2}{*}{Reference Model} & \multicolumn{5}{c}{AUROC} \\
        & & Telescope & Binoculars & Perplexity & DetectLLM LRR & Fast-DetectGPT\\
        \midrule

\multirow{12}{*}{GB Spelling Error Perturbation}

& Gemma2 2B & (0.89989, 0.96925) & (0.92012, 0.98707) & (0.86572, 0.95682) & (0.87086, 0.95920) & (0.92540, 0.97865)  \\
& Gemma2 9B & (0.97207, 0.99992) & (0.93764, 0.98627) & (0.86572, 0.95577) & (0.92041, 0.99221) & (0.91320, 0.97413)  \\
& Llama3 8B & (0.92188, 0.98531) & (0.77753, 0.89283) & (0.89162, 0.97418) & (0.91567, 0.98922) & (0.44545, 0.60848)  \\
& Falcon 7B & (0.75326, 0.87174) & (0.94422, 1.00269) & (0.84473, 0.94247) & (0.83952, 0.93745) & (0.92151, 0.98359)  \\
& SmolLM 135M & (0.91310, 0.97862) & (0.35586, 0.52604) & (0.79624, 0.91070) & (0.82338, 0.92767) & (0.14875, 0.29669)  \\
\bottomrule
    \end{tabular}
    }
\end{table*}

\begin{table*}[t]
    \caption{Two standard deviation confidence interval of Telescope Perplexity, Binoculars, Perplexity, DetectLLM LRR, and Fast-DetectGPT across a variety of perturbation schemes and reference models. We report the confidence interval of the AUROC of each reference model on each perturbation scheme and detection technique.
    }
    \centering
    \resizebox{17cm}{!}{
    \begin{tabular}{l|lcccccc}
        \toprule
        \multirow{2}{*}{Dataset} & \multirow{2}{*}{Reference Model} & \multicolumn{5}{c}{AUROC} \\
        & & Telescope & Binoculars & Perplexity & DetectLLM LRR & Fast-DetectGPT\\
        \midrule

\multirow{12}{*}{GB Swap Adjacent Sentences Perturbation}

& Gemma2 2B & (0.97790, 0.99857) & (0.96978, 0.99973) & (0.94978, 0.99887) & (0.93792, 0.99149) & (0.89712, 0.96397)  \\
& Gemma2 9B & (0.98880, 1.00179) & (0.98373, 0.99888) & (0.93414, 0.98750) & (0.95097, 0.99461) & (0.90950, 0.97061)  \\
& Llama3 8B & (0.96177, 0.99732) & (0.86639, 0.94703) & (0.94562, 0.99546) & (0.92639, 0.98789) & (0.55384, 0.70905)  \\
& Falcon 7B & (0.89741, 0.96961) & (0.98723, 1.00172) & (0.92433, 0.98463) & (0.90332, 0.97230) & (0.95585, 0.99709)  \\
& SmolLM 135M & (0.98690, 0.99960) & (0.77561, 0.89338) & (0.93156, 0.98988) & (0.95636, 0.99802) & (0.38334, 0.55303)  \\
\bottomrule
    \end{tabular}
    }
\end{table*}

\begin{table*}[t]
    \caption{Detection performance of Telescope Perplexity, Binoculars, Perplexity, DetectLLM LRR, and Fast-DetectGPT when transfering a classification threshold from all of the other Ghostbusters datasets to this dataset. We report the F1-Score of each reference model on each test dataset.}
    \centering
    \resizebox{14cm}{!}{
    \begin{tabular}{l|lcccccc}
        \toprule
        \multirow{2}{*}{Test Dataset} & \multirow{2}{*}{Reference Model} & \multicolumn{5}{c}{F1-Score} \\
        & & Telescope & Binoculars & Perplexity & DetectLLM LRR & Fast-DetectGPT\\
        
\midrule
\multirow{12}{*}{GB Essay ChatGPT}

& Gemma2 2B & \textbf{0.98331} & 0.97774 & 0.98128 & 0.97623 & 0.90996  \\
& Gemma2 9B & 0.97724 & 0.97521 & 0.98331 & \textbf{0.98887} & 0.93627  \\
& Llama3 8B & \textbf{0.99140} & 0.83308 & 0.96965 & 0.97673 & 0.59737  \\
& Falcon 7B & 0.92757 & \textbf{0.97775} & 0.96586 & 0.96793 & 0.91878  \\
& SmolLM 135M & \textbf{0.93374} & 0.72534 & 0.85230 & 0.86191 & 0.70157  \\
& SmolLM 360M & \textbf{0.96105} & 0.60445 & 0.87759 & 0.90794 & 0.78705  \\
\bottomrule
    \end{tabular}
    }
\end{table*}

\begin{table*}[t]
    \caption{Detection performance of Telescope Perplexity, Binoculars, Perplexity, DetectLLM LRR, and Fast-DetectGPT when transfering a classification threshold from all of the other Ghostbusters datasets to this dataset. We report the F1-Score of each reference model on each test dataset.}
    \centering
    \resizebox{14cm}{!}{
    \begin{tabular}{l|lcccccc}
        \toprule
        \multirow{2}{*}{Test Dataset} & \multirow{2}{*}{Reference Model} & \multicolumn{5}{c}{F1-Score} \\
        & & Telescope & Binoculars & Perplexity & DetectLLM LRR & Fast-DetectGPT\\

\midrule
\multirow{12}{*}{GB News Claude}

& Gemma2 2B & 0.78034 & \textbf{0.88465} & 0.80441 & 0.79238 & 0.67503  \\
& Gemma2 9B & 0.89819 & \textbf{0.93330} & 0.72217 & 0.86108 & 0.71063  \\
& Llama3 8B & \textbf{0.86409} & 0.85005 & 0.84554 & 0.82297 & 0.66851  \\
& Falcon 7B & 0.61840 & \textbf{0.81818} & 0.79314 & 0.77518 & 0.80348  \\
& SmolLM 135M & 0.57472 & \textbf{0.73521} & 0.69960 & 0.63842 & 0.38415  \\
& SmolLM 360M & 0.70712 & \textbf{0.75677} & 0.74975 & 0.72518 & 0.42126  \\
\bottomrule
    \end{tabular}
    }
\end{table*}

\begin{table*}[t]
    \caption{Two standard deviation confidence interval of Telescope Perplexity, Binoculars, Perplexity, DetectLLM LRR, and Fast-DetectGPT when transfering a classification threshold from all of the other Ghostbusters datasets to this dataset. We report the Confidence Interval of the F1-Score of each reference model on each test dataset.}
    \centering
    \resizebox{17cm}{!}{
    \begin{tabular}{l|lcccccc}
        \toprule
        \multirow{2}{*}{Test Dataset} & \multirow{2}{*}{Reference Model} & \multicolumn{5}{c}{F1-Score} \\
        & & Telescope & Binoculars & Perplexity & DetectLLM LRR & Fast-DetectGPT\\

\midrule
\multirow{12}{*}{GB Essay ChatGPT}

& Gemma2 2B & (0.97766, 0.98896) & (0.97124, 0.98425) & (0.97531, 0.98726) & (0.96951, 0.98294) & (0.89735, 0.92258)  \\
& Gemma2 9B & (0.97066, 0.98381) & (0.96836, 0.98207) & (0.97766, 0.98896) & (0.98425, 0.99350) & (0.92550, 0.94703)  \\
& Llama3 8B & (0.98733, 0.99547) & (0.81664, 0.84952) & (0.96209, 0.97721) & (0.97009, 0.98338) & (0.57575, 0.61899)  \\
& Falcon 7B & (0.91602, 0.93913) & (0.97118, 0.98433) & (0.95776, 0.97395) & (0.96007, 0.97578) & (0.90660, 0.93096)  \\
& SmolLM 135M & (0.92277, 0.94470) & (0.70567, 0.74502) & (0.83666, 0.86794) & (0.84670, 0.87712) & (0.68140, 0.72174)  \\
\bottomrule
    \end{tabular}
    }
\end{table*}

\begin{table*}[t]
    \caption{Two standard deviation confidence interval of Telescope Perplexity, Binoculars, Perplexity, DetectLLM LRR, and Fast-DetectGPT when transfering a classification threshold from all of the other Ghostbusters datasets to this dataset. We report the Confidence Interval of the F1-Score of each reference model on each test dataset.}
    \centering
    \resizebox{14cm}{!}{
    \begin{tabular}{l|lcccccc}
        \toprule
        \multirow{2}{*}{Test Dataset} & \multirow{2}{*}{Reference Model} & \multicolumn{5}{c}{F1-Score} \\
        & & Telescope & Binoculars & Perplexity & DetectLLM LRR & Fast-DetectGPT\\

        \midrule

\midrule
\multirow{12}{*}{GB Creative GPT4o}

& Gemma2 2B & (0.94462, 0.96313) & (0.94518, 0.96359) & (0.84004, 0.87106) & (0.75706, 0.79388) & (0.62513, 0.66732)  \\
& Gemma2 9B & (0.98668, 0.99507) & (0.90794, 0.93190) & (0.78581, 0.82088) & (0.80628, 0.83995) & (0.60511, 0.64780)  \\
& Llama3 8B & (0.98361, 0.99308) & (0.89768, 0.92290) & (0.76593, 0.80224) & (0.77743, 0.81304) & (0.54937, 0.59305)  \\
& Falcon 7B & (0.94490, 0.96358) & (0.95870, 0.97474) & (0.66834, 0.70972) & (0.63768, 0.68006) & (0.93864, 0.95839)  \\
& SmolLM 135M & (0.95243, 0.96952) & (0.79052, 0.82529) & (0.64621, 0.68780) & (0.71752, 0.75638) & (0.58769, 0.63075)  \\
\bottomrule
    \end{tabular}
    }
\end{table*}


\begin{table*}[t]
    \captionof{table}{Detection performance of Telescope Perplexity, Binoculars, Perplexity, DetectLLM LRR, and Fast-DetectGPT across datasets and reference models. We report the true positive rate at a false positive rate of 5\% of each reference model on each dataset and detection technique. The best performance on a dataset is bolded. 
    }
    \centering
    \resizebox{13cm}{!}{
    \begin{tabular}{l|lcccccc}
        \toprule
        \multirow{2}{*}{Dataset} & \multirow{2}{*}{Reference Model} & \multicolumn{5}{c}{True Positive Rate at False Positive Rate of 5\%} \\
        & & Telescope & Binoculars & Perplexity & DetectLLM LRR & Fast-DetectGPT\\

\midrule
\multirow{12}{*}{Detect LLM Text}

& Gemma2 2B & \textbf{0.98689} & 0.70103 & 0.65742 & 0.76557 & 0.62869  \\
& Gemma2 9B & \textbf{0.99622} & 0.75296 & 0.55029 & 0.89463 & 0.65516  \\
& Llama3 8B & \textbf{0.96723} & 0.53390 & 0.65793 & 0.57373 & 0.39551  \\
& Falcon 7B & \textbf{0.91009} & 0.72005 & 0.65798 & 0.64725 & 0.62759  \\
& SmolLM 135M & \textbf{0.98412} & 0.49508 & 0.65163 & 0.84825 & 0.26368  \\
& SmolLM 360M & \textbf{0.98538} & 0.43257 & 0.66473 & 0.84321 & 0.21553  \\
\bottomrule
    \end{tabular}
    }
\end{table*}

\begin{table*}[t]
    \caption{Detection performance of Telescope Perplexity, Binoculars, Perplexity, DetectLLM LRR, and Fast-DetectGPT across datasets and reference models. We report the true positive rate at a false positive rate of 5\% of each reference model on each dataset and detection technique. The best performance on a dataset is bolded. 
    }
    \centering
    \resizebox{14cm}{!}{
    \begin{tabular}{l|lcccccc}
        \toprule
        \multirow{2}{*}{Dataset} & \multirow{2}{*}{Reference Model} & \multicolumn{5}{c}{True Positive Rate at False Positive Rate of 5\%} \\
        & & Telescope & Binoculars & Perplexity & DetectLLM LRR & Fast-DetectGPT\\

\midrule
\multirow{12}{*}{GB Essay ChatGPT}

& Gemma2 2B & \textbf{1.00000} & 0.99200 & \textbf{1.00000} & 0.99800 & 0.92700  \\
& Gemma2 9B & \textbf{1.00000} & 0.99600 & \textbf{1.00000} & 0.99700 & 0.97700  \\
& Llama3 8B & \textbf{1.00000} & 0.66800 & \textbf{1.00000} & 0.99800 & 0.15400  \\
& Falcon 7B & 0.99899 & \textbf{1.00000} & 0.99899 & 0.99194 & 0.90927  \\
& SmolLM 135M & \textbf{1.00000} & 0.42800 & 0.99900 & 0.99000 & 0.00800  \\
& SmolLM 360M & \textbf{1.00000} & 0.24200 & 0.99900 & 0.99500 & 0.00000  \\
\bottomrule
    \end{tabular}
    }
\end{table*}

\begin{table*}[t]
    \caption{Detection performance of Telescope Perplexity, Binoculars, Perplexity, DetectLLM LRR, and Fast-DetectGPT across datasets and reference models. We report the true positive rate at a false positive rate of 5\% of each reference model on each dataset and detection technique. The best performance on a dataset is bolded. 
    }
    \centering
    \resizebox{14cm}{!}{
    \begin{tabular}{l|lcccccc}
        \toprule
        \multirow{2}{*}{Dataset} & \multirow{2}{*}{Reference Model} & \multicolumn{5}{c}{True Positive Rate at False Positive Rate of 5\%} \\
        & & Telescope & Binoculars & Perplexity & DetectLLM LRR & Fast-DetectGPT\\

\midrule
\multirow{12}{*}{GB News Claude}

& Gemma2 2B & 0.65700 & \textbf{0.81400} & 0.65100 & 0.65500 & 0.40500  \\
& Gemma2 9B & \textbf{0.93700} & 0.93400 & 0.55600 & 0.90500 & 0.41500  \\
& Llama3 8B & 0.73300 & 0.73900 & \textbf{0.83200} & 0.63200 & 0.37800  \\
& Falcon 7B & 0.20278 & \textbf{0.82966} & 0.63615 & 0.56779 & 0.63499  \\
& SmolLM 135M & \textbf{0.52100} & 0.46900 & 0.27700 & 0.34900 & 0.17600  \\
& SmolLM 360M & \textbf{0.74700} & 0.45100 & 0.35500 & 0.49400 & 0.17900  \\
\bottomrule
    \end{tabular}
    }
\end{table*}

\begin{table*}[t]
    \caption{Detection performance of Telescope Perplexity, Binoculars, Perplexity, DetectLLM LRR, and Fast-DetectGPT across a variety of perturbation schemes and reference models. We report the true positive rate at a false positive rate of 5\% of each reference model on each dataset and detection technique. The best performance on a dataset is bolded. 
    }
    \centering
    \resizebox{16cm}{!}{
    \begin{tabular}{l|lcccccc}
        \toprule
        \multirow{2}{*}{Dataset} & \multirow{2}{*}{Reference Model} & \multicolumn{5}{c}{True Positive Rate at False Positive Rate of 5\%} \\
        & & Telescope & Binoculars & Perplexity & DetectLLM LRR & Fast-DetectGPT\\

\midrule
\multirow{12}{*}{GB Spelling Error Perturbation}

& Gemma2 2B & 0.77174 & \textbf{0.90217} & 0.77174 & 0.75000 & 0.76087  \\
& Gemma2 9B & \textbf{0.94565} & 0.86957 & 0.73913 & 0.92391 & 0.79348  \\
& Llama3 8B & \textbf{0.89130} & 0.52174 & 0.82609 & \textbf{0.89130} & 0.10870  \\
& Falcon 7B & 0.51087 & \textbf{0.95652} & 0.72826 & 0.65217 & 0.88043  \\
& SmolLM 135M & \textbf{0.84783} & 0.18478 & 0.64130 & 0.70652 & 0.09783  \\
& SmolLM 360M & \textbf{0.85870} & 0.23913 & 0.70652 & 0.80435 & 0.14130  \\
\bottomrule
    \end{tabular}
    }
\end{table*}

\begin{table*}[t]
    \caption{Detection performance of Telescope Perplexity, Binoculars, Perplexity, DetectLLM LRR, and Fast-DetectGPT across a variety of perturbation schemes and reference models. We report the true positive rate at a false positive rate of 5\% of each reference model on each dataset and detection technique. The best performance on a dataset is bolded. 
    }
    \centering
    \resizebox{17cm}{!}{
    \begin{tabular}{l|lcccccc}
        \toprule
        \multirow{2}{*}{Dataset} & \multirow{2}{*}{Reference Model} & \multicolumn{5}{c}{True Positive Rate at False Positive Rate of 5\%} \\
        & & Telescope & Binoculars & Perplexity & DetectLLM LRR & Fast-DetectGPT\\

\midrule
\multirow{12}{*}{GB Swap Adjacent Sentences Perturbation}

& Gemma2 2B & 0.93617 & \textbf{0.95745} & 0.92553 & 0.85106 & 0.70213  \\
& Gemma2 9B & \textbf{0.97872} & 0.94681 & 0.86170 & 0.90426 & 0.68085  \\
& Llama3 8B & \textbf{0.94681} & 0.68085 & 0.91489 & 0.87234 & 0.21277  \\
& Falcon 7B & 0.80851 & \textbf{0.97872} & 0.85106 & 0.77660 & 0.92553  \\
& SmolLM 135M & \textbf{0.94681} & 0.55319 & 0.89362 & 0.90426 & 0.18085  \\
& SmolLM 360M & \textbf{0.95745} & 0.57447 & 0.90426 & 0.94681 & 0.19149  \\
\bottomrule
    \end{tabular}
    }
\end{table*}

\begin{table*}[t]
    \caption{Two standard deviation confidence interval of Telescope Perplexity, Binoculars, Perplexity, DetectLLM LRR, and Fast-DetectGPT across datasets and reference models. We report the confidence interval of the true positive rate at a false positive rate of 5\% for each reference model on each perturbation scheme and detection technique.
    }
    \centering
    \resizebox{16cm}{!}{
    \begin{tabular}{l|lcccccc}
        \toprule
        \multirow{2}{*}{Dataset} & \multirow{2}{*}{Reference Model} & \multicolumn{5}{c}{True Positive Rate at False Positive Rate of 5\%} \\
        & & Telescope & Binoculars & Perplexity & DetectLLM LRR & Fast-DetectGPT\\

\midrule
\multirow{12}{*}{Detect LLM Text}

& Gemma2 2B & (0.98336, 0.99143) & (0.67960, 0.71894) & (0.64254, 0.67357) & (0.75145, 0.78171) & (0.60599, 0.65592)  \\
& Gemma2 9B & (0.99395, 0.99798) & (0.73305, 0.77136) & (0.53617, 0.57020) & (0.88177, 0.90774) & (0.63119, 0.68164)  \\
& Llama3 8B & (0.96017, 0.97480) & (0.51549, 0.55357) & (0.63473, 0.67835) & (0.55709, 0.59692) & (0.37887, 0.41671)  \\
& Falcon 7B & (0.89423, 0.92210) & (0.70625, 0.73615) & (0.64291, 0.67357) & (0.62119, 0.67280) & (0.61149, 0.64343)  \\
& SmolLM 135M & (0.97933, 0.98840) & (0.47820, 0.51248) & (0.63599, 0.66726) & (0.83160, 0.86413) & (0.24679, 0.28058)  \\
\bottomrule
    \end{tabular}
    }
\end{table*}

\begin{table*}[t]
    \caption{Two standard deviation confidence interval of Telescope Perplexity, Binoculars, Perplexity, DetectLLM LRR, and Fast-DetectGPT across datasets and reference models. We report the confidence interval of the true positive rate at a false positive rate of 5\% for each reference model on each perturbation scheme and detection technique.
    }
    \centering
    \resizebox{17cm}{!}{
    \begin{tabular}{l|lcccccc}
        \toprule
        \multirow{2}{*}{Dataset} & \multirow{2}{*}{Reference Model} & \multicolumn{5}{c}{True Positive Rate at False Positive Rate of 5\%} \\
        & & Telescope & Binoculars & Perplexity & DetectLLM LRR & Fast-DetectGPT\\

\midrule
\multirow{12}{*}{GB Essay ChatGPT}

& Gemma2 2B & (1.00000, 1.00000) & (0.98700, 0.99900) & (1.00000, 1.00000) & (0.99100, 1.00000) & (0.88398, 0.95100)  \\
& Gemma2 9B & (1.00000, 1.00000) & (0.98900, 0.99900) & (1.00000, 1.00000) & (0.99400, 1.00000) & (0.96198, 0.98600)  \\
& Llama3 8B & (1.00000, 1.00000) & (0.62100, 0.73400) & (1.00000, 1.00000) & (0.99300, 1.00000) & (0.12400, 0.19000)  \\
& Falcon 7B & (0.99698, 1.00000) & (1.00000, 1.00000) & (0.99698, 1.00000) & (0.98185, 0.99698) & (0.88808, 0.92744)  \\
& SmolLM 135M & (1.00000, 1.00000) & (0.37400, 0.48000) & (0.99600, 1.00000) & (0.98200, 0.99600) & (0.00300, 0.01600)  \\
\bottomrule
    \end{tabular}
    }
\end{table*}

\begin{table*}[t]
    \caption{Two standard deviation confidence interval of Telescope Perplexity, Binoculars, Perplexity, DetectLLM LRR, and Fast-DetectGPT across datasets and reference models. We report the confidence interval of the true positive rate at a false positive rate of 5\% for each reference model on each perturbation scheme and detection technique.
    }
    \centering
    \resizebox{17cm}{!}{
    \begin{tabular}{l|lcccccc}
        \toprule
        \multirow{2}{*}{Dataset} & \multirow{2}{*}{Reference Model} & \multicolumn{5}{c}{True Positive Rate at False Positive Rate of 5\%} \\
        & & Telescope & Binoculars & Perplexity & DetectLLM LRR & Fast-DetectGPT\\

\midrule
\multirow{12}{*}{GB News Claude}

& Gemma2 2B & (0.57700, 0.69600) & (0.78400, 0.85502) & (0.55200, 0.71602) & (0.58098, 0.71500) & (0.34400, 0.47600)  \\
& Gemma2 9B & (0.91900, 0.95800) & (0.90298, 0.95300) & (0.50595, 0.60702) & (0.87798, 0.92500) & (0.34395, 0.47302)  \\
& Llama3 8B & (0.65098, 0.76800) & (0.69700, 0.77905) & (0.76600, 0.88900) & (0.59700, 0.67700) & (0.33400, 0.41700)  \\
& Falcon 7B & (0.16107, 0.24450) & (0.77520, 0.86327) & (0.57819, 0.69293) & (0.50521, 0.61066) & (0.56663, 0.67442)  \\
& SmolLM 135M & (0.47897, 0.57600) & (0.39500, 0.53102) & (0.22800, 0.32800) & (0.26993, 0.40400) & (0.13500, 0.22100)  \\
\bottomrule
    \end{tabular}
    }
\end{table*}

\begin{table*}[t]
    \caption{Detection performance of Telescope Perplexity, Binoculars, Perplexity, DetectLLM LRR, and Fast-DetectGPT across a variety of perturbation schemes and reference models. We report the confidence interval of the true positive rate at a false positive rate of 5\% for each reference model on each perturbation scheme and detection technique.
    }
    \centering
    \resizebox{17cm}{!}{
    \begin{tabular}{l|lcccccc}
        \toprule
        \multirow{2}{*}{Dataset} & \multirow{2}{*}{Reference Model} & \multicolumn{5}{c}{True Positive Rate at False Positive Rate of 5\%} \\
        & & Telescope & Binoculars & Perplexity & DetectLLM LRR & Fast-DetectGPT\\

\midrule
\multirow{12}{*}{GB Spelling Error Perturbation}

& Gemma2 2B & (0.56522, 0.89130) & (0.78261, 0.96739) & (0.65217, 0.86957) & (0.64130, 0.85870) & (0.67391, 0.90217)  \\
& Gemma2 9B & (0.88043, 0.98913) & (0.77174, 0.93478) & (0.50000, 0.86957) & (0.85870, 0.97826) & (0.54348, 0.88043)  \\
& Llama3 8B & (0.69565, 0.95652) & (0.35870, 0.67391) & (0.73886, 0.91304) & (0.82609, 0.96739) & (0.04348, 0.18505)  \\
& Falcon 7B & (0.38043, 0.63043) & (0.90217, 1.00000) & (0.48913, 0.83696) & (0.54348, 0.79348) & (0.73913, 0.94565)  \\
& SmolLM 135M & (0.71739, 0.91304) & (0.09783, 0.26087) & (0.43478, 0.77174) & (0.51087, 0.80462) & (0.04348, 0.16304)  \\
\bottomrule
    \end{tabular}
    }
\end{table*}

\begin{table*}[t]
    \caption{Detection performance of Telescope Perplexity, Binoculars, Perplexity, DetectLLM LRR, and Fast-DetectGPT across a variety of perturbation schemes and reference models. We report the confidence interval of the true positive rate at a false positive rate of 5\% for each reference model on each perturbation scheme and detection technique.
    }
    \centering
    \resizebox{17cm}{!}{
    \begin{tabular}{l|lcccccc}
        \toprule
        \multirow{2}{*}{Dataset} & \multirow{2}{*}{Reference Model} & \multicolumn{5}{c}{True Positive Rate at False Positive Rate of 5\%} \\
        & & Telescope & Binoculars & Perplexity & DetectLLM LRR & Fast-DetectGPT\\

\midrule
\multirow{12}{*}{GB Swap Adjacent Sentences Perturbation}

& Gemma2 2B & (0.88298, 0.97872) & (0.88298, 0.98936) & (0.80851, 0.98936) & (0.76596, 0.94681) & (0.53165, 0.87234)  \\
& Gemma2 9B & (0.94681, 1.00000) & (0.84043, 1.00000) & (0.72340, 0.94681) & (0.82979, 0.95745) & (0.47872, 0.89362)  \\
& Llama3 8B & (0.89362, 0.98936) & (0.54255, 0.77686) & (0.84043, 0.96809) & (0.79787, 0.94681) & (0.12766, 0.33005)  \\
& Falcon 7B & (0.71277, 0.89362) & (0.93617, 1.00000) & (0.72340, 0.93617) & (0.69122, 0.89362) & (0.85106, 0.97872)  \\
& SmolLM 135M & (0.89335, 1.00000) & (0.43617, 0.71303) & (0.79787, 0.94707) & (0.84043, 0.97872) & (0.10638, 0.25532)  \\
\bottomrule
    \end{tabular}
    }
\end{table*}



\end{document}